\documentclass[journal,twoside,web]{ieeecolor}
\usepackage{generic}
\usepackage{cite}
\usepackage{amsmath,amssymb,amsfonts}
\usepackage{algorithm,algorithmic}
\usepackage[pdftex]{graphicx}
\usepackage[hidelinks]{hyperref}
\usepackage{textcomp}
\usepackage{array}
\usepackage{booktabs}
\usepackage{url}
\usepackage{multirow}
\usepackage{subcaption}
\usepackage{rotating}
\usepackage{caption}
\usepackage{tabularx}
\usepackage{xcolor}
\usepackage{colortbl}
\definecolor{VLMRow}{gray}{0.92}
\usepackage{makecell}
\usepackage{stfloats}

\definecolor{IEEEBLUE}{RGB}{0, 98, 155}

\newcolumntype{Y}{>{\centering\arraybackslash}X}

\begin{document}

\title {VL-AcneSeg: A Vision-Language Framework for Region-Aware Acne Lesion Segmentation}

\markboth{IEEE JOURNAL OF BIOMEDICAL AND HEALTH INFORMATICS}%
{Oh \MakeLowercase{\textit{et al.}}: VL-AcneSeg: A Vision-Language Framework for Region-Aware Acne Lesion Segmentation}%

\author{Sukju~Oh,
        Soo~Ick~Cho,
        Dae~Hun~Suh,
        and~Sukkyu~Sun
\thanks{\copyright~2026 IEEE. Accepted for publication in the IEEE Journal of
Biomedical and Health Informatics. Personal use of this material is permitted;
all other uses require permission from IEEE.}%
\thanks{This research was supported by the ``Regional Innovation System \& Education (RISE)'' through the Seoul RISE Center, funded by the MOE (Ministry of Education) and the Seoul Metropolitan Government, and conducted in collaboration with Inskinlab Co., Ltd. (https://inskinlab.com/) (2026-RISE-01-007-05). This work was also supported by the Artificial Intelligence Convergence Innovation Human Resources Development supervised by the MSIT (Ministry of Science and ICT) and the IITP (Institute for Information \& Communications Technology Planning \& Evaluation) (IITP-2026-RS-2023-00254592). The authors thank Dong Hyo Kim, MD and Ji Won Lee, MD for the lesion annotations produced in the course of the study reported in \cite{kim2023automated}. (Corresponding author: Sukkyu Sun.)}%
\thanks{Ethical Approval: This study was approved by the Institutional Review Board of Seoul National University Hospital (No. 2510-149-1690) and conducted in accordance with the Declaration of Helsinki.}
\thanks{Sukju Oh and Sukkyu Sun are with the Department of Computer Science and Artificial Intelligence, Dongguk University, Seoul 04620, Republic of Korea (e-mail: dhtjrwn119@dgu.ac.kr; sukkyu.sun@dgu.ac.kr).}%
\thanks{Soo Ick Cho is with InSkin Lab Inc., Seoul 06193, Republic of Korea (e-mail: sooickcho@inskinlab.com).}
\thanks{Dae Hun Suh is with the Department of Dermatology, Seoul National University College of Medicine, and also with the Department of Dermatology, Seoul National University Hospital, Seoul 03080, Republic of Korea (e-mail: daehun@snu.ac.kr).}}

\maketitle

\begin{abstract}
Acne assessment is crucial for clinical decision-making, yet traditional grading and counting are subjective and fail to account for lesion size. While area-based assessment has emerged as a promising alternative, acne segmentation has continued to rely on general-purpose architectures. To address this gap, we propose VL-AcneSeg, a multimodal framework for acne lesion segmentation that leverages CLIP and region-level text prompts to incorporate spatial priors, enabling lesions to be localized across the whole face. Because region-level prompts indicate which facial areas contain lesions, we report a single global prompt, which requires no such information, as our primary setting. On our internal clinical dataset, VL-AcneSeg achieves a Dice score of 0.5082 and an IoU of 0.3407 under this protocol, the highest among all compared methods, including recent vision-language segmentation methods that are themselves given region-level prompts; region-level prompting raises these to 0.5296 and 0.3602. Moreover, lesion area measurements derived from our segmentation correlate with IGA scores at a level comparable to expert annotations (Pearson r = 0.719 versus 0.658). Notably, our framework maintains consistent performance across external validation datasets, performing reliably even on uncontrolled smartphone images without requiring additional training or fine-tuning. By pairing a protocol that requires no lesion-location information with area-based severity estimation, this work provides a foundation for objective acne assessment outside the clinic. Our implementation is publicly available at: \url{https://github.com/sukjuoh/VL-AcneSeg}
\end{abstract}


\begin{figure}[t]
    \centering
    {\small \textbf{Input Prompt:} \textit{``acne lesion on the left cheek''}\par\medskip}

    \setlength{\tabcolsep}{1pt} 
    \begin{tabular}{ccc}
        \includegraphics[width=0.31\columnwidth]{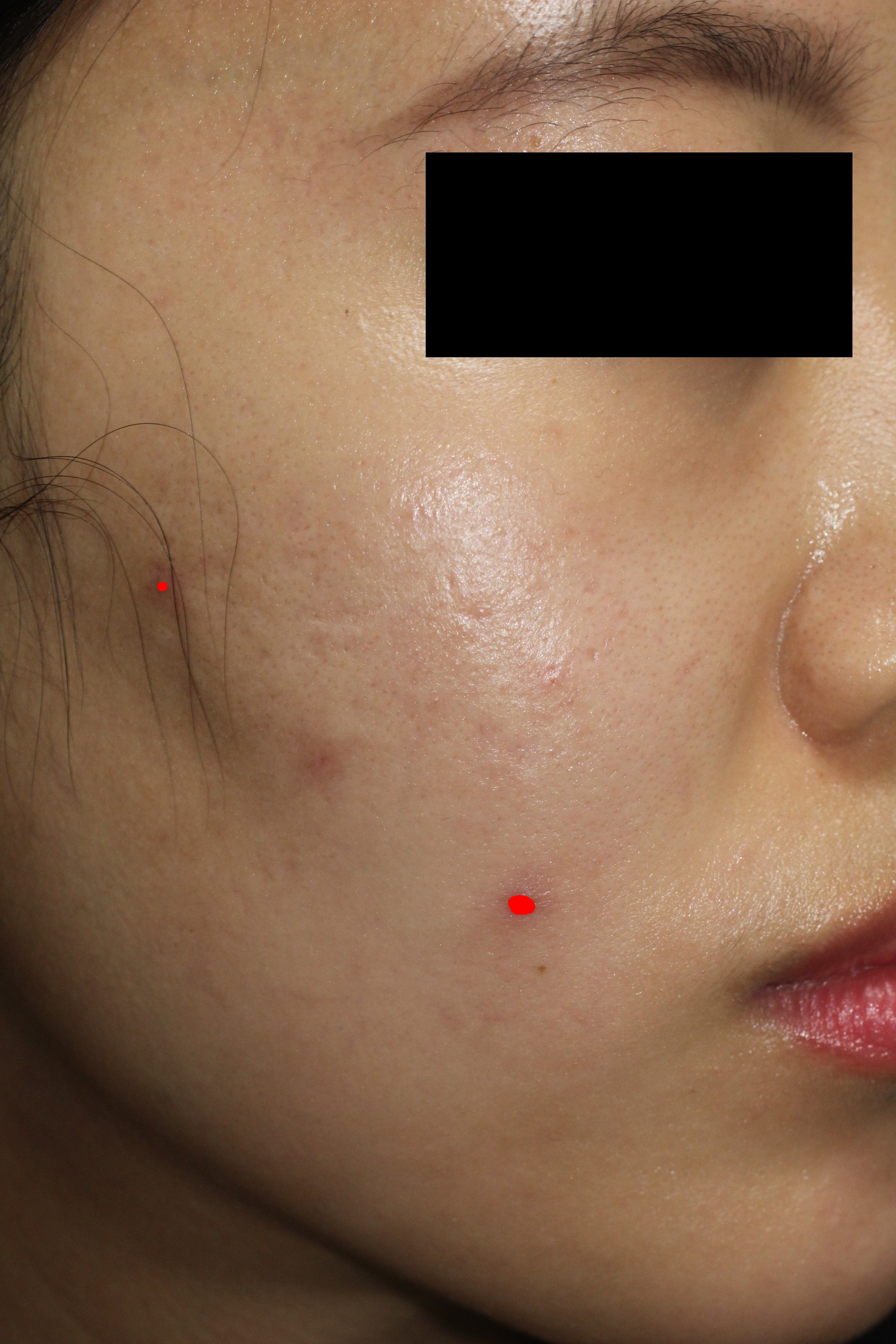}%
        & \includegraphics[width=0.31\columnwidth]{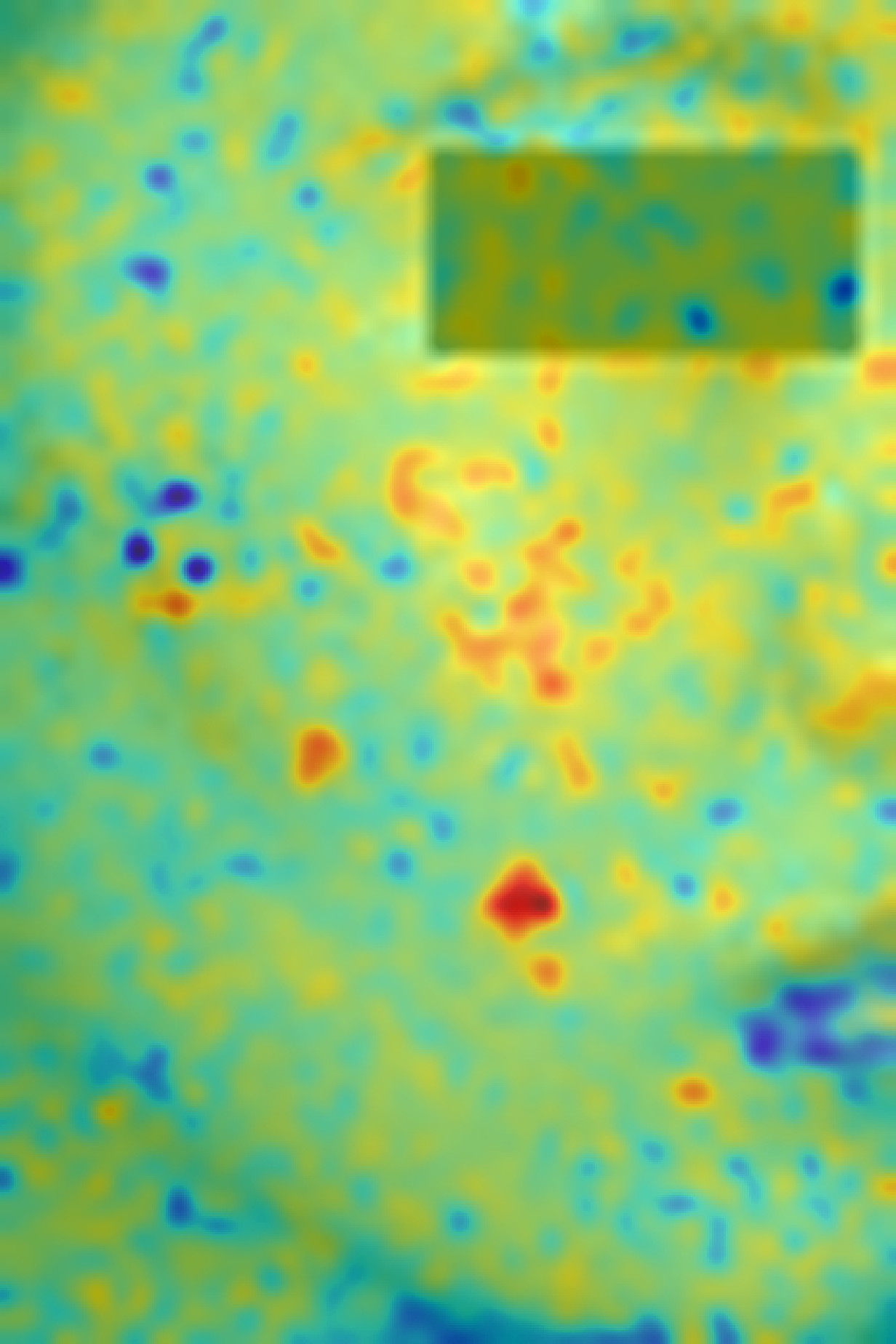}%
        & \includegraphics[width=0.31\columnwidth]{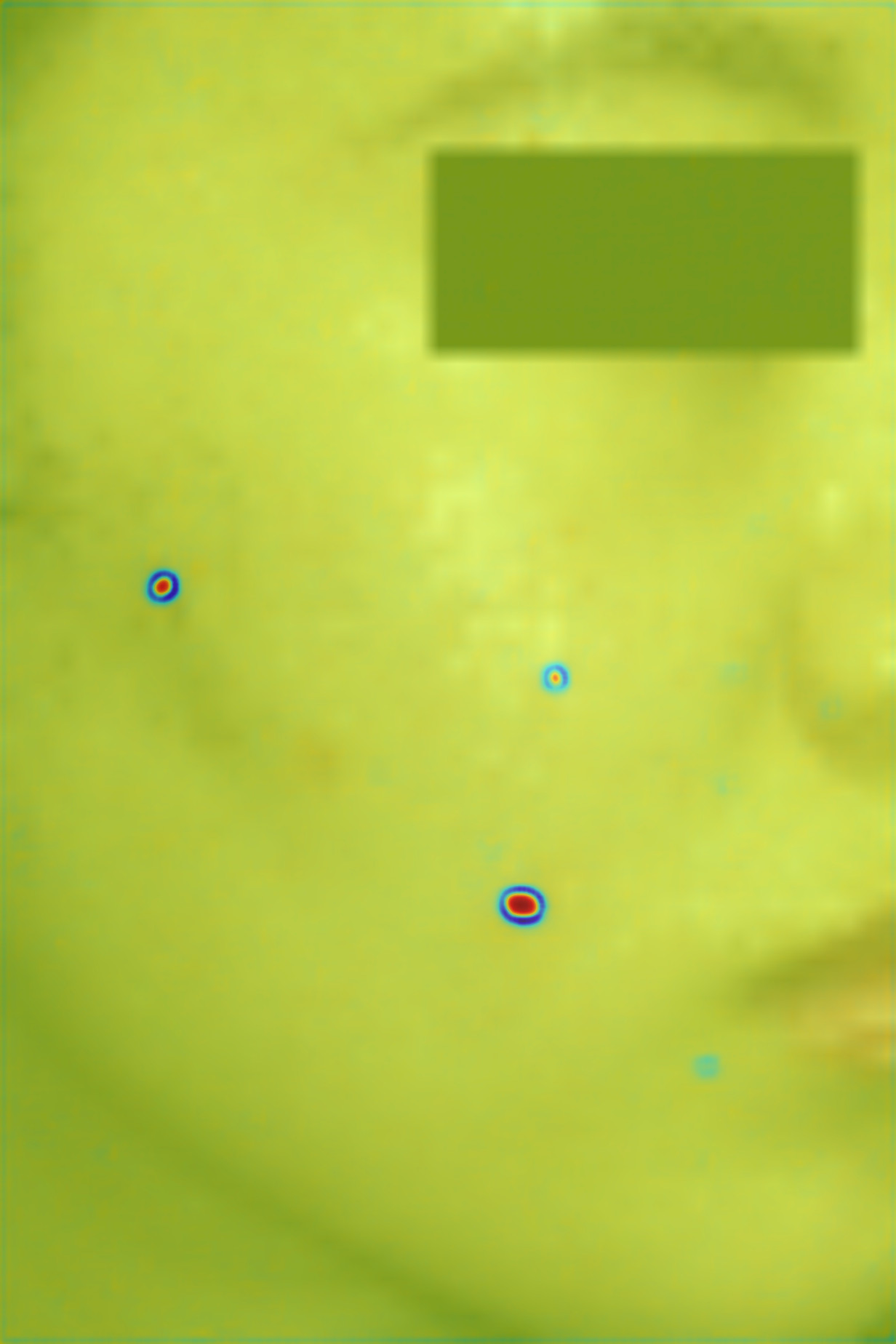} \\
        \small (a) Ground Truth & \small (b) Localization & \small (c) Refinement
    \end{tabular}

    \caption{\textbf{Conceptual framework of VL-AcneSeg.} Given a regional text prompt, VL-AcneSeg localizes potential lesion sites through CLIP-based semantic alignment (Middle) and then performs refinement to produce the segmentation mask (Right).}
    \label{fig:fig1}
\end{figure}

\section{Introduction}
Acne vulgaris is one of the most prevalent chronic skin disorders worldwide \cite{grada2022trends}, ranking among the top three most common skin conditions and affecting up to 85\% of adolescents and young adults during their lifetime \cite{tan2015global, law2010acne, bhate2013epidemiology}. Because it predominantly involves the facial region, where lesions are highly visible, acne can lead to considerable psychosocial stress \cite{park2016cross}. Acne lesions are broadly categorized into non-inflammatory types (e.g., comedones), which can be removed through simple extraction, and inflammatory types (e.g., papules, pustules, nodules), which often require pharmacological treatment depending on severity. Since inflammatory lesions are more directly tied to disease progression and therapeutic decisions, developing automated tools that can precisely quantify them is a critical priority for objective acne management.

Severity is assessed in clinical practice by global grading or by lesion counting, and each has its own limitation. Global grading compares a patient's presentation against standardized reference cases \cite{cho2021analysis} and carries the inter- and intra-observer variability that entails \cite{beylot2010inter, agnew2016comprehensive}. Counting individual lesions by type and region is more quantitative \cite{bae2024comprehensive, tan2006reliability} and has been automated by recent detection methods \cite{huynh2022automatic, wang2023novel, kim2023automated}, but it records a lesion as present or absent, so a small papule and a large confluent nodule contribute equally \cite{thiboutot2021parametric, gold2022picture}.

Area-based assessment is a promising alternative because it reflects lesion extent as well as lesion number. Gazeau \textit{et al.} combined lesion areas with lesion-specific severity scores into an overall index, offering a more objective basis for severity assessment \cite{gazeau2024acneai}.

Despite this potential, acne segmentation itself has received limited attention, largely due to the scarcity of annotated data and the inherent difficulty of the task \cite{moncho2023segmentation, traini2025artificial}. Pixel-level annotation of numerous small lesions requires dermatological expertise and is both costly and time-consuming. In addition, variations in skin tone, illumination, and imaging conditions introduce substantial appearance variability, while lesions often have blurred boundaries, irregular shapes, and subtle contrasts with surrounding skin, with confounding factors such as moles, scars, and pigmentation further complicating discrimination.

Reflecting these challenges, previous studies on acne segmentation have typically relied on general-purpose architectures and conventional training strategies, often applied to cropped patches around lesions rather than full-face images \cite{yadav2022hsv, junayed2022transformer, kim2024semi}. While such approaches simplify the task, they limit the ability to capture the global distribution of lesions, which is essential for area-based severity assessment. Even full-face models \cite{gazeau2024acneai, kim2023facial} have introduced little segmentation methodology tailored to the specific challenges of acne.

Vision-language models offer a way forward, since they allow simple textual cues---such as the approximate facial region in which lesions are present---to be supplied as spatial priors for segmentation. Existing vision-language segmentation methods, however, are built around a different question: they use text to specify \textit{which} class to segment, and their alignment relies on the contrast between classes, which is absent when there is a single lesion class. What acne does offer is a face whose anatomy is stable and nameable, so text can instead specify \textit{where} the target may appear. Such region-level descriptions carry only coarse spatial information at inference time, yet can narrow where the model searches for lesions across the face. Accordingly, we propose \textbf{VL-AcneSeg}, a multimodal framework that utilizes region-level text prompts to guide the segmentation of inflammatory lesions. As illustrated in \textcolor{IEEEBLUE}{Fig.~\ref{fig:fig1}}, the model aligns these textual prompts with visual features via CLIP to identify potential lesion sites, which are then refined into precise segmentation masks. Under region-level prompting the design additionally yields a severity estimate for each facial region. Our contributions are as follows:
\begin{itemize}
    \item We recast text-guided segmentation for a setting in which open-vocabulary alignment does not apply. Existing paradigms use text to specify \textit{which} class to segment, which carries little information when there is a single lesion class; we instead use it to specify \textit{where} the target may appear, so that the text channel supplies a spatial rather than a categorical constraint.

    \item We adapt the patch-wise similarity paradigm to acne through tiled high-resolution CLIP encoding, layer-targeted V-V attention and CLS-token-guided cross-attention, so that the alignment operates at the resolution the lesions demand.

    \item We report a deployment-realistic protocol that requires no reference-derived information, evaluate it on external clinical and smartphone datasets without retraining, and show that the resulting lesion areas track Investigator's Global Assessment severity at the level of the reference annotations.
\end{itemize}
\section{Related Work}

\subsection{Acne Lesion Segmentation}
Acne lesion segmentation has been explored only to a limited extent compared to other dermatological conditions. Existing studies often rely on cropped image patches rather than full-face images, limiting their ability to capture the global lesion distribution \cite{yadav2022hsv, junayed2022transformer, kim2024semi}, and typically adopt standard architectures such as U-Net without acne-specific design \cite{kim2023facial}. Gazeau \textit{et al.} \cite{gazeau2024acneai} employed a conventional segmentation backbone within their area-based severity framework, but their focus was on the evaluation methodology rather than the segmentation model itself. Moreover, absolute segmentation metrics reported in this domain remain modest across studies, reflecting the intrinsic difficulty of distinguishing acne from visually similar skin features such as PIH, scars, and pores~\cite{kim2023facial, gazeau2024acneai}. A systematic review covering 2017--2025 screened 345 articles and identified 29 eligible studies, the majority addressing severity grading or lesion counting rather than pixel-level segmentation \cite{traini2025artificial}. Pixel-level acne segmentation has therefore received limited attention, and the methods that do address it were not designed for the setting that area-based assessment requires: the whole face at once, with lesions that are numerous and easily confused with the marks they leave behind.

\subsection{Open-Vocabulary and Text-Guided Segmentation}
Vision-language models such as CLIP \cite{radford2021learning} and ALIGN \cite{jia2021scaling} learn joint image--text representations that support open-vocabulary understanding and zero-shot transfer \cite{zhou2022conditional, zhong2022regionclip, jeong2023winclip}. This has motivated their use in dense prediction, through transformer-based fusion \cite{ding2022vlt, yang2022lavt, wang2022cris} and open-vocabulary paradigms such as OVSeg \cite{liang2023open} and ODISE \cite{xu2023open}, which match class-agnostic region proposals with text embeddings.

However, these paradigms face significant practical and logical hurdles in acne segmentation. For ROI-based methods, if a mask generator can successfully propose an ROI for a specialized single-class lesion, the subsequent text-alignment step becomes effectively redundant. Furthermore, the visual features within an acne lesion are often too localized and subtle to provide sufficient discriminative information for meaningful alignment, while random masking strategies like MaskCLIP \cite{zhou2022extract} fail to maintain the structural integrity of such small-scale lesions during training. Moreover, the lack of distinct positive-negative pairs in localized skin imagery makes it difficult to establish a viable contrastive learning framework.

More recently, CAT-Seg \cite{cho2024cat} introduced a patch-wise cosine similarity paradigm that formulates segmentation as a direct alignment between CLIP’s visual and textual embeddings, with subsequent extensions such as SED \cite{xie2024sed} and ESC-Net \cite{lee2025effective} refining this formulation for natural-image open-vocabulary settings; we compare against SED directly, whereas ESC-Net couples the cost volume with SAM, which enters our comparison as a separate fine-tuned baseline. The most fundamental distinction between our approach and CAT-Seg lies in the role of text: CAT-Seg consumes class-level prompts that specify what object to segment, whereas VL-AcneSeg consumes region-level prompts that specify where the target may appear. Building on this distinction, our framework repurposes the alignment itself: the text channel supplies a spatial constraint rather than a categorical one, which extends the formulation to dense single-class segmentation, a setting where prior open-vocabulary methods struggle for want of inter-class contrast. The architectural components that follow are adaptations required to make this alignment operate at the resolution the task demands. A complementary line of work prompts CLIP with visual rather than textual references: PBIP \cite{tang2025prototype} constructs class prototypes from a training image bank and uses them as image prompts for weakly supervised histopathological segmentation. Such prototype prompts encode appearance, whereas the region-level prompts used here encode anatomical location, and the two forms of guidance are therefore complementary.

\subsection{Text-Guided Segmentation Model in the Medical Domain}
In the medical domain, text has mainly been used to identify the target structure. LViT \cite{li2023lvit} integrates BERT \cite{devlin2019bert} text embeddings into a hybrid CNN-Transformer framework, while VLM-based approaches such as MedCLIP-SAM \cite{koleilat2024medclip}, SegICL \cite{shen2024segicl}, and OMT-SAM \cite{zhang2025organ} combine pretrained text encoders with segmentation frameworks for zero-shot or prompt-based segmentation of anatomical structures.

More recent studies have begun to exploit the spatial content of clinical text explicitly. TVE-Net \cite{fang2025driven} converts the location descriptions in radiology reports into a ``text view'' through a hand-designed positional probability function, assigning a lesion probability to each image region so that textual spatial information is injected into the segmentation network. STPNet \cite{shan2025stpnet} instead retrieves multi-scale textual descriptions from a curated medical text repository during training and thereby removes the need for text input at inference, while EviVLM \cite{pan2025evivlm} introduces evidential learning to quantify and mitigate the modality gap between image and text representations.

Our framework shares with TVE-Net the premise that textual location information can serve as a spatial prior, but realises it differently. TVE-Net maps free-text radiology reports to a probability map through a hand-designed positional function, whereas VL-AcneSeg keeps the prior inside the joint embedding space: fixed-template region prompts are encoded by the CLIP text encoder and aligned with patch embeddings through cosine similarity, so that the prior is a learned visual--textual correspondence rather than a predefined geometric mapping, and no narrative report is required. These advances have moreover addressed radiology, where lesions are comparatively large and few in number; to our knowledge text-guided segmentation has not been applied to acne, whose lesions are small, numerous and distributed across the whole face.

\begin{figure}[t]
    \centering
    \includegraphics[width=0.8\columnwidth]{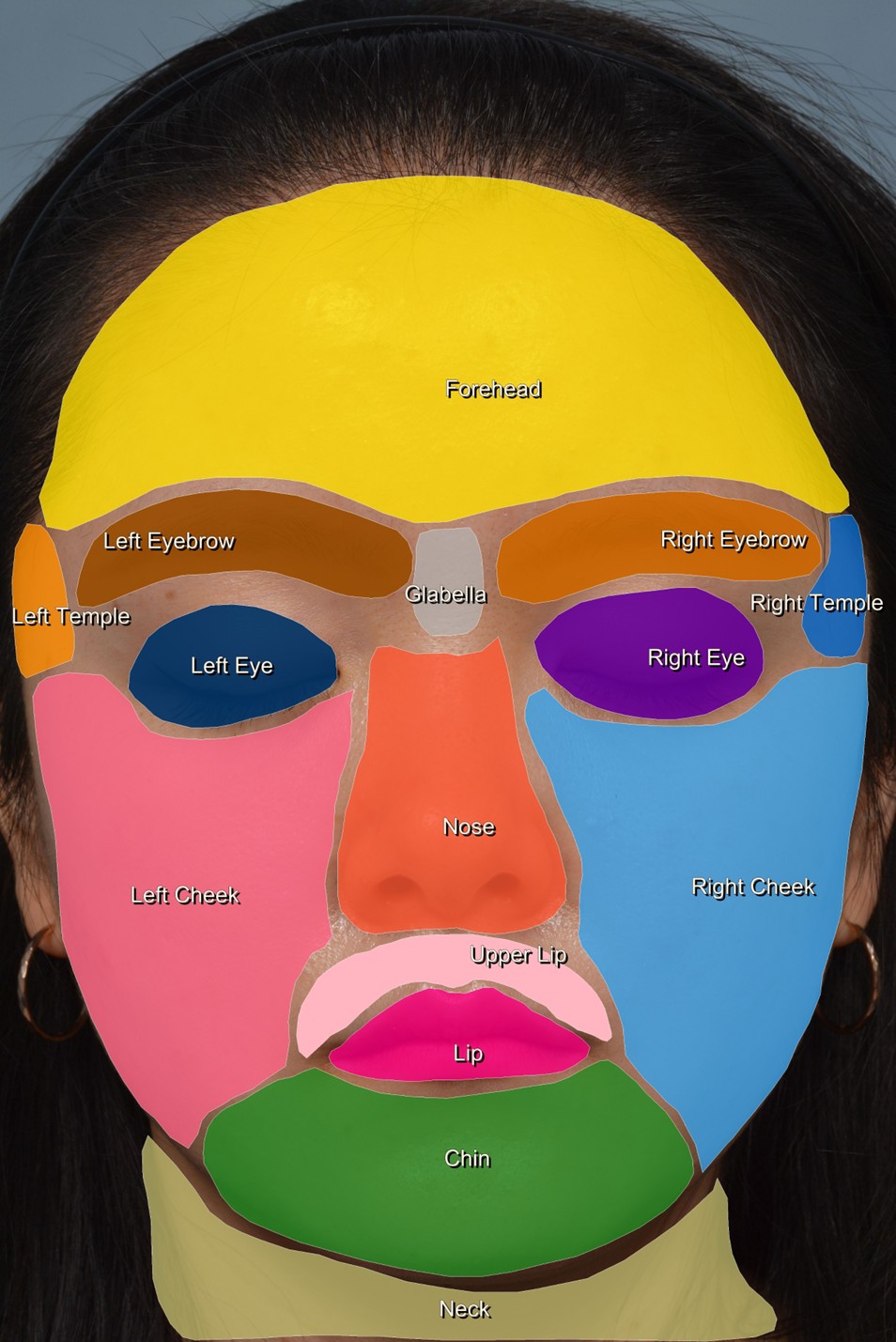}
    \caption{Facial regions were defined using Mediapipe Face Mesh \cite{lugaresi2019mediapipe}, and lesion masks were assigned to each region according to their centroid location. Region-wise masks and text prompts (``acne lesion on the \{\textit{region}\}'') were then generated to provide spatial guidance for segmentation.}
    \label{fig:fig2}
\end{figure}

\begin{figure*}[t]
    \centering
    \includegraphics[width=0.9\textwidth]{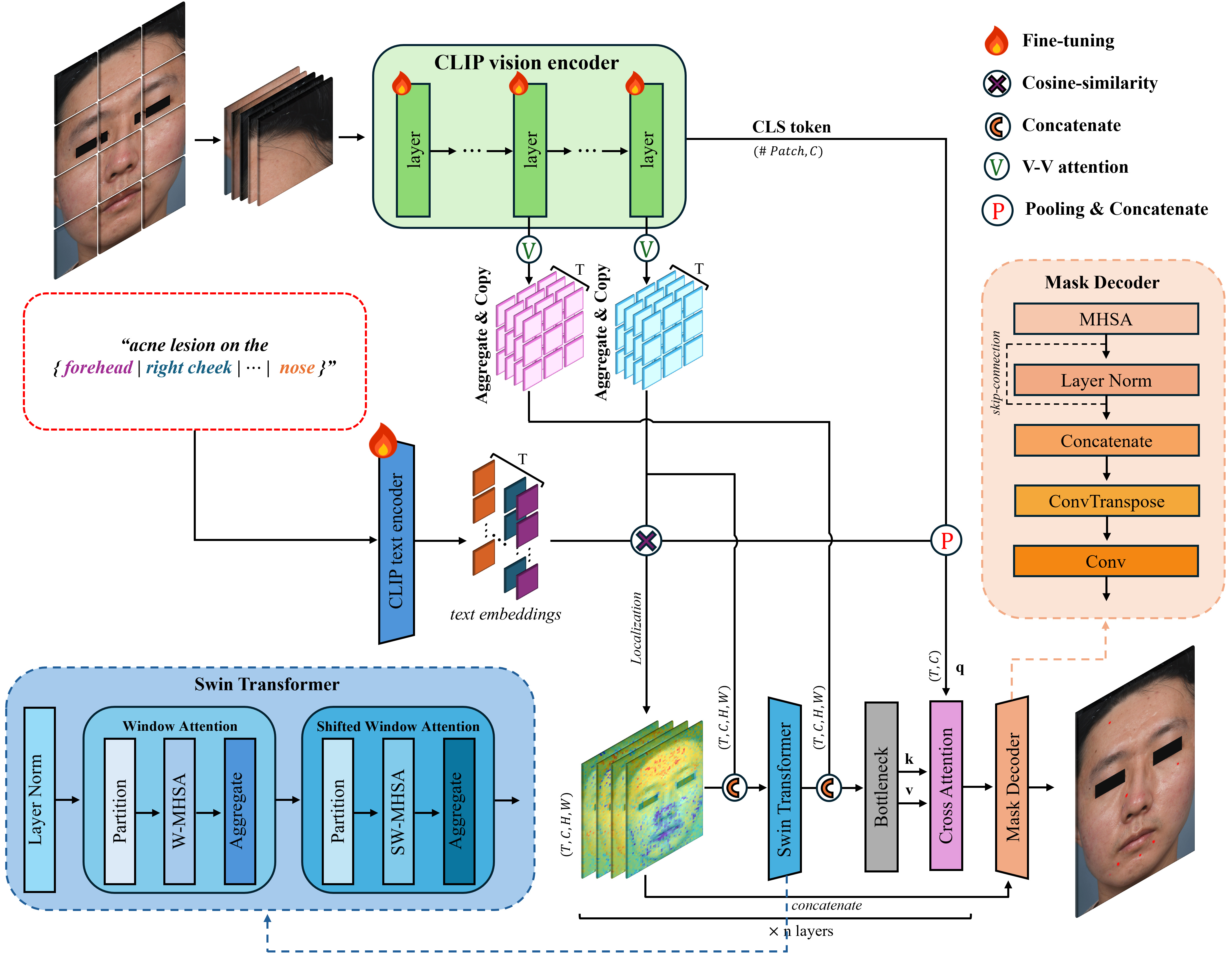} 
    \caption{\textbf{Overview of VL-AcneSeg.} Overlapping tiles are encoded by CLIP and reassembled into a unified feature map, whose patch-wise cosine similarity with the text embeddings yields the localization map; intermediate vision features and the CLS token provide additional guidance. A Swin Transformer, a CNN bottleneck and a decoder then refine this map into the lesion mask.}
    \label{fig:fig3}
\end{figure*}

\section{Method}
\subsection{Data Preparation}
\label{sec:data_prep}
We collected standardized clinical photographs from 258 patients (38.4\% male; mean age 22.7 $\pm$ 5.97 years) who visited the Department of Dermatology, Seoul National University Hospital, using Canon EOS 550D and Nikon D7100 cameras. All participants were of Korean ethnicity, and Fitzpatrick phototype was not routinely recorded. The same cohort was described previously in the context of automated lesion detection and counting \cite{kim2023automated}; the present study is the first to derive pixel-level segmentation masks from these data. A total of 1{,}213 images carry 20{,}699 annotated acne lesions. The lesion inventory originates from the study of this cohort reported in \cite{kim2023automated}, where two dermatology residents independently marked each lesion. For the present study these annotations were converted into pixel-level masks with LabelMe \cite{russell2008labelme}, and a board-certified dermatologist reviewed every image, corrected the delineations and adjudicated the final set. The reference standard is therefore fixed by a single senior reader rather than by agreement among readers. The annotations cover five lesion types (closed and open comedones, papules, nodules/cysts, and pustules). Because clinical acne severity assessment is based primarily on inflammatory lesions, and because comedones are frequently indistinguishable at the pixel level in standardized clinical photographs, this study targets inflammatory lesions only (papules, nodules/cysts, and pustules); patients without any inflammatory lesion were excluded, leaving 222 patients, 821 images and 4{,}655 annotated lesion instances. 

To generate region masks and text prompts, each facial image was divided into anatomical regions (forehead, cheeks, nose, and chin) using MediaPipe Face Mesh landmarks \cite{lugaresi2019mediapipe}. As illustrated in \textcolor{IEEEBLUE}{Fig.~\ref{fig:fig2}}, the center coordinate of each lesion mask was used to determine its corresponding facial region, and the mask was assigned accordingly.

For every region containing one or more lesions, an individual text prompt was generated following a fixed template format, ``acne lesion on the \{\textit{region}\}.'' For example, if acne lesions were present on the forehead, chin, and neck, three separate prompts (``acne lesion on the forehead,'' ``acne lesion on the chin,'' and ``acne lesion on the neck'') were produced. These prompts indicate the approximate location of lesions across facial regions.

\subsection{Model Overview}
Our goal is to segment acne lesions from facial images by leveraging textual descriptions of facial regions as spatial priors. Given an input image and corresponding region-level text prompts, our model outputs a pixel-wise segmentation mask indicating the locations of acne lesions.

The overall architecture is illustrated in \textcolor{IEEEBLUE}{Fig.~\ref{fig:fig3}}. The input image is divided into tiles, encoded by the CLIP vision encoder and reassembled into a visual feature map, whose patch-wise cosine similarity with the CLIP text embeddings yields a localization map that is then passed to a Swin Transformer.

Intermediate visual features and the CLS tokens provide additional guidance, and a cross-attention module conditioned on the text embeddings refines the representation before the mask decoder, so that region-level cues are combined with visual features in a patch-aligned manner.

\subsection{Localization: High-Resolution Features and Patch-wise Similarity}

Acne lesions are typically small and have blurry boundaries, making high-resolution input crucial for precise segmentation. However, the CLIP vision encoder $\Phi$ (ViT-L/14@336px) does not support inputs larger than $336 \times 336$. To overcome this limitation, we divide each input image $X \in \mathbb{R}^{3 \times H \times W}$ into overlapping tiles of size $k \times k$ ($k = 336$) and process each tile individually through $\Phi$. The tiling layout follows from two constraints rather than from a search: the tile size is fixed at $336$ by the encoder, and clinical facial photographs have a $3{:}4$ width-to-height ratio, so tiling the image with square patches requires four rows and three columns. With the longer side resampled to $1024$, the smallest such layout is the $4 \times 3$ configuration used here; coarser layouts reduce the pixel density per tile, while the next admissible refinement quadruples the number of CLIP forward passes for tiles that each cover less anatomical context. For each tile $P_{i,j}$, we obtain the CLS token and patch embeddings:
\begin{equation}
[z_{i,j}^{\text{cls}}, Z_{i,j}] = \Phi(P_{i,j}),
\end{equation}
where $Z_{i,j} \in \mathbb{R}^{c \times h \times w}$ is the token grid for the $(i,j)$-th tile ($h=w=24, \;c=1024$). All tile embeddings are then reassembled into a single feature map $S$ by placing them at their corresponding spatial locations and averaging the overlapping regions using a Gaussian weight window $G$:
\begin{equation}
S = \frac{\sum_{i,j} Z_{i,j} \odot G_{i,j}}{\sum_{i,j} G_{i,j} + \epsilon},
\end{equation}
where $\odot$ denotes element-wise multiplication, $G_{i,j}$ is a 2D Gaussian kernel of size $h \times w$ centered on tile $(i,j)$ with standard deviation $\sigma = h/4$ that downweights border patches to reduce stitching artifacts at tile boundaries, and $\epsilon = 10^{-6}$ is a small constant ensuring numerical stability. Finally, the stitched feature map $S$ is upsampled to a fixed resolution and normalized at each spatial location:
\begin{equation}
S_{\text{norm}}(x, y) = \frac{S(x, y)}{\| S(x, y) \|_2}.
\end{equation}

Lesions are localized by aligning visual features with the text prompts (\textcolor{IEEEBLUE}{Fig.~\ref{fig:fig1}}~(b)). The $N_T$ region prompts $\{T_k\}_{k=1}^{N_T}$ are passed through the CLIP text encoder $\Gamma$:
\begin{equation}
\{ t_k = \Gamma(T_k) \in \mathbb{R}^C \}_{k=1}^{N_T}.
\end{equation}
Let $S \in \mathbb{R}^{C \times H \times W}$ denote the visual feature map extracted from the image encoder $\Phi$. Before computing similarity, both the visual features $S$ and text embeddings $t_k$ are normalized along the channel dimension. For each text embedding $t_k$, we compute a patch-wise cosine similarity map with the visual features as:
\begin{equation}
M_k(h, w) = \frac{S_{h,w}^\top t_k}{\| S_{h,w} \|_2 \| t_k \|_2},
\end{equation}
yielding a set of similarity maps:
\begin{equation}
M = \{M_k\}_{k=1}^{N_T} \in \mathbb{R}^{H \times W \times N_T},
\end{equation}
each encoding the alignment between visual patches and one region-level description.

\subsection{Feature Enhancement}
CLIP \cite{radford2021learning} aligns visual and textual representations through a contrastive objective, but its attention is biased toward large and salient regions and is often dominated by a few tokens, which suppresses the fine-grained local features on which small lesions depend---a limitation also noted in AnomalyCLIP \cite{zhou2024anomalyclip}.

To address this issue, we adopt the V-V attention mechanism introduced by AnomalyCLIP, but redesign its deployment strategy for fine-grained lesion segmentation. Whereas AnomalyCLIP applies V-V attention across multiple intermediate layers for global anomaly detection, we apply it only at the final feature extraction layer. The reason is that V-V attention replaces query--key routing with value--value affinity, so each token is re-expressed as a weighted average of tokens with similar values. Applied once, this suppresses the few dominant tokens that otherwise absorb the attention mass; applied repeatedly, the same averaging acts as a low-pass filter over the token grid, and the patch-aligned spatial structure that dense prediction depends on is progressively smoothed away. Anomaly detection tolerates this because it requires only a coarse image-level score, whereas lesions a few pixels across do not. \textcolor{IEEEBLUE}{Table~\ref{tab:table2}} confirms the distinction empirically. Specifically, let  $V \in \mathbb{R}^{N \times d}$
 denote the value embeddings
extracted from this target layer of the CLIP vision 
encoder, where $N$ is the number of tokens and $d$ is the 
embedding dimension. We compute a token-token affinity 
matrix through a scaled dot product between the value vectors:
\begin{equation}
A_{vv} = \text{Softmax}\left(\frac{VV^\top}{\sqrt{d}}\right),
\end{equation}
and use it to refine the visual features:
\begin{equation}
V' = A_{vv}V.
\end{equation}
The refined features are then passed through the original output projection of that layer and added to the original layer output to form the final feature map.

Fine-grained local information diminishes with layer depth, so we additionally take the feature map of the 12th CLIP layer as guidance and fuse it before the bottleneck (\textcolor{IEEEBLUE}{Section~\ref{sec:refine}}).

We further utilize CLS tokens to inject coarse spatial 
information derived from the image tiles. Since each image is 
divided into $4 \times 3$ tiles and individually processed by the CLIP 
vision encoder, we obtain one CLS token $c_n \in \mathbb{R}^C$
for each tile 
$n=1,\dots,N$. Given the normalized text embeddings $\{t_k\}_{k=1}^{N_T}$
, we compute the cosine similarity between all CLS tokens and 
text embeddings to produce tile-text similarity weights:
\begin{equation}
w_{n,k} = \frac{\exp (c_n^\top t_k)}{\sum_{n'=1}^N \exp (c_{n'}^\top t_k)}.
\end{equation}
These weights are then used to compute a weighted combination of CLS tokens for each text embedding, representing the pooling operation (P) illustrated in \textcolor{IEEEBLUE}{Fig.~\ref{fig:fig3}}:
\begin{equation}
\bar{c}_k = \sum_{n=1}^N w_{n,k} c_n.
\end{equation}
The resulting $\bar{c_k}$ encodes spatial cues indicating which tiles 
are semantically aligned with each text prompt. Finally, $\bar{c_k}$ is 
concatenated with the corresponding text embedding $t_k$, and 
this combined representation is used as the query for the cross-attention module.

\subsection{Refinement and Mask Decoding}
\label{sec:refine}
The patch-wise similarity maps $M \in \mathbb{R}^{B \times N_T \times H \times W}$ are first passed through a convolution layer to introduce a channel dimension, resulting in $M' = \text{Conv}(M) \in \mathbb{R}^{(B \times N_T) \times C_M \times H \times W}$. In parallel, the original CLIP vision encoder feature map $S$ is replicated and concatenated with $M'$ to form the input feature $F_{\text{in}}$. This feature map is processed by a Swin Transformer with W-MHSA and SW-MHSA blocks to capture long-range spatial dependencies (\textcolor{IEEEBLUE}{Fig.~\ref{fig:fig3}}).

The guidance feature $F_{\text{guide}}$, taken from the 12th layer of the CLIP vision encoder, is passed through a convolution layer, replicated $N_T$ times and concatenated with $F_{\text{swin}}$ before the bottleneck:
\begin{equation}
F_{\text{bottle\_in}} = \text{Concat}[F_{\text{swin}}, \text{Repeat}_{N_T}(F_{\text{guide}})],
\end{equation}
where $F_{\text{guide}} \in \mathbb{R}^{B \times C_g \times H' \times W'}$. The bottleneck refines this representation to integrate spatial similarity cues with low-level visual features:
\begin{equation}
F_{\text{bottle}} = \text{Bottleneck}(F_{\text{bottle\_in}}) \in \mathbb{R}^{(B \times N_T) \times C_b \times H' \times W'}.
\end{equation}

The weighted CLS token $\bar{c}_k$ and the text embedding $t_k$ form the query $Q$:
\begin{equation}
Q = \{\text{Concat}[\bar{c}_k, t_k]\}_{k=1}^{N_T} \in \mathbb{R}^{N_T \times C_q}.
\end{equation}
which is broadcast to $Q' \in \mathbb{R}^{(B \times N_T) \times C_q \times H' \times W'}$ and interacts with the refined visual features through cross-attention:
\begin{equation}
F_{\text{fused}} = \text{CrossAttn}(Q', F_{\text{bottle}}, F_{\text{bottle}}).
\end{equation}

The mask decoder produces the segmentation mask $\hat{Y}$: the fused feature $F_{\text{fused}}$ is processed through Multi-Head Self-Attention (MHSA) and Layer Normalization (LN), then concatenated with the similarity map $M'$ to anchor the predictions to the initial localization results. The final mask is produced through learned upsampling via a Convtranspose layer followed by a convolution:
\begin{equation}
\begin{split}
\hat{Y} = \text{Conv}( & \text{ConvTranspose}( \\
& \text{Concat}(\text{LN}(\text{MHSA}(F_{\text{fused}})), M'))).
\end{split}
\end{equation}

\subsection{Loss Function}
We supervise the network with two complementary losses: a global segmentation loss over the full-face mask, and a region-specific loss for each regional prediction.

For the global segmentation, we adopt the weighted IoU and weighted BCE losses from PraNet \cite{fan2020pranet}, which emphasize hard pixels by upweighting challenging regions:
\begin{equation}
\mathcal{L}_{\text{total}} = \mathcal{L}_{\text{wIoU}}(\hat{Y}, Y) + \mathcal{L}_{\text{wBCE}}(\hat{Y}, Y).
\end{equation}

For region-specific supervision, we apply focal loss \cite{lin2017focal} (with $\alpha=0.75$) to each regional prediction $\hat{Y}_k$ associated with a text prompt $T_k$:
\begin{equation}
\mathcal{L}_{\text{region}} = \frac{1}{N_T} \sum_{k=1}^{N_T} \mathcal{L}_{\text{focal}}(\hat{Y}_k, Y_k),
\end{equation}
where $Y_k$ is the ground-truth mask for the $k$-th facial region and $N_T$ is the number of region-level prompts. This term supervises the similarity map $M$ directly, encouraging alignment between visual patches and their regional prompts, while the focal weighting emphasizes small or ambiguous lesions.

The final objective combines both terms:
\begin{equation}
\mathcal{L} = 0.6\mathcal{L}_{\text{total}} + 0.4\mathcal{L}_{\text{region}}.
\end{equation}

\section{Experiments}
\subsection{Setup}
\label{sec:dataset}
We evaluated our model on three datasets. The first is the internal test set described in \textcolor{IEEEBLUE}{Section~\ref{sec:data_prep}}. The internal dataset was split into training, validation, and test sets at an 8:1:1 ratio on a patient-wise basis to prevent data leakage across multiple images from the same subject; the split was defined on the full cohort of 258 patients, and after excluding patients without inflammatory lesions the realized split comprised 190, 16 and 16 patients. Investigator's Global Assessment (IGA) scores were not recorded at acquisition and were assigned retrospectively at the visit level by three board-certified dermatologists, for the internal test set only, since IGA is used solely for the clinical correlation analysis (\textcolor{IEEEBLUE}{Section~\ref{sec:iga}}) and never for model training or selection. The 58 internal test images correspond to 22 visits from 16 patients, of which 3, 7, 8 and 4 were graded 1 to 4 on the standard five-point scale and none as grade 0; the grading protocol is given in Section~\ref{sec:supp_data} of the Supplementary Material. The other two datasets were derived from the AI Hub Korean Skin Condition Measurement Dataset\footnote{https://www.aihub.or.kr}, developed to support facial skin analysis for Korean individuals.

\begin{table}[t]
\centering
\caption{Composition of the internal splits and the two external test sets, counted over the inflammatory lesion masks used in this study. The full annotated cohort comprises 258 patients, 1{,}213 images and 20{,}699 lesions across five types. The internal dataset was divided patient-wise, and each external image originates from a distinct subject.}
\label{tab:splits}
\footnotesize
\begin{tabular}{lccc}
\hline
\textbf{Split} & \textbf{Patients} & \textbf{Images} & \textbf{Lesions} \\
\hline
\multicolumn{4}{l}{\textit{Internal (Seoul National University Hospital)}} \\
Train & 190 & 703 & 3{,}814 \\
Validation & 16 & 60 & 494 \\
Test & 16 & 58 & 347 \\
Total & 222 & 821 & 4{,}655 \\
\hline
\multicolumn{4}{l}{\textit{External (AI Hub)}} \\
Controlled & 71 & 71 & 171 \\
Real-world & 56 & 56 & 101 \\
\hline
\end{tabular}
\end{table}

From this collection we constructed two external test sets, one for a controlled imaging setting and one for an uncontrolled, real-world setting. Because the collection contains many images without visible inflammatory lesions, candidates were screened with an acne detection model trained on the public ACNE04 dataset~\cite{wu2019joint} and then sampled at random; the dermatologist who annotated the internal set reviewed the sampled images and excluded those without confirmed inflammatory acne, yielding 71 controlled-setting and 56 real-world images. The screening model predicts bounding boxes rather than masks and its outputs were used only for candidate selection, never in the annotations or in any evaluation. Pixel-level masks were newly constructed for the present study by the same board-certified dermatologist who adjudicated the internal set, so both sets share one reference reader; unlike the internal set, no prior box inventory was available here, so the lesions were located as well as delineated. Text prompts were generated as described in \textcolor{IEEEBLUE}{Section~\ref{sec:data_prep}}, and the composition of every split is summarized in \textcolor{IEEEBLUE}{Table~\ref{tab:splits}}. The two subsets differ from the internal cohort in lesion burden and probe different kinds of imaging shift: the controlled subset was acquired with digital cameras under a setup comparable to the internal cohort, whereas the real-world subset was acquired with smartphones. The screening procedure, the resulting sampling bias and the available metadata are detailed in Section~\ref{sec:supp_data} of the Supplementary Material.

\subsection{Evaluation Metrics}
We used the Dice coefficient and Intersection over Union (IoU):

\begin{equation}
\text{Dice} = \frac{2|Y \cap \hat{Y}|}{|Y| + |\hat{Y}|},
\end{equation}
\begin{equation}
\text{IoU} = \frac{|Y \cap \hat{Y}|}{|Y \cup \hat{Y}|}.
\end{equation}

We further report pixel-wise Precision and Recall, together with lesion-level counts per image. A predicted connected component is counted as a true positive (TP/img) if it overlaps a reference lesion in at least one pixel, and as a false positive (FP/img) otherwise. Specificity is not reported: background pixels dominate facial images, so it exceeds 0.999 for every method and carries no information. Let $\Omega$ denote the set of valid image pixels, $Y$ the ground-truth lesion mask and $\hat{Y}$ the predicted mask. The pixel-wise counts are defined as $\mathrm{TP} = |Y \cap \hat{Y}|$, $\mathrm{FP} = |\hat{Y} \setminus Y|$, $\mathrm{FN} = |Y \setminus \hat{Y}|$, and $\mathrm{TN} = |\Omega| - |Y \cup \hat{Y}|$, giving

\begin{equation}
\text{Precision} = \frac{\mathrm{TP}}{\mathrm{TP} + \mathrm{FP}},
\end{equation}
\begin{equation}
\text{Recall} = \frac{\mathrm{TP}}{\mathrm{TP} + \mathrm{FN}}.
\end{equation}

All metrics are aggregated over the entire test set by accumulating $\mathrm{TP}$, $\mathrm{FP}$, $\mathrm{FN}$, and $\mathrm{TN}$ across images before computing the ratios (micro-averaging). Lesion-level recall, precision and F1 restricted to the smallest quartile of lesions (below 500\,px$^2$), together with a threshold-free comparison based on the area under the precision--recall curve, are reported for all methods in Table~\ref{tab:supp_metrics} of the Supplementary Material. Zero-padded regions (\textcolor{IEEEBLUE}{Section~\ref{sec:impl}}) are excluded from $\Omega$, so the denominator is the same effective area for every method.

\subsection{Implementation and Training Protocol}
\label{sec:impl}
The model was implemented in PyTorch and trained on an NVIDIA A6000 GPU (48 GB). We used the AdamW \cite{loshchilov2017decoupled} optimizer with a learning rate of $2 \times 10^{-4}$ for the main network and $2 \times 10^{-6}$ for the CLIP encoders, for 40 epochs with a batch size of 4. The optimizer’s betas were set to 
(0.9, 0.999).

Baselines used the default hyperparameters of their original papers with no additional search, and were initialized from the pretrained weights those implementations provide---ViT-B/16 for TransUNet, Swin Transformer for Swin-UNet, MiT for SegFormer, ImageNet ResNet encoders for PSPNet, DeepLabV3+ and CE-Net, VMamba for Swin-UMamba$\dagger$, and DeiT-Base~\cite{touvron2021training} with Bio\_ClinicalBERT~\cite{alsentzer2019publicly} for EviVLM---while U-Net, U-Net++ and nnU-Net were trained from random initialization. SAM was initialized from the pretrained ViT-L checkpoint, with its image encoder and mask decoder fully fine-tuned and the prompt encoder left unused, as no point or box prompts were supplied. Every model, including ours, used the identical patient-wise 8:1:1 split and no data augmentation. For each model we selected the checkpoint with the highest validation IoU, and fixed the binarization threshold at 0.5. CAT-Seg and SED share our input resolution and $4 \times 3$ tiling, so the three CLIP-based methods are compared at matched resolution. All others take the image resized to a height of 1024, zero-padded along the width where the architecture requires a fixed input shape, with padded regions excluded from the loss and the evaluation.

\subsection{Results}
\begin{table*}[t]
\renewcommand{\arraystretch}{1.3}
\setlength{\tabcolsep}{3pt}
\centering
\caption{Quantitative comparison of acne lesion segmentation on the internal and external datasets. Shaded rows ($\ast$) are vision-language methods given region-level prompts derived from the reference masks, an oracle setting; CAT-Seg and SED use the same tiling scheme and input resolution as the proposed method, so that the CLIP-based methods are compared at matched resolution. All remaining rows, including VL-AcneSeg (Global), require no reference-derived information. TP/img and FP/img are the numbers of predicted lesion components per image that do or do not overlap a reference lesion; neither is ranked. Point estimates are computed over all test-set pixels; 95\% confidence intervals are obtained by patient-level cluster bootstrap (1{,}000 resamples). Best values are in \textbf{bold} and second-best are \underline{underlined}.}
\label{tab:table1}
\resizebox{\textwidth}{!}{
\begin{tabular}{l|c|cccccc|cccccc|cccccc} 
\hline
\multirow{3}{*}{\textbf{Model}} & \multirow{3}{*}{\textbf{Backbone}} & \multicolumn{6}{c|}{\textbf{Internal Data}} & \multicolumn{12}{c}{\textbf{External Data (Evaluation Only)}} \\
\cline{9-20}
 & & \multicolumn{6}{c|}{\textbf{(Training \& Evaluation)}} & \multicolumn{6}{c|}{\textbf{Controlled Setting (1)}} & \multicolumn{6}{c}{\textbf{Real-world Setting (2)}} \\
\cline{3-20}
 & & IoU & Dice & Precision & Recall & TP/img & FP/img & IoU & Dice & Precision & Recall & TP/img & FP/img & IoU & Dice & Precision & Recall & TP/img & FP/img \\
\hline
\rowcolor{VLMRow} \textbf{VL-AcneSeg (Region)}$^\ast$ & ViT-L & \textbf{0.3602} {\scriptsize [0.309, 0.399]} & \textbf{0.5296} {\scriptsize [0.472, 0.570]} & 0.5202 & 0.5393 & 4.29 & 2.48 & \textbf{0.4232} {\scriptsize [0.376, 0.470]} & \textbf{0.5948} {\scriptsize [0.546, 0.640]} & 0.6110 & 0.5794 & 1.64 & 0.60 & \textbf{0.3044} {\scriptsize [0.2476, 0.3483]} & \textbf{0.4667} {\scriptsize [0.3969, 0.5166]} & 0.4882 & 0.4471 & 1.50 & 0.82 \\
\rowcolor{VLMRow} SED~\cite{xie2024sed}$^\ast$ & ConvNeXt-L & 0.3182 {\scriptsize [0.255, 0.373]} & 0.4828 {\scriptsize [0.406, 0.544]} & \textbf{0.6291} & 0.3917 & 4.09 & 3.74 & 0.3070 {\scriptsize [0.263, 0.355]} & 0.4698 {\scriptsize [0.416, 0.524]} & \textbf{0.7509} & 0.3419 & 1.97 & 1.51 & 0.1650 {\scriptsize [0.121, 0.229]} & 0.2833 {\scriptsize [0.215, 0.373]} & \textbf{0.6365} & 0.1822 & 0.98 & 0.32 \\
\rowcolor{VLMRow} CAT-Seg~\cite{cho2024cat}$^\ast$ & ViT-L & 0.2909 {\scriptsize [0.226, 0.374]} & 0.4507 {\scriptsize [0.368, 0.545]} & \underline{0.5832} & 0.3673 & 3.09 & 1.45 & 0.2287 {\scriptsize [0.202, 0.255]} & 0.3722 {\scriptsize [0.336, 0.407]} & 0.4101 & 0.3407 & 1.70 & 4.44 & 0.1300 {\scriptsize [0.098, 0.160]} & 0.2300 {\scriptsize [0.179, 0.276]} & 0.2050 & 0.2622 & 0.82 & 0.46 \\
\rowcolor{VLMRow} EviVLM~\cite{pan2025evivlm}$^\ast$ & ViT-B & 0.3290 {\scriptsize [0.261, 0.385]} & 0.4952 {\scriptsize [0.414, 0.556]} & 0.5289 & 0.4654 & 4.57 & 5.72 & 0.2584 {\scriptsize [0.213, 0.311]} & 0.4106 {\scriptsize [0.351, 0.474]} & \underline{0.7417} & 0.2839 & 1.00 & 0.24 & 0.2487 {\scriptsize [0.182, 0.312]} & 0.3984 {\scriptsize [0.308, 0.476]} & \underline{0.5134} & 0.3254 & 0.66 & 0.82 \\
\hline
\textbf{VL-AcneSeg (Global)} & ViT-L & \underline{0.3407} {\scriptsize [0.282, 0.392]} & \underline{0.5082} {\scriptsize [0.440, 0.563]} & 0.4938 & 0.5238 & 3.69 & 2.43 & \underline{0.3814} {\scriptsize [0.345, 0.425]} & \underline{0.5522} {\scriptsize [0.513, 0.597]} & 0.5011 & 0.6149 & 1.83 & 0.94 & \underline{0.2671} {\scriptsize [0.213, 0.313]} & \underline{0.4216} {\scriptsize [0.352, 0.477]} & 0.3277 & 0.5908 & 1.23 & 0.93 \\
SAM~\cite{kirillov2023segment} & ViT-L & 0.3216 {\scriptsize [0.251, 0.409]} & 0.4867 {\scriptsize [0.401, 0.581]} & 0.5601 & 0.4303 & 3.21 & 1.66 & 0.3764 {\scriptsize [0.3306, 0.4284]} & 0.5469 {\scriptsize [0.4969, 0.5999]} & 0.7064 & 0.4462 & 1.54 & 0.64 & 0.2546 {\scriptsize [0.1568, 0.2919]} & 0.4058 {\scriptsize [0.2710, 0.4519]} & 0.3452 & 0.4923 & 0.91 & 1.34 \\
U-Net~\cite{ronneberger2015u} & CNN & 0.1418 {\scriptsize [0.107, 0.167]} & 0.2483 {\scriptsize [0.193, 0.286]} & 0.1490 & \textbf{0.7443} & 4.86 & 19.07 & 0.1771 {\scriptsize [0.1549, 0.1980]} & 0.3009 {\scriptsize [0.2682, 0.3305]} & 0.1857 & \textbf{0.7931} & 2.13 & 13.34 & 0.0863 {\scriptsize [0.0620, 0.1148]} & 0.1590 {\scriptsize [0.1167, 0.2060]} & 0.0884 & \textbf{0.7869} & 1.43 & 9.82 \\
PSPNet~\cite{zhao2017pyramid} & CNN & 0.1802 {\scriptsize [0.134, 0.212]} & 0.3053 {\scriptsize [0.236, 0.350]} & 0.2337 & 0.4402 & 2.33 & 1.21 & 0.3368 {\scriptsize [0.3015, 0.3724]} & 0.5038 {\scriptsize [0.4633, 0.5427]} & 0.4157 & \underline{0.6394} & 1.79 & 1.43 & 0.1199 {\scriptsize [0.0814, 0.1599]} & 0.2141 {\scriptsize [0.1506, 0.2757]} & 0.1375 & 0.4826 & 0.91 & 1.62 \\
DeepLabV3+~\cite{chen2018encoder} & CNN & 0.1807 {\scriptsize [0.118, 0.229]} & 0.3061 {\scriptsize [0.211, 0.373]} & 0.2592 & 0.3737 & 2.22 & 4.71 & 0.2603 {\scriptsize [0.2058, 0.3185]} & 0.4131 {\scriptsize [0.3413, 0.4832]} & 0.5224 & 0.3416 & 1.23 & 0.56 & 0.1681 {\scriptsize [0.1018, 0.2344]} & 0.2878 {\scriptsize [0.1848, 0.3798]} & 0.4078 & 0.2224 & 0.52 & 0.38 \\
U-Net++~\cite{zhou2018unet++} & CNN & 0.1575 {\scriptsize [0.103, 0.221]} & 0.2721 {\scriptsize [0.186, 0.362]} & 0.3802 & 0.2118 & 3.34 & 11.29 & 0.1986 {\scriptsize [0.1627, 0.2342]} & 0.3315 {\scriptsize [0.2798, 0.3796]} & 0.4420 & 0.2650 & 1.71 & 4.37 & 0.2517 {\scriptsize [0.1817, 0.3309]} & 0.4022 {\scriptsize [0.3076, 0.4973]} & 0.3455 & 0.4822 & 1.39 & 3.88 \\
CE-Net~\cite{gu2019net} & CNN & 0.2215 {\scriptsize [0.161, 0.278]} & 0.3626 {\scriptsize [0.277, 0.434]} & 0.2836 & 0.5028 & 3.34 & 4.31 & 0.1248 {\scriptsize [0.0810, 0.1922]} & 0.2220 {\scriptsize [0.1499, 0.3224]} & 0.1495 & 0.4308 & 1.81 & 8.06 & 0.1234 {\scriptsize [0.0748, 0.1900]} & 0.2197 {\scriptsize [0.1392, 0.3194]} & 0.1306 & \underline{0.6915} & 1.57 & 7.93 \\
SegFormer~\cite{xie2021segformer} & Transformer & 0.2047 {\scriptsize [0.146, 0.263]} & 0.3399 {\scriptsize [0.254, 0.417]} & 0.2865 & 0.4177 & 3.90 & 10.83 & 0.1328 {\scriptsize [0.1119, 0.1538]} & 0.2345 {\scriptsize [0.2013, 0.2667]} & 0.1537 & 0.4974 & 1.79 & 17.21 & 0.0222 {\scriptsize [0.0142, 0.0317]} & 0.0434 {\scriptsize [0.0281, 0.0614]} & 0.0233 & 0.3072 & 0.88 & 16.38 \\
TransUNet~\cite{chen2021transunet} & Transformer & 0.2624 {\scriptsize [0.189, 0.352]} & 0.4157 {\scriptsize [0.317, 0.520]} & 0.5210 & 0.3459 & 3.09 & 2.36 & 0.1360 {\scriptsize [0.0937, 0.1845]} & 0.2394 {\scriptsize [0.1713, 0.3115]} & 0.3155 & 0.1928 & 0.93 & 0.60 & 0.0328 {\scriptsize [0.0097, 0.0794]} & 0.0634 {\scriptsize [0.0191, 0.1471]} & 0.0426 & 0.1245 & 0.23 & 1.57 \\
Swin-UNet~\cite{cao2022swin} & Transformer & 0.2089 {\scriptsize [0.150, 0.264]} & 0.3456 {\scriptsize [0.262, 0.418]} & 0.3266 & 0.3670 & 3.12 & 7.53 & 0.2005 {\scriptsize [0.1563, 0.2476]} & 0.3341 {\scriptsize [0.2703, 0.3969]} & 0.4446 & 0.2676 & 1.19 & 3.79 & 0.1454 {\scriptsize [0.0737, 0.2346]} & 0.2539 {\scriptsize [0.1373, 0.3800]} & 0.2390 & 0.2709 & 0.73 & 3.84 \\
nnU-Net~\cite{isensee2021nnunet} & CNN & 0.1909 {\scriptsize [0.142, 0.242]} & 0.3206 {\scriptsize [0.248, 0.390]} & 0.2311 & 0.5234 & 4.09 & 17.97 & 0.2079 {\scriptsize [0.177, 0.240]} & 0.3442 {\scriptsize [0.301, 0.387]} & 0.2927 & 0.4177 & 1.53 & 3.50 & 0.0454 {\scriptsize [0.017, 0.084]} & 0.0869 {\scriptsize [0.033, 0.155]} & 0.0575 & 0.1779 & 0.39 & 4.62 \\
Swin-UMamba$\dagger$~\cite{liu2024swinumamba} & Mamba & 0.2877 {\scriptsize [0.207, 0.355]} & 0.4468 {\scriptsize [0.343, 0.524]} & 0.3329 & \underline{0.6793} & 4.98 & 7.84 & 0.3278 {\scriptsize [0.295, 0.364]} & 0.4937 {\scriptsize [0.456, 0.534]} & 0.4933 & 0.4941 & 1.69 & 2.13 & 0.2552 {\scriptsize [0.188, 0.326]} & 0.4066 {\scriptsize [0.316, 0.492]} & 0.3566 & 0.4729 & 0.75 & 11.68 \\
\hline
\end{tabular}
}
\end{table*}

\begin{figure*}[!t]
    \centering
    \includegraphics[width=\textwidth]{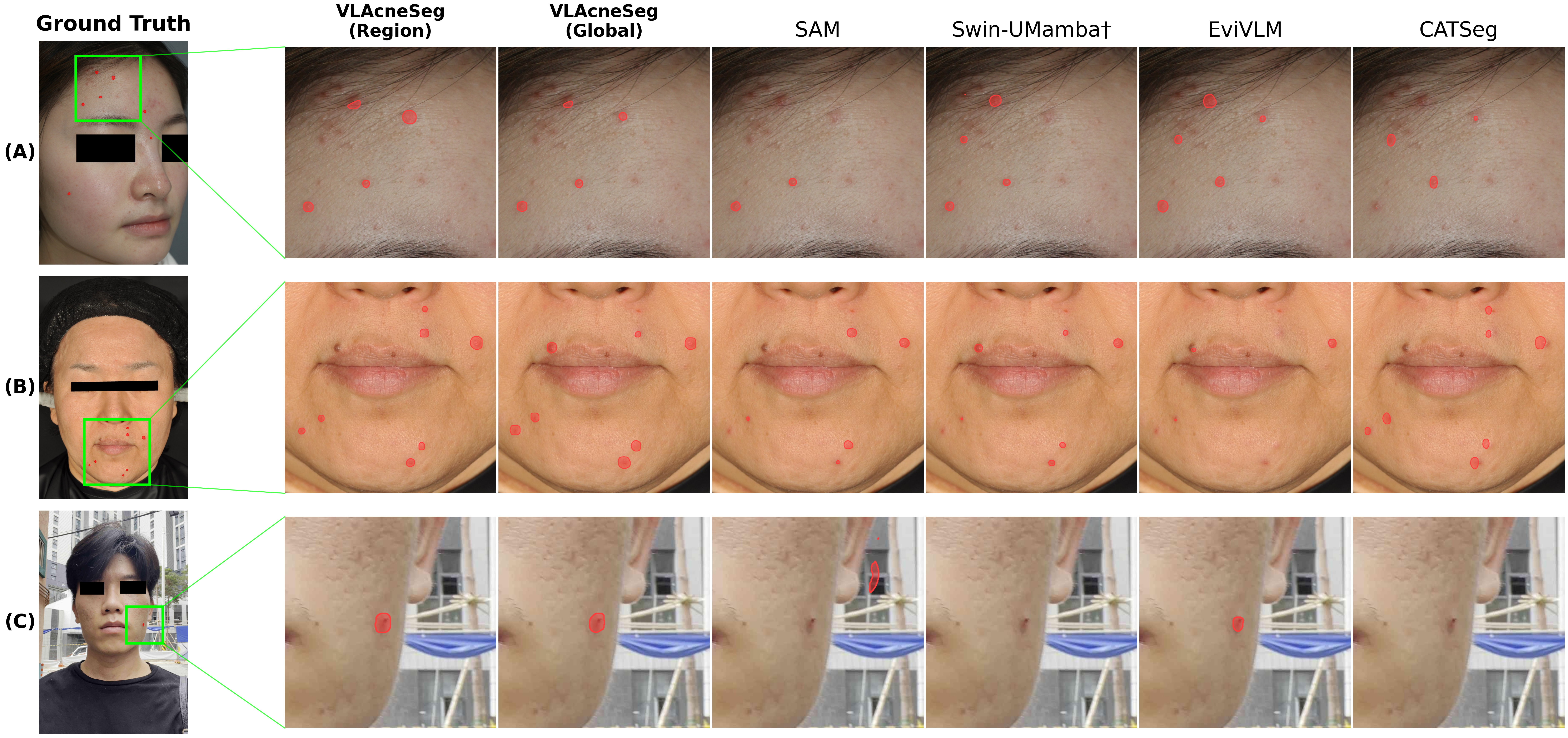}
    \caption{\textbf{Qualitative comparison across the three datasets}: (A) internal, (B) external controlled, and (C) external smartphone images. The left column shows the reference annotation on the full face; the green box marks the region enlarged in the remaining columns, so that lesions a few pixels across remain visible at print size. Predicted masks are overlaid in red, and both prompting conditions of the proposed method are shown. A comparison including all baselines and both prompting conditions is given in Figs.~\ref{fig:comprehensive_a}--\ref{fig:comprehensive} of the Supplementary Material.}
    \label{fig:fig4}
\end{figure*}

As summarized in \textcolor{IEEEBLUE}{Table~\ref{tab:table1}}, VL-AcneSeg attains the highest IoU and Dice scores in every evaluation setting. While traditional architectures like U-Net achieve high recall, they tend to be indiscriminate and often fail to distinguish acne lesions from visually similar conditions such as scars, post-inflammatory hyperpigmentation (PIH), and rashes. As shown in \textcolor{IEEEBLUE}{Fig.~\ref{fig:fig4}}, baseline models frequently misclassify these non-acne conditions as lesions, whereas VL-AcneSeg correctly identifies true acne through vision-language alignment.

Because the vision-language baselines are all supplied with region-level prompts, they are compared against VL-AcneSeg under the same regional prompting condition, whereas the remaining methods, which take no text input, are compared against the global-prompt configuration; in both cases the difference is assessed by paired patient-level cluster bootstrap, with the two-sided $p$ value taken as twice the proportion of resamples in which the difference lies on the opposite side of zero from the point estimate. Every baseline other than SAM differs significantly: $p \leq 0.005$ for the convolutional, Transformer and Mamba architectures, and 0.013, 0.034 and 0.048 for CAT-Seg, SED and EviVLM. SAM is the one method from which the proposed model is not distinguished ($p$ = 0.58). We therefore retrained it and the global-prompt configuration three times with different seeds. IoU was 0.3460 $\pm$ 0.0049 against 0.3219 $\pm$ 0.0052 and Dice 0.5141 $\pm$ 0.0054 against 0.4871 $\pm$ 0.0065; the variation across seeds is about five times smaller than the gap between them, and the ordering held every time. The remaining configurations were trained once, and \textcolor{IEEEBLUE}{Table~\ref{tab:table1}} reports that run. The direction of every comparison favours VL-AcneSeg and the ordering is preserved across all three datasets, although with 16 test patients the intervals around the smaller differences remain wide; these are reported in Table~\ref{tab:supp_diffci} of the Supplementary Material.

On external datasets, VL-AcneSeg maintains consistent performance across both controlled and real-world settings. Most baseline models show notable degradation, with Transformer-based architectures being particularly vulnerable to distribution shifts compared to CNN-based models, likely due to their weaker inductive biases~\cite{wang2022can}. The gap becomes more pronounced in the real-world smartphone setting, where CAT-Seg and SED degrade substantially and fall below several unimodal baselines, indicating that general-purpose VLM architectures do not readily transfer to domain-shifted clinical environments. VL-AcneSeg, by contrast, retains stable segmentation under unpredictable lighting and varying facial angles.

\subsection{Ablation Study}
\label{sec:ablation}
We evaluated the contributions of VL-AcneSeg's modules and hyperparameters through three studies: (1) architectural components, (2) image patch resolutions, and (3) text prompt strategies. Results on the external datasets are included in \textcolor{IEEEBLUE}{Table~\ref{tab:table2}}, and the corresponding patch-resolution comparison is given in Table~\ref{tab:appendix_ablation_final} of the Supplementary Material.

\subsubsection{Effectiveness of Architectural Components}
\begin{table*}[t]
\centering
\caption{Ablation study on the model components of VL-AcneSeg. Confidence intervals and $p$ values are obtained by paired cluster bootstrap against the Full Model, at the patient level for the internal dataset and at the image level for the external datasets. Specificity exceeds 0.999 everywhere and is omitted. FP lesions/img counts predicted lesion components matching no reference lesion. Best values are in \textbf{bold} and second-best are \underline{underlined}.}
\label{tab:table2}
\footnotesize
\begin{tabular}{lccccc}
\hline
\multirow{2}{*}{\textbf{Metrics}} & \multirow{2}{*}{\textbf{Full Model}} & \textbf{w/o cls} & \textbf{w/o v-v} & \textbf{w/o region} & \textbf{all-layer} \\
 & & \textbf{token} & \textbf{attention} & \textbf{mask loss} & \textbf{v-v} \\ \hline
\multicolumn{6}{l}{\textit{Internal}} \\
IoU          & \textbf{0.3602} {\scriptsize [0.309, 0.399]} & \underline{0.3563} {\scriptsize [0.294, 0.398]} & 0.3420 {\scriptsize [0.290, 0.380]} & 0.3510 {\scriptsize [0.290, 0.394]} & 0.2832 {\scriptsize [0.215, 0.336]} \\
Dice         & \textbf{0.5296} {\scriptsize [0.472, 0.570]} & \underline{0.5254} {\scriptsize [0.458, 0.568]} & 0.5097 {\scriptsize [0.451, 0.552]} & 0.5196 {\scriptsize [0.454, 0.565]} & 0.4414 {\scriptsize [0.357, 0.505]} \\
Precision    & \textbf{0.5202} & \underline{0.5149} & 0.4936 & 0.4935 & 0.3700 \\
Recall       & 0.5393 & 0.5363 & 0.5268 & \textbf{0.5486} & \underline{0.5449} \\
$p$-value & --- & 0.554 & 0.057 & 0.526 & $<$0.001 \\
TP lesions/img & 4.29 & 3.49 & 3.52 & 3.55 & 4.38 \\ 
FP lesions/img & 2.48 & 1.90 & 2.27 & 2.09 & 3.57 \\ \hline
\multicolumn{6}{l}{\textit{External (Controlled)}} \\
IoU          & \textbf{0.4232} {\scriptsize [0.376, 0.470]} & 0.4064 {\scriptsize [0.364, 0.452]} & 0.4020 {\scriptsize [0.359, 0.449]} & \underline{0.4097} {\scriptsize [0.366, 0.455]} & 0.3314 {\scriptsize [0.294, 0.374]} \\
Dice         & \textbf{0.5948} {\scriptsize [0.546, 0.640]} & 0.5780 {\scriptsize [0.534, 0.622]} & 0.5735 {\scriptsize [0.528, 0.620]} & \underline{0.5813} {\scriptsize [0.536, 0.625]} & 0.4979 {\scriptsize [0.454, 0.544]} \\
Precision    & \textbf{0.6110} & \underline{0.6107} & 0.5879 & 0.5947 & 0.4588 \\
Recall       & \textbf{0.5794} & 0.5486 & 0.5597 & \underline{0.5684} & 0.5442 \\
$p$-value & --- & 0.119 & 0.114 & 0.332 & $<$0.001 \\
TP lesions/img & 1.64 & 1.57 & 1.27 & 1.07 & 1.68 \\ 
FP lesions/img & 0.60 & 0.47 & 0.67 & 0.45 & 0.78 \\ \hline
\multicolumn{6}{l}{\textit{External (Real-world)}} \\
IoU          & \textbf{0.3044} {\scriptsize [0.246, 0.351]} & 0.2873 {\scriptsize [0.190, 0.365]} & 0.2819 {\scriptsize [0.218, 0.343]} & \underline{0.2905} {\scriptsize [0.216, 0.352]} & 0.1627 {\scriptsize [0.114, 0.217]} \\
Dice         & \textbf{0.4667} {\scriptsize [0.395, 0.520]} & 0.4464 {\scriptsize [0.320, 0.535]} & 0.4398 {\scriptsize [0.358, 0.511]} & \underline{0.4502} {\scriptsize [0.355, 0.520]} & 0.2798 {\scriptsize [0.204, 0.356]} \\
Precision    & \textbf{0.4882} & \underline{0.4571} & 0.3785 & 0.4269 & 0.1965 \\
Recall       & 0.4471 & 0.4361 & \textbf{0.5248} & \underline{0.4763} & 0.4860 \\
$p$-value & --- & 0.573 & 0.346 & 0.520 & $<$0.001 \\
TP lesions/img & 1.50 & 1.34 & 0.95 & 0.83 & 1.55 \\ 
FP lesions/img & 0.82 & 0.77 & 0.56 & 0.48 & 0.93 \\ \hline
\end{tabular}
\end{table*}

As shown in \textcolor{IEEEBLUE}{Table~\ref{tab:table2}}, the Full Model attains the highest IoU and Dice on all three datasets. Removing any single component lowers performance consistently but by a margin that the present test sets do not resolve individually, indicating that the CLS token, V-V attention and the region mask loss act in a complementary manner rather than through one dominant factor. The placement of V-V attention, by contrast, is clearly resolved: applying it across all layers, as in the original formulation, degrades performance on every dataset ($p < 0.01$) and raises the number of false positive lesions per image from 2.48 to 3.57 internally and from 0.82 to 0.93 in the real-world setting, consistent with the view that repeated value--value mixing erases the fine spatial detail on which small lesions depend.

\subsubsection{Impact of Image Patch Resolution}
\begin{table}[t]
\centering
\caption{Ablation study on different image patch resolutions for CLIP-based processing in VL-AcneSeg. Best values are in \textbf{bold} and second-best are \underline{underlined}.}
\label{tab:table3}
\footnotesize
\begin{tabular}{lccc}
\hline
\textbf{Metrics} & \textbf{$4 \times 3$ patches} & \textbf{$3 \times 2$ patches} & \textbf{$2 \times 1$ patches} \\ \hline
IoU          & \textbf{0.3602} & 0.3194 & 0.2700 \\
Dice         & \textbf{0.5296} & 0.4841 & 0.4253 \\
Precision    & \textbf{0.5202} & 0.4308 & 0.4070 \\
Recall       & \underline{0.5393} & \textbf{0.5524} & 0.4452 \\
TP lesions/img & \textbf{4.29} & 3.67 & 3.14 \\
FP lesions/img & 2.48 & 2.72 & 1.71 \\ \hline
\end{tabular}
\end{table}

We tested tiling configurations of $4\times3$, $3\times2$, and $2\times1$ patches. As summarized in \textcolor{IEEEBLUE}{Table~\ref{tab:table3}}, while the $3\times2$ configuration yields the highest Recall, it suffers from low Precision due to its inability to distinguish acne from scars. In contrast, $4\times3$ patches provide the most balanced results. Localization maps for the three configurations are shown in Fig.~\ref{fig:supp_patch} of the Supplementary Material, where finer granularity yields the detailed spatial priors needed to discriminate true lesions from confounding skin features. Furthermore, qualitative evidence of precise localization in unconstrained smartphone environments is provided in Fig.~\ref{fig:smartphone_localization_ablation} of the Supplementary Material.

\subsubsection{Role of the Text Prompt}

\begin{table}[t]
\centering
\caption{Where the performance originates. Each row removes one element of the vision--language pipeline. \textit{Unseen synonyms} and \textit{Shuffled text prompt} keep the CLIP text encoder but substitute synonyms absent from training or pair each image with the prompts of another image, testing whether the prompt is read as a reference to a particular area. \textit{Learnable embedding} and \textit{One-hot region encoding} replace the text embedding with a non-linguistic region code, separating language from spatial conditioning. \textit{CLIP vision encoder only} discards the text branch altogether, and \textit{w/o CLIP} removes the CLIP encoders as well, so that the interval between the two isolates the contribution of the pretrained visual representation. The last four settings were retrained.}
\label{tab:table5}
\footnotesize
\setlength{\tabcolsep}{3pt}
\resizebox{\columnwidth}{!}{
\begin{tabular}{lcccc}
\hline
\textbf{Prompt representation} & \textbf{IoU} & \textbf{Dice} & \textbf{Precision} & \textbf{Recall} \\ \hline
\multicolumn{5}{l}{\textit{Internal}} \\
Regional text prompt & 0.3602 {\scriptsize [0.309, 0.399]} & 0.5296 {\scriptsize [0.472, 0.570]} & 0.5202 & 0.5393 \\
Unseen synonyms & 0.3422 {\scriptsize [0.291, 0.392]} & 0.5100 {\scriptsize [0.451, 0.563]} & 0.5272 & 0.4938 \\
Shuffled text prompt & 0.3039 {\scriptsize [0.251, 0.356]} & 0.4661 {\scriptsize [0.402, 0.525]} & 0.5425 & 0.4085 \\
Learnable embedding & 0.3565 {\scriptsize [0.301, 0.399]} & 0.5256 {\scriptsize [0.463, 0.570]} & 0.4654 & 0.6037 \\
One-hot region encoding & 0.3111 {\scriptsize [0.260, 0.357]} & 0.4745 {\scriptsize [0.413, 0.526]} & 0.4313 & 0.5274 \\
CLIP vision encoder only & 0.2358 {\scriptsize [0.184, 0.270]} & 0.3816 {\scriptsize [0.311, 0.426]} & 0.3642 & 0.4007 \\
w/o CLIP (Swin only) & 0.1694 {\scriptsize [0.117, 0.219]} & 0.2897 {\scriptsize [0.210, 0.359]} & 0.2047 & 0.4957 \\ \hline
\multicolumn{5}{l}{\textit{External (Controlled)}} \\
Regional text prompt & 0.4232 {\scriptsize [0.376, 0.470]} & 0.5948 {\scriptsize [0.546, 0.640]} & 0.6110 & 0.5794 \\
Unseen synonyms & 0.3724 {\scriptsize [0.337, 0.414]} & 0.5427 {\scriptsize [0.504, 0.585]} & 0.6476 & 0.4671 \\
Shuffled text prompt & 0.2170 {\scriptsize [0.171, 0.267]} & 0.3567 {\scriptsize [0.292, 0.421]} & 0.5597 & 0.2617 \\
Learnable embedding & 0.4292 {\scriptsize [0.393, 0.474]} & 0.6006 {\scriptsize [0.565, 0.643]} & 0.6246 & 0.5784 \\
One-hot region encoding & 0.3732 {\scriptsize [0.335, 0.417]} & 0.5435 {\scriptsize [0.502, 0.588]} & 0.4885 & 0.6125 \\
CLIP vision encoder only & 0.2354 {\scriptsize [0.205, 0.272]} & 0.3811 {\scriptsize [0.340, 0.427]} & 0.3638 & 0.4002 \\
w/o CLIP (Swin only) & 0.2150 {\scriptsize [0.181, 0.251]} & 0.3539 {\scriptsize [0.307, 0.401]} & 0.2725 & 0.5049 \\ \hline
\multicolumn{5}{l}{\textit{External (Real-world)}} \\
Regional text prompt & 0.3044 {\scriptsize [0.248, 0.348]} & 0.4667 {\scriptsize [0.397, 0.517]} & 0.4882 & 0.4471 \\
Unseen synonyms & 0.2708 {\scriptsize [0.199, 0.334]} & 0.4262 {\scriptsize [0.332, 0.500]} & 0.4509 & 0.4041 \\
Shuffled text prompt & 0.2435 {\scriptsize [0.175, 0.290]} & 0.3916 {\scriptsize [0.298, 0.449]} & 0.4685 & 0.3364 \\
Learnable embedding & 0.2900 {\scriptsize [0.227, 0.347]} & 0.4496 {\scriptsize [0.369, 0.515]} & 0.5289 & 0.3910 \\
One-hot region encoding & 0.2734 {\scriptsize [0.212, 0.329]} & 0.4294 {\scriptsize [0.350, 0.495]} & 0.3864 & 0.4831 \\
CLIP vision encoder only & 0.1470 {\scriptsize [0.107, 0.188]} & 0.2563 {\scriptsize [0.194, 0.316]} & 0.2447 & 0.2692 \\
w/o CLIP (Swin only) & 0.0180 {\scriptsize [0.008, 0.035]} & 0.0353 {\scriptsize [0.016, 0.067]} & 0.0186 & 0.3475 \\ \hline
\end{tabular}
}
\end{table}

To test whether the spatial prior rests on the semantics of the anatomical terms or on memorised region identifiers, we prepared two clinical synonyms for each of the fifteen regions and at inference replaced every prompt by one of the two at random, without retraining or adapting the model (Table~\ref{tab:supp_synonyms} of the Supplementary Material). Accuracy is largely preserved: internal IoU moves from 0.3602 to 0.3422, with precision rising slightly and recall falling, a shift towards more conservative predictions rather than a loss of spatial grounding. Randomly pairing each image with the prompts belonging to another image is by contrast costly. On the controlled set IoU falls from 0.4232 to 0.2170, close to the level obtained without CLIP at all; internally the drop is more moderate, from 0.3602 to 0.3039. The prompt is therefore followed as a reference to a particular area rather than treated as an undifferentiated conditioning signal, although how much the correct reference is worth varies between the two datasets. Alternative prompt contents, adding a severity level or a lesion count to the region name, are compared in Table~\ref{tab:supp_promptstrategy}.

Read from the bottom, \textcolor{IEEEBLUE}{Table~\ref{tab:table5}} separates where the performance originates. Removing CLIP altogether leaves internal IoU at 0.1694 and real-world IoU at 0.0180, precision collapsing while recall is preserved: without the pretrained visual representation the model no longer distinguishes active lesions from post-inflammatory hyperpigmentation, pores and scars. Restoring the CLIP vision encoder alone recovers 0.2358, and adding region conditioning a further 0.3602, so both contribute substantially, while the visual representation is what sustains performance under domain shift. Among the conditioning schemes, a fixed one-hot index reaches only 0.3111 whereas a learnable per-region vector matches the text prompts (0.3565 against 0.3602): what matters is not that a region is named but that its code can be aligned with the patch features, which an orthogonal index cannot be. Either source yields one vector per region that enters the same cosine-similarity localization, so with a fixed set of fifteen areas the two are mechanistically equivalent and differ only in that the text encoder also resolves terms outside that vocabulary---a distinction that becomes practical only where the vocabulary is open or prompts arrive as free-form clinical language.

\begin{figure}[t!]
    \centering
    \setlength{\tabcolsep}{1pt} 
    \begin{tabular}{cccc}
        \multirow{4}{*}{\makecell{\includegraphics[width=0.23\columnwidth]{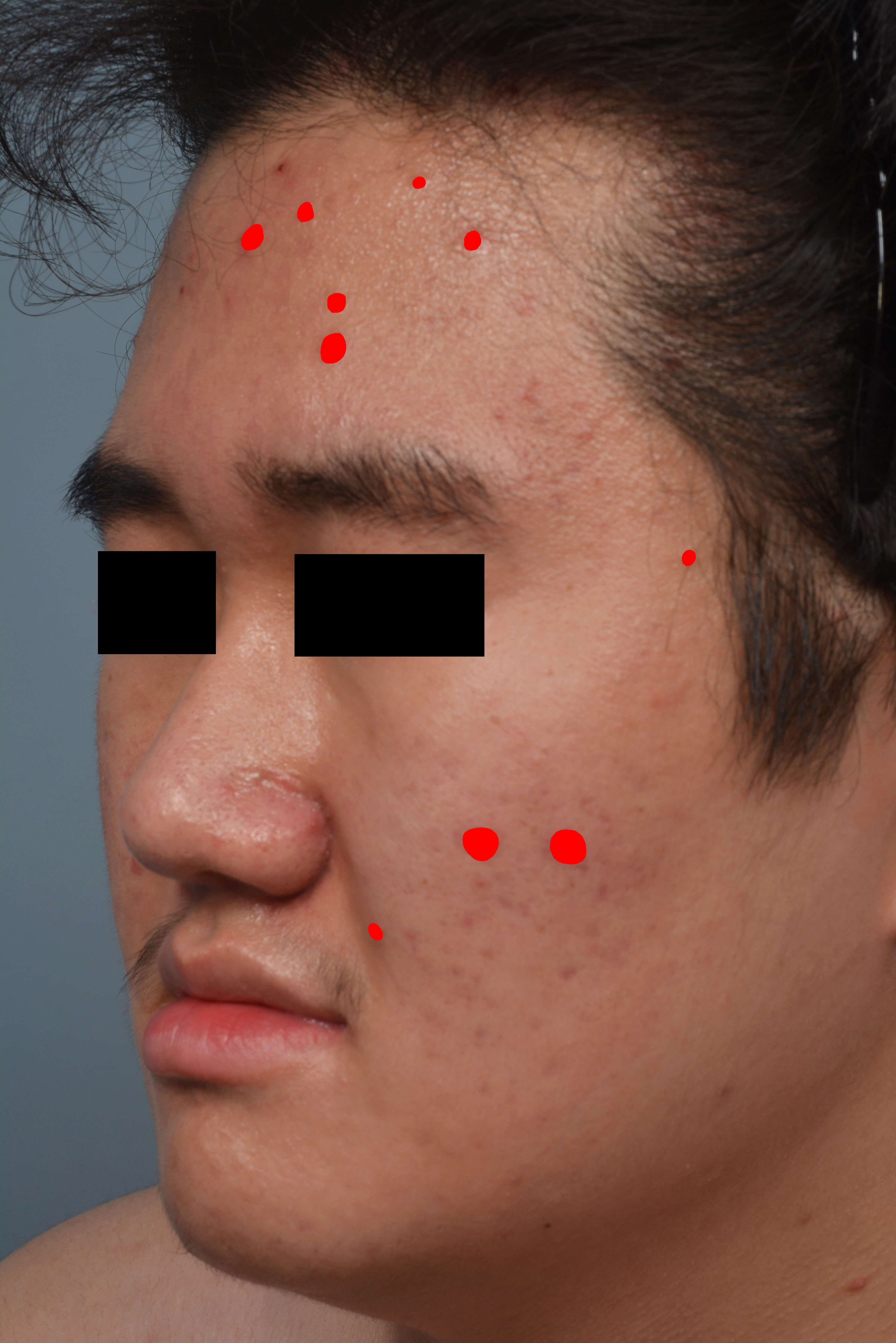}\\ \tiny (a) GT}} &
        \includegraphics[width=0.23\columnwidth]{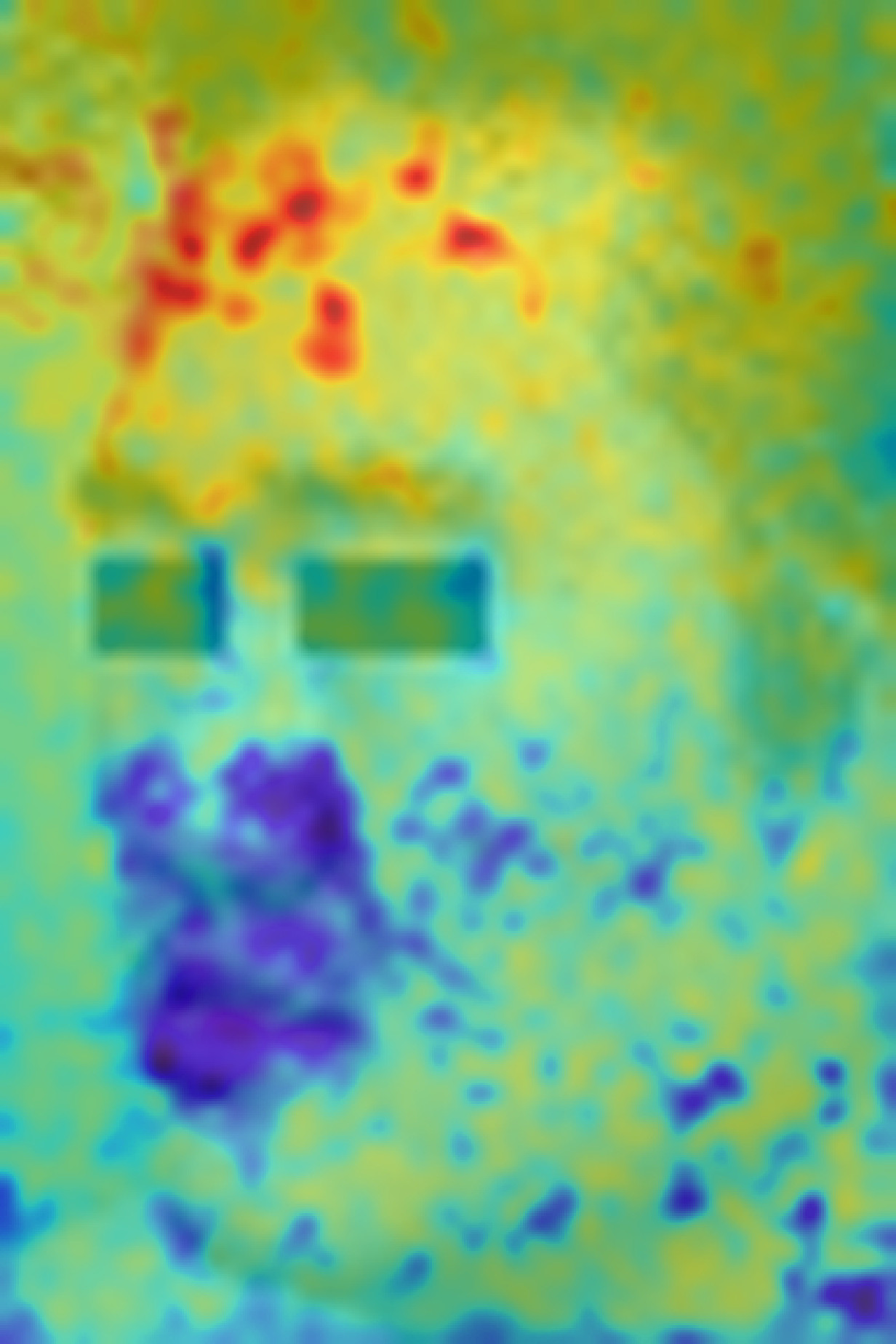} & 
        \includegraphics[width=0.23\columnwidth]{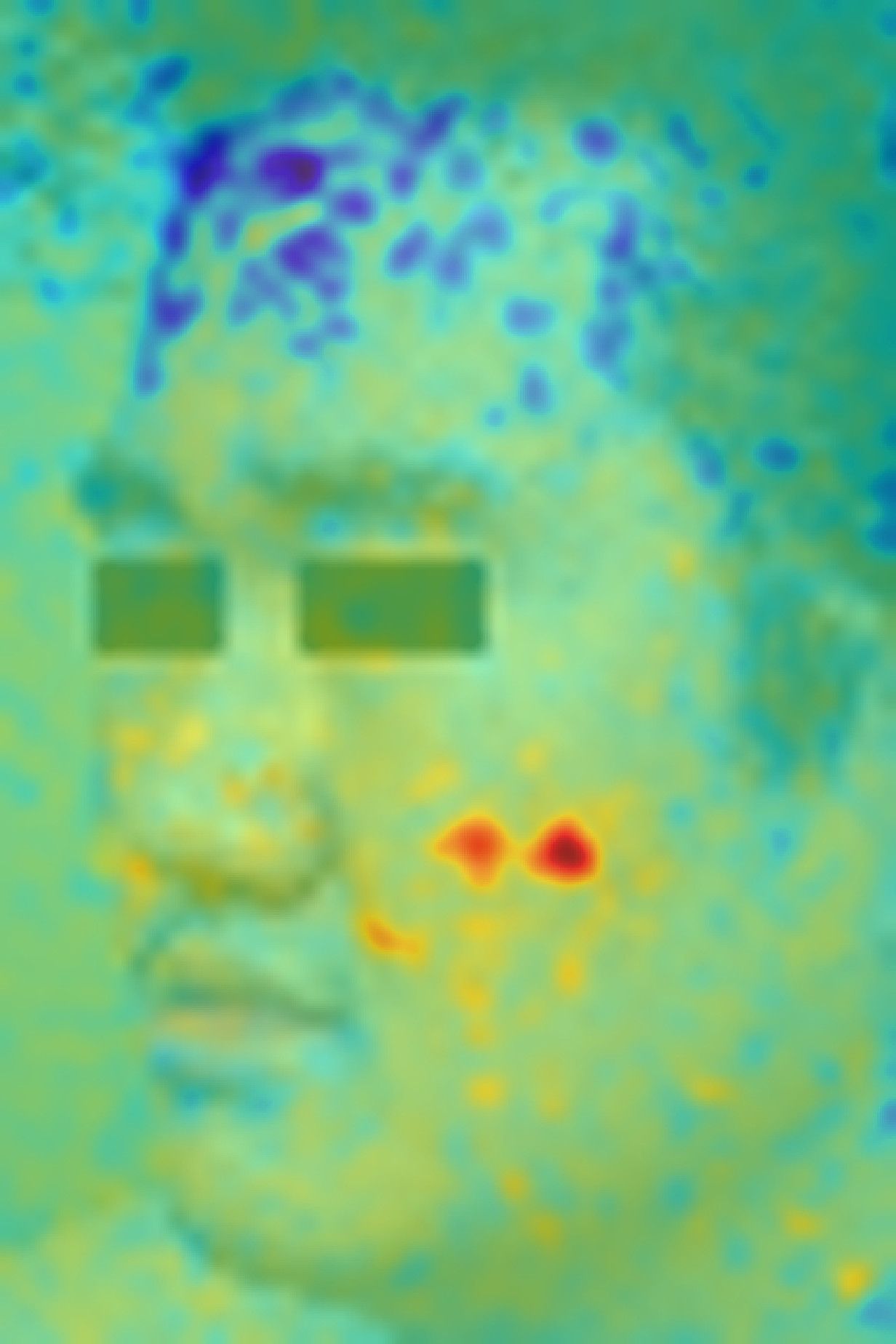} & 
        \includegraphics[width=0.23\columnwidth]{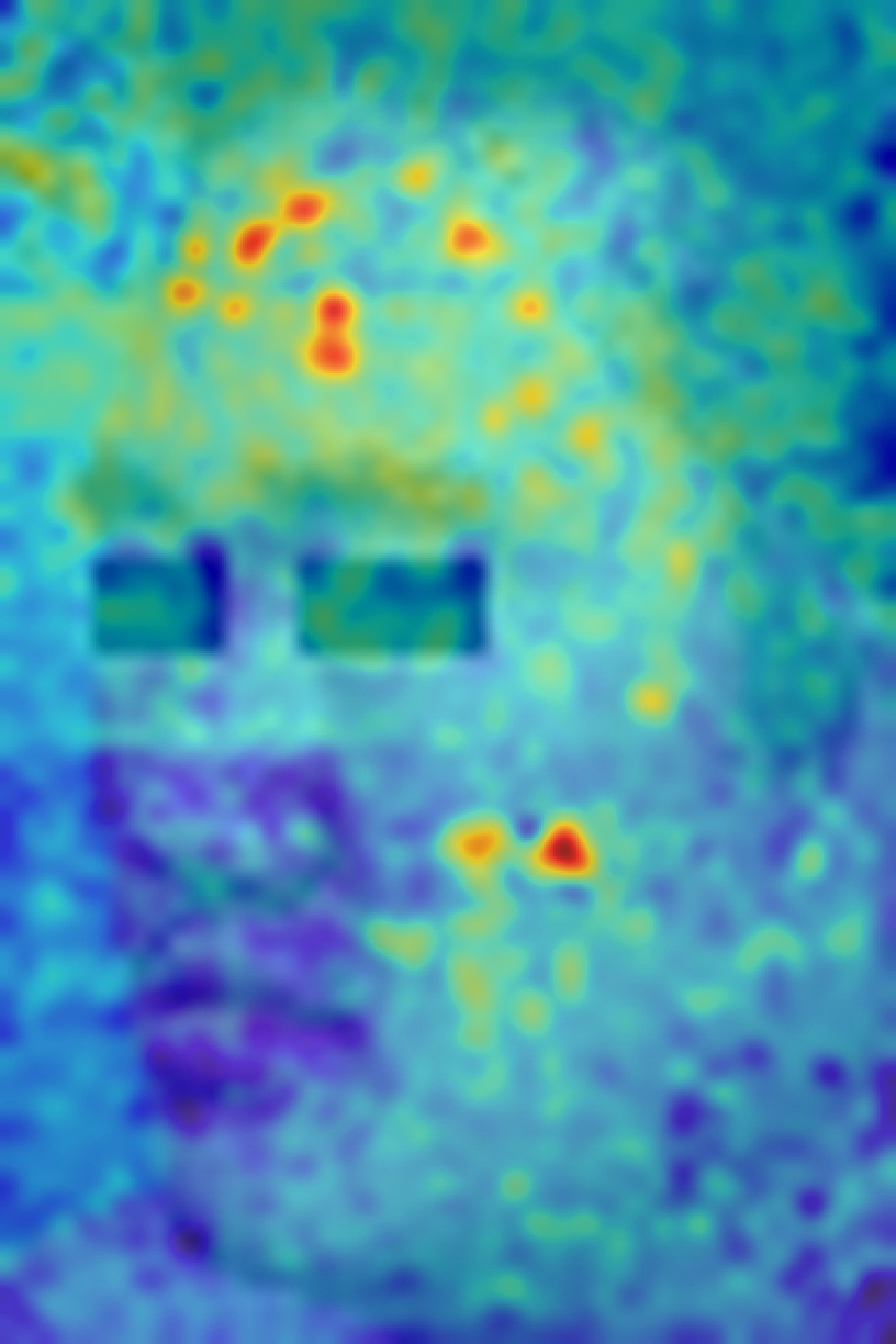} \\
        & \tiny (b) Forehead & \tiny (c) Right cheek & \tiny (d) Right temple \\[2pt]
        & \includegraphics[width=0.23\columnwidth]{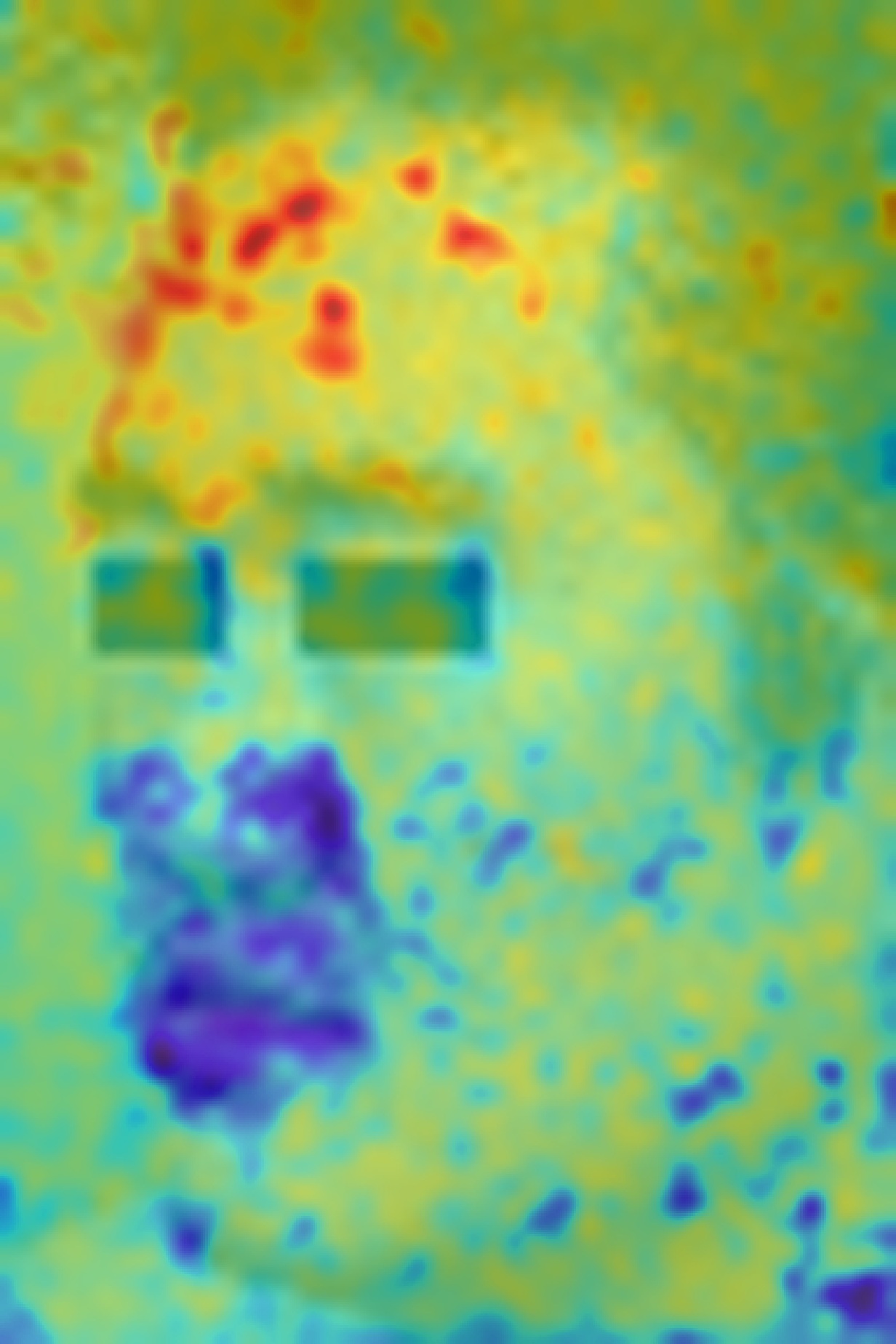} & 
        \includegraphics[width=0.23\columnwidth]{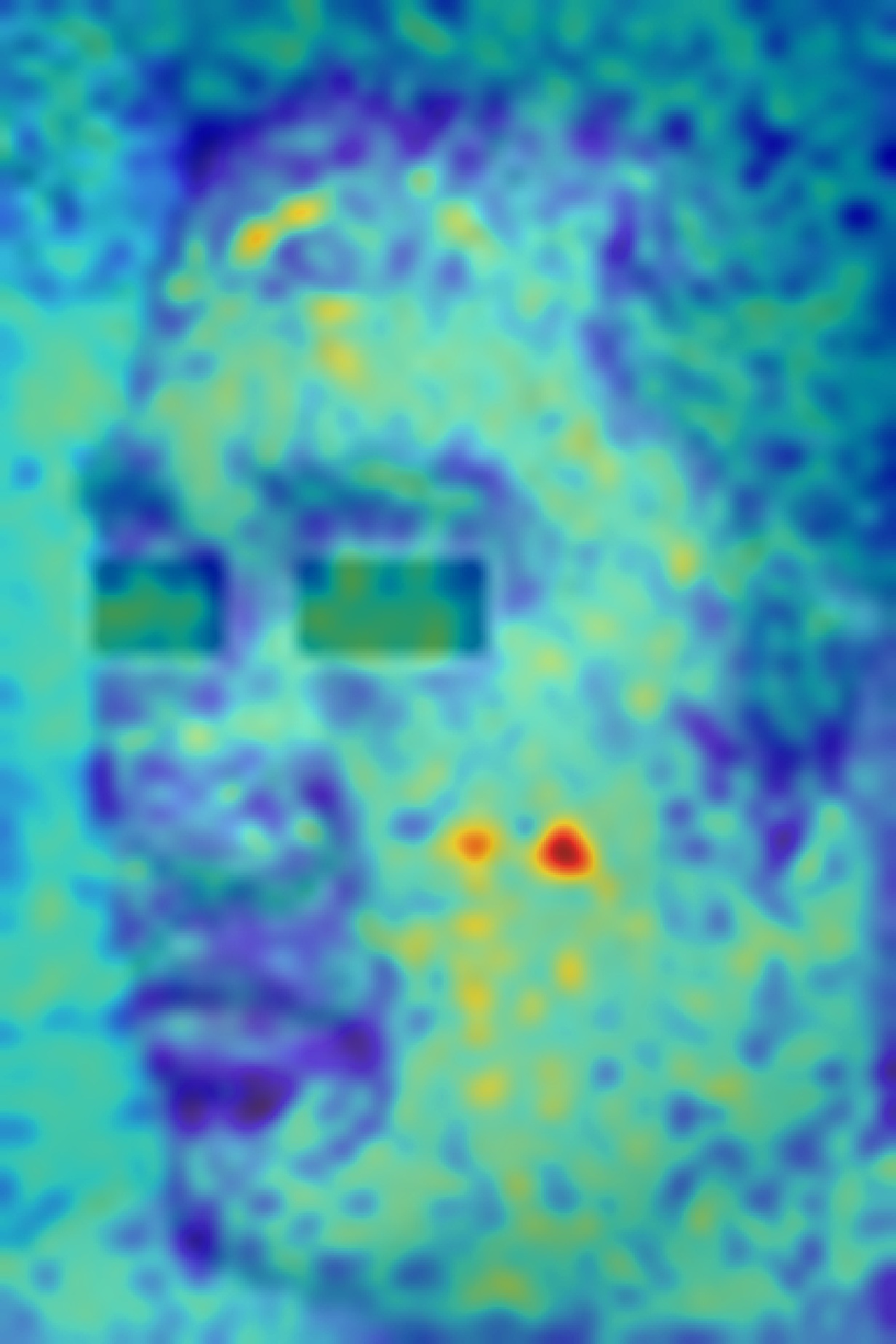} & 
        \includegraphics[width=0.23\columnwidth]{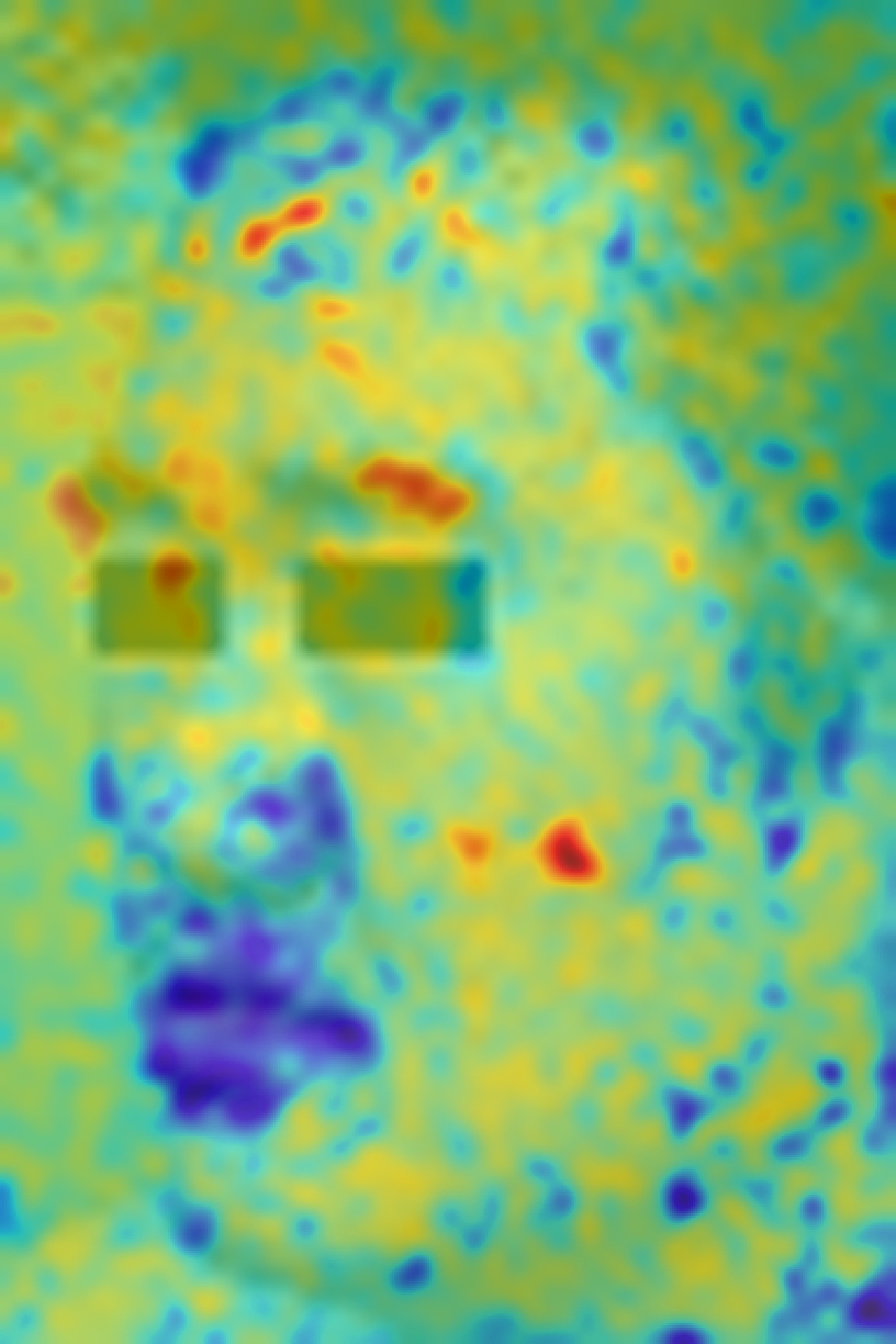} \\
        & \tiny (e) Frontal region & \tiny (f) Malar area & \tiny (g) Lateral forehead \\
    \end{tabular}
    \caption{\textbf{Anatomical region discrimination through text-guided spatial priors.} Top row: region names used during training. Bottom row: anatomical terms never seen during training, paired column-wise with the region above.}
    \label{fig:fig7}
\end{figure}

\textcolor{IEEEBLUE}{Fig.~\ref{fig:fig7}} shows how prompts steer spatial attention. For distinct areas such as the forehead (b) and cheek (c) the response is regionally dominant, whereas in the narrow temple region (d) the model falls back on the broader concept of an acne lesion. The bottom row probes terms absent from training: single-word clinical synonyms transfer, with ``frontal region'' (e) concentrating over the forehead and ``malar area'' (f) isolating the mid-face while suppressing the forehead response seen in (b) and (e). The compositional expression ``lateral forehead'' (g) instead yields a diffuse map, indicating that a spatial modifier is not fully composed with the anatomical noun.

\begin{figure}[t]
    \centering
    \includegraphics[width=\columnwidth]{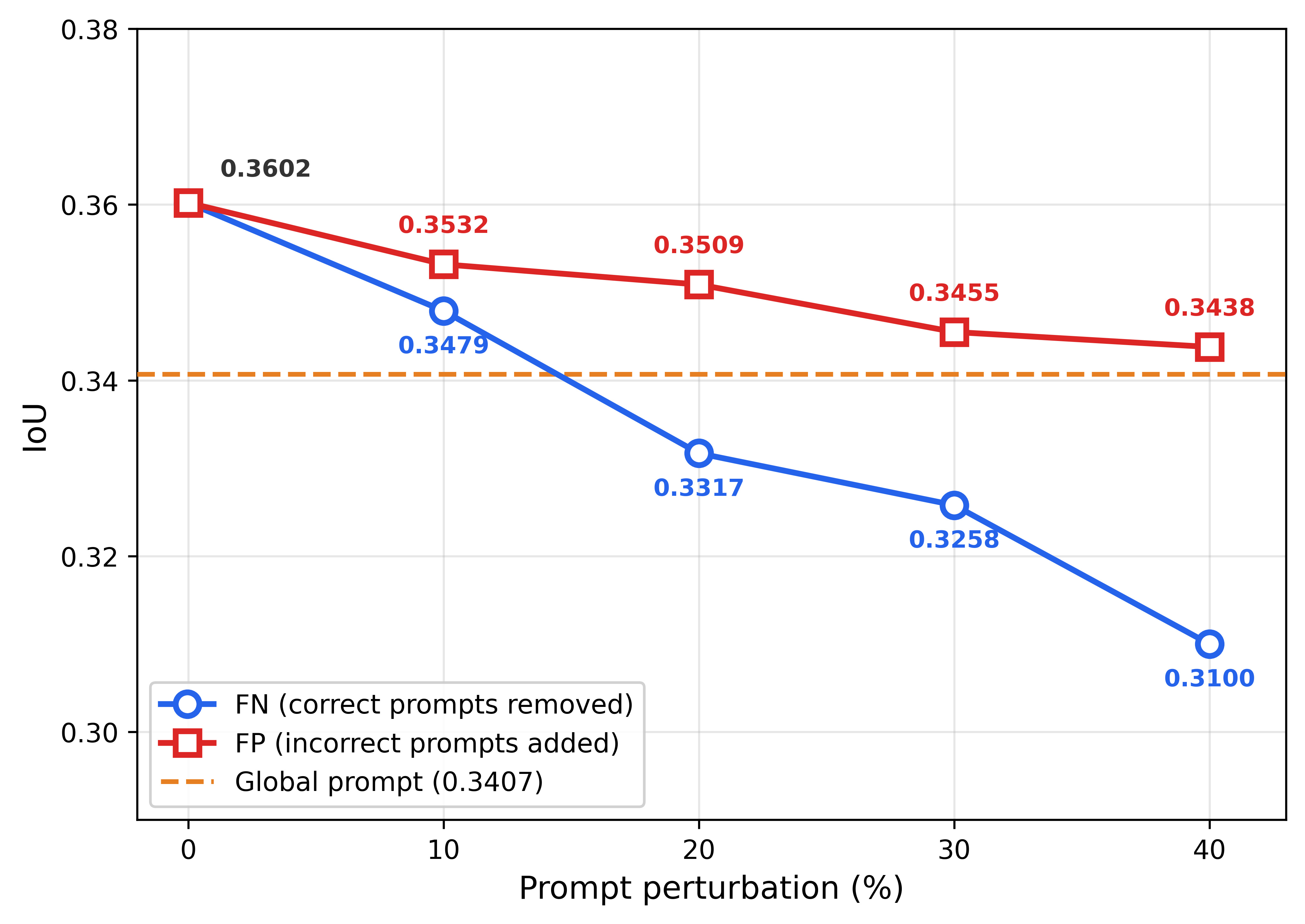}
    \caption{\textbf{Segmentation accuracy under prompt perturbation on the internal test set.} The 58 test images carry 165 lesion-bearing (positive) region prompts and 705 lesion-free (negative) regions. Each pool was shuffled once with a fixed seed, and the perturbation level denotes the fraction of that pool that is withheld (FN) or added (FP); the perturbed sets are nested. When every prompt of an image is withheld, it falls back to the single global prompt, marked by the dashed line.}
    \label{fig:fig_perturb}
\end{figure}

\begin{figure}[t!]
    \centering
    \setlength{\tabcolsep}{2pt}
    \begin{tabular}{>{\centering\arraybackslash}m{0.30\columnwidth} >{\centering\arraybackslash}m{0.30\columnwidth} >{\centering\arraybackslash}m{0.30\columnwidth}}
        \toprule
        \textbf{\footnotesize GT} & \textbf{\footnotesize Original Result} & \textbf{\footnotesize Adversarial Result} \\ \midrule

        \includegraphics[width=\linewidth]{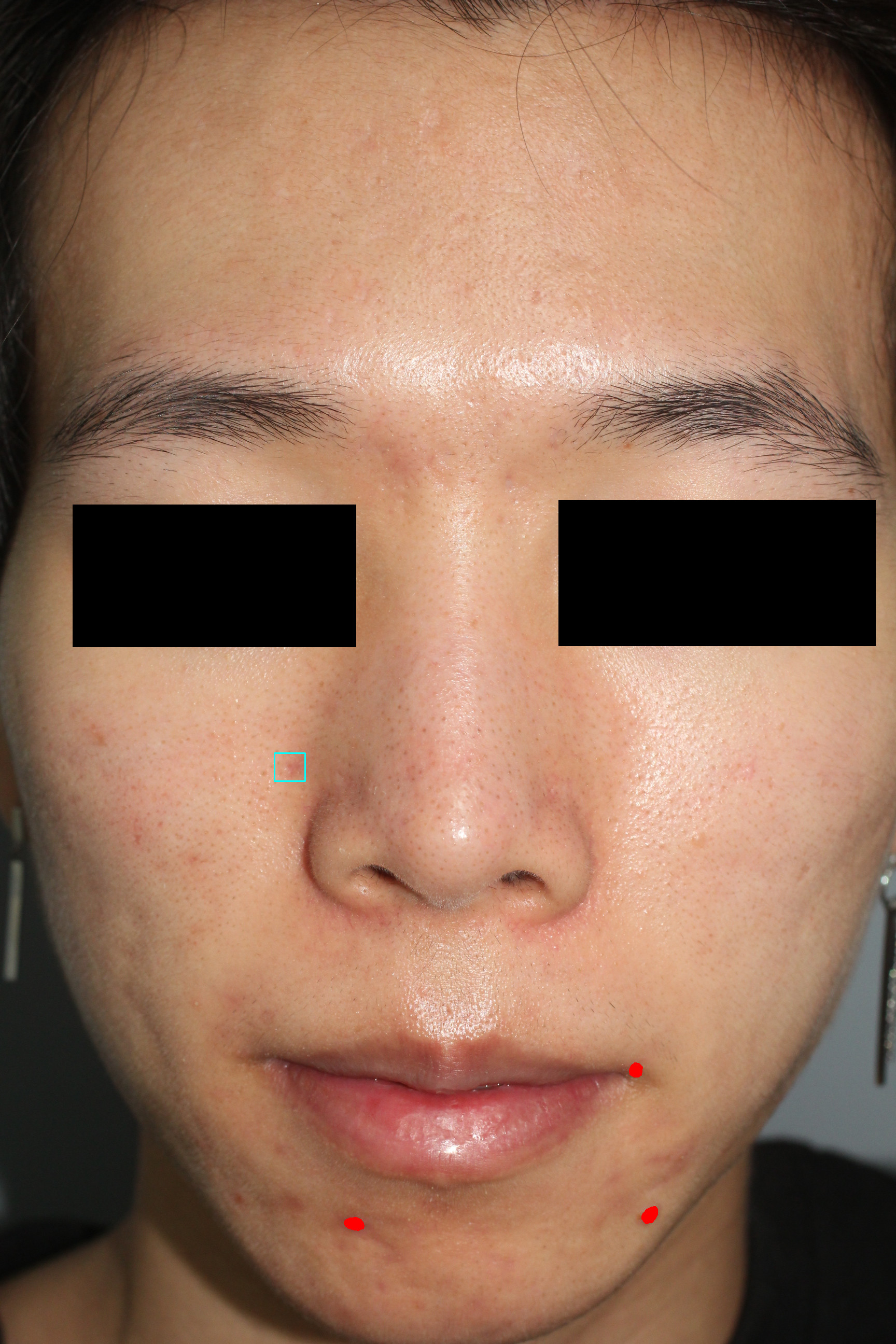} &
        \includegraphics[width=\linewidth]{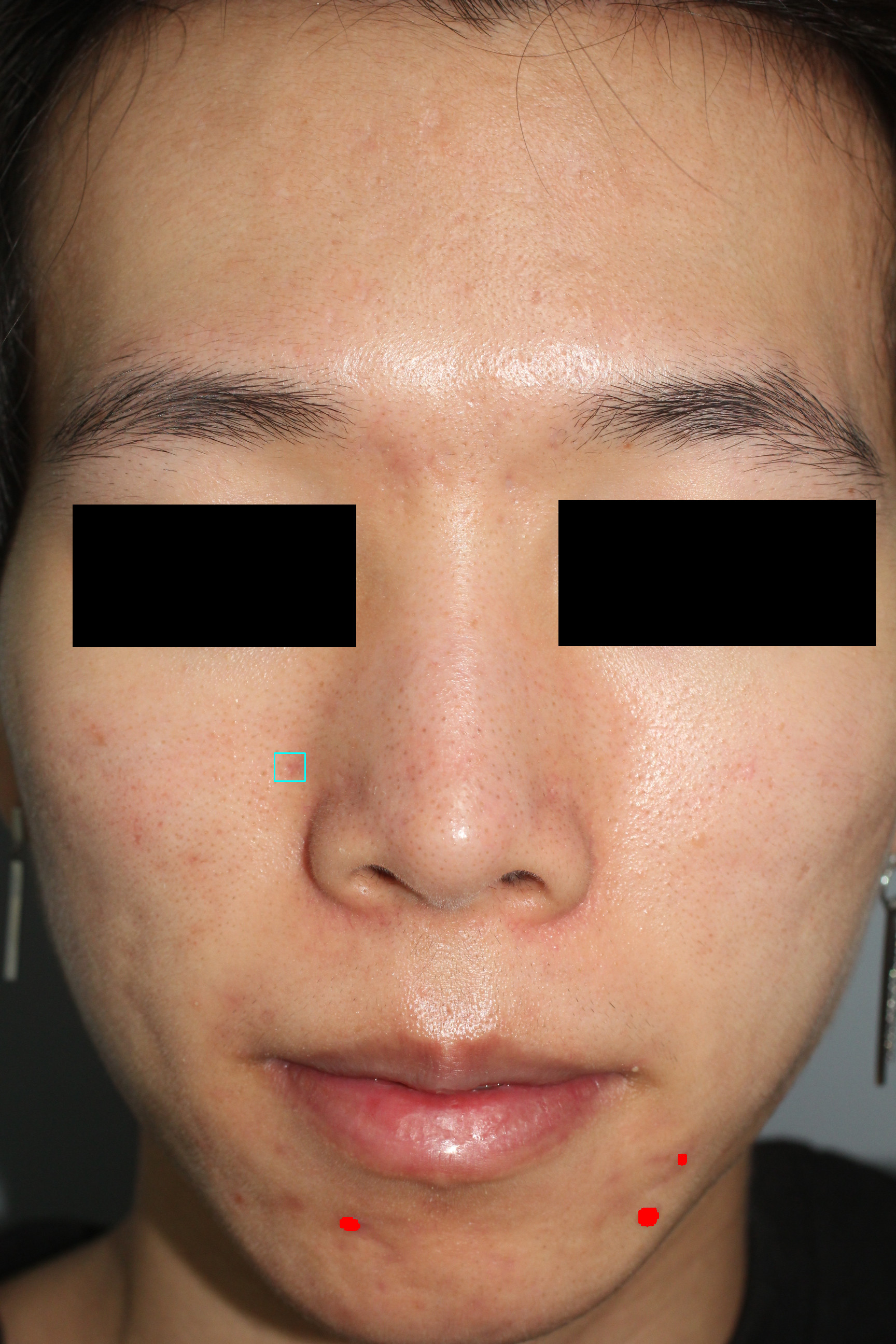} &
        \includegraphics[width=\linewidth]{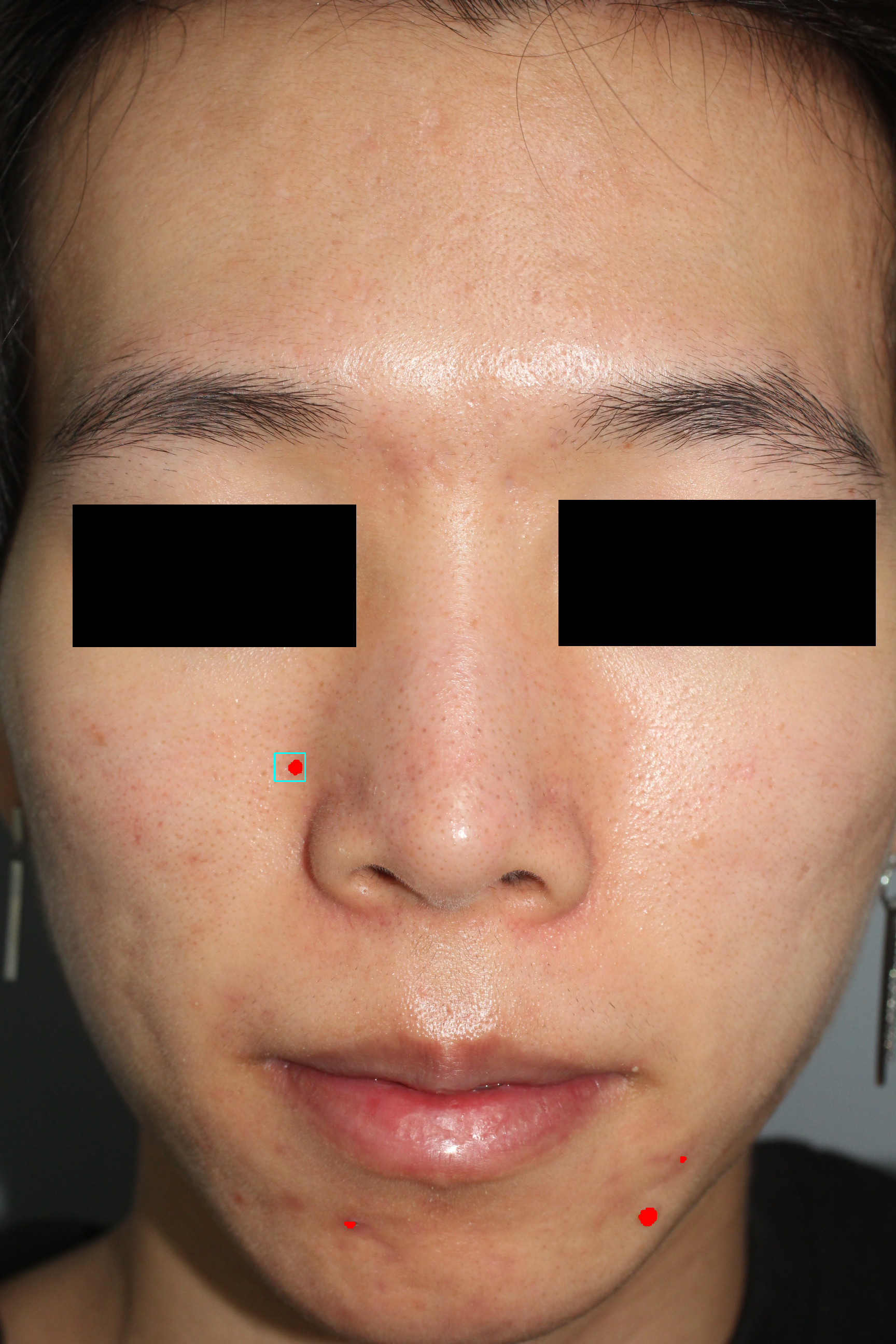} \\

        \fontsize{5pt}{6pt}\selectfont GT &
        \fontsize{5pt}{6pt}\selectfont \textbf{Ori.:} ``acne on \textit{lip}; \textcolor{red}{\textit{chin}}" (actual) &
        \fontsize{5pt}{6pt}\selectfont \textbf{Adv.:} ``acne on \textit{lip}; \textcolor{blue}{\textit{left cheek}}" (remove \textcolor{red}{\textit{chin}}, add \textcolor{blue}{\textit{left cheek}}) \\ \midrule

        \includegraphics[width=\linewidth]{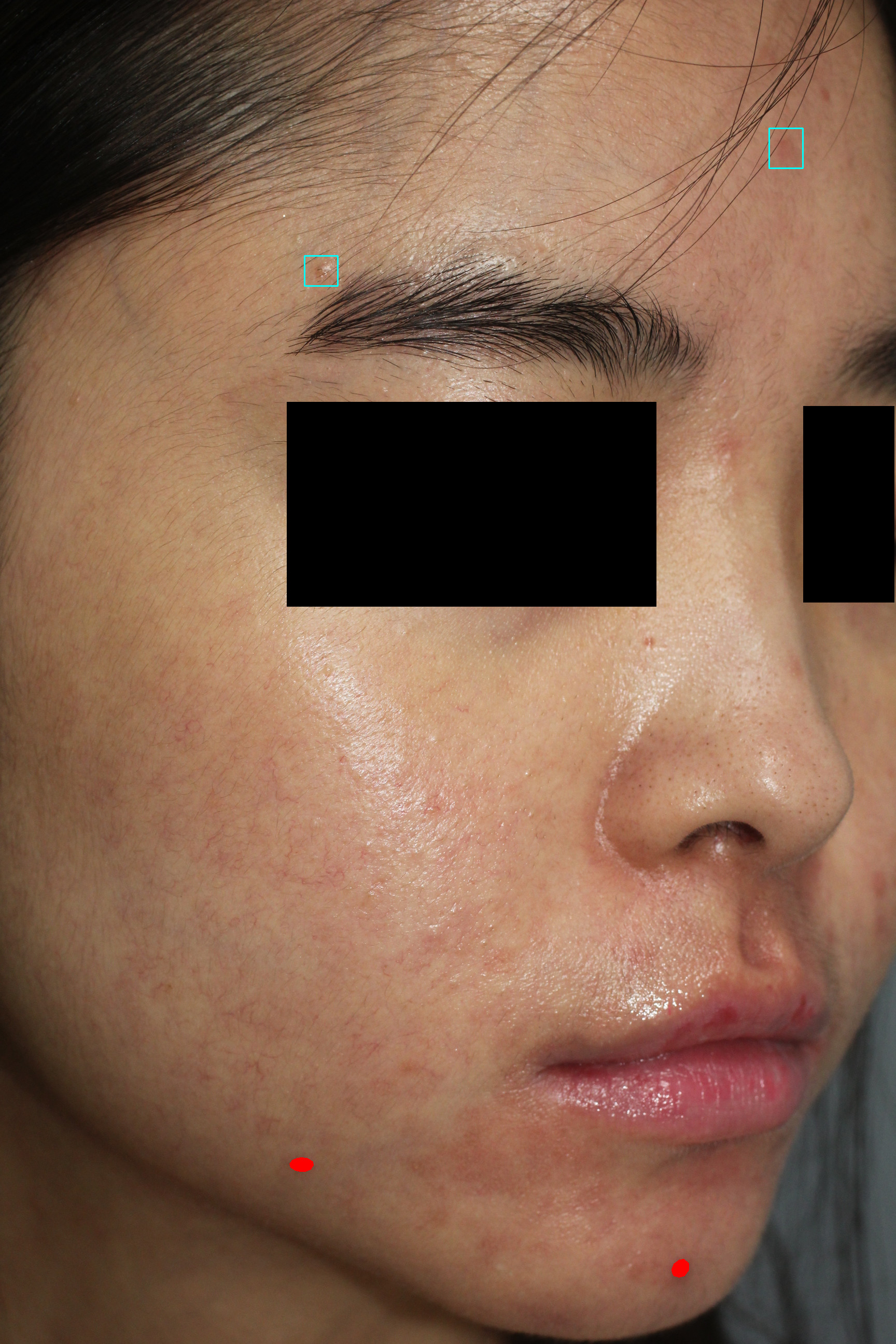} &
        \includegraphics[width=\linewidth]{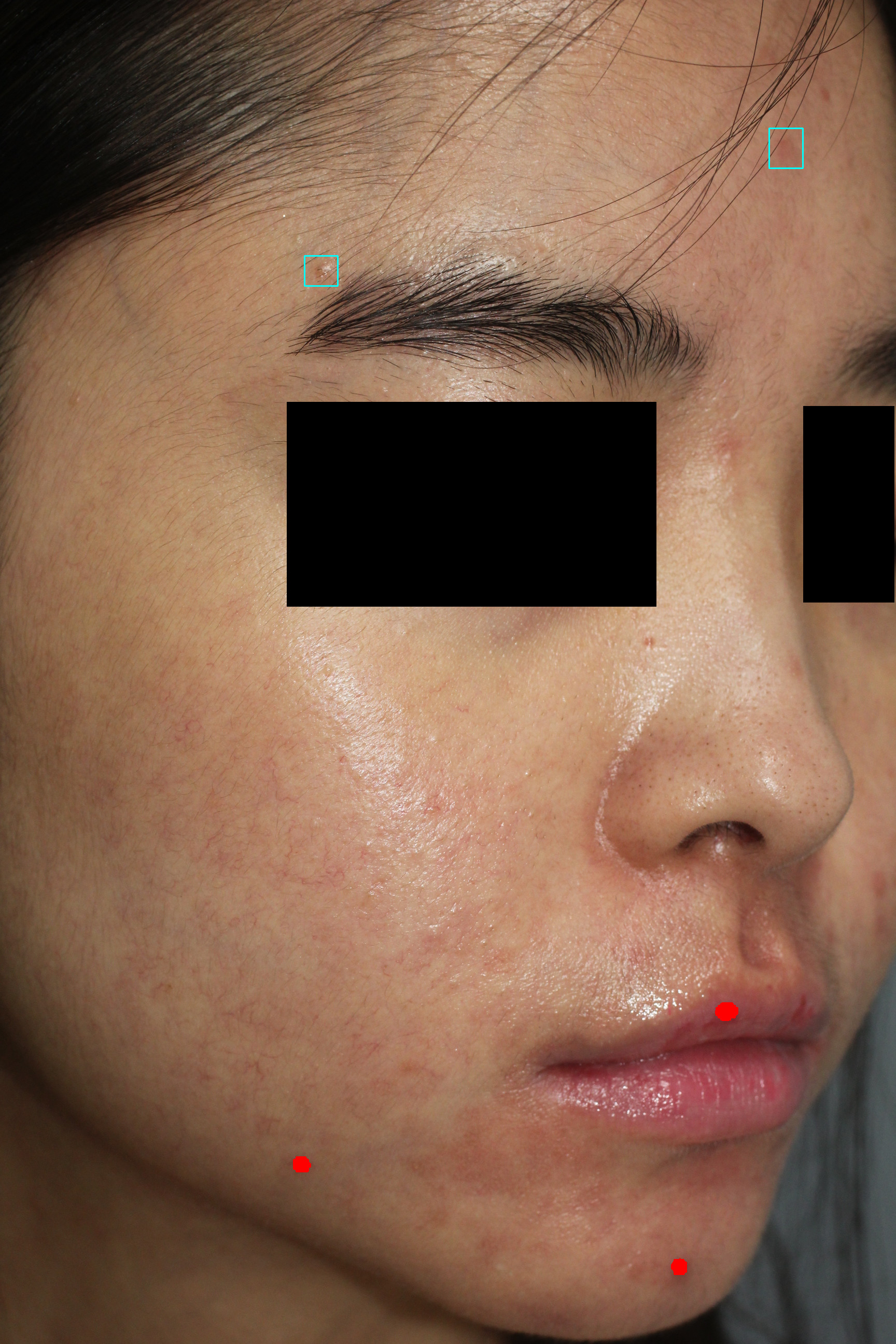} &
        \includegraphics[width=\linewidth]{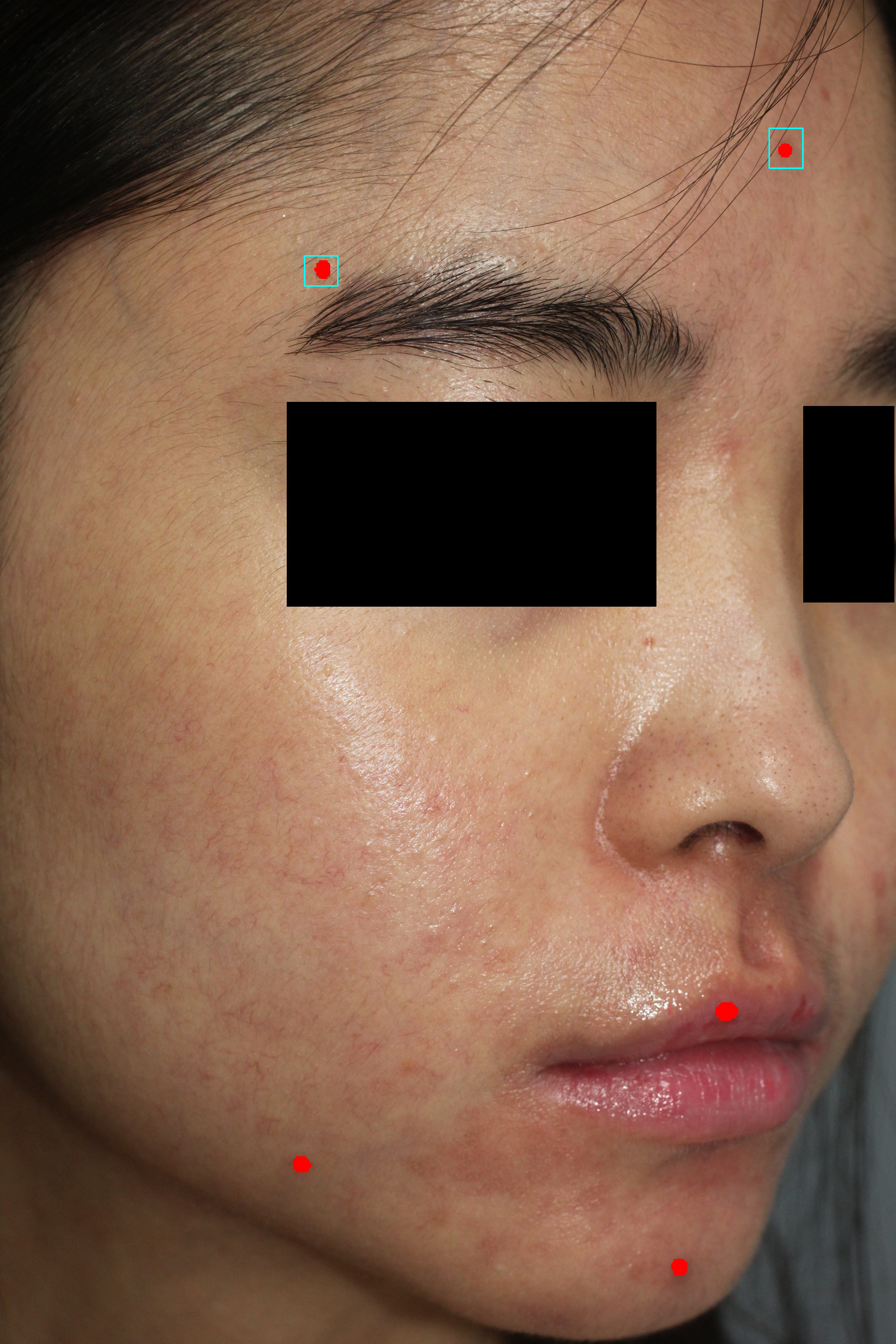} \\

        \fontsize{5pt}{6pt}\selectfont GT &
        \fontsize{5pt}{6pt}\selectfont \textbf{Ori.:} ``acne on \textcolor{red}{\textit{left cheek}}; \textit{chin}" (actual) &
        \fontsize{5pt}{6pt}\selectfont \textbf{Adv.:} ``acne on \textcolor{blue}{\textit{forehead}}; \textit{chin}" (remove \textcolor{red}{\textit{left cheek}}, add \textcolor{blue}{\textit{forehead}}) \\ \bottomrule
    \end{tabular}
    \caption{\textbf{Model response under adversarial prompting.} Prompts for lesion-bearing regions are omitted and misleading prompts are issued for lesion-free regions.}
    \label{fig:fig6}
\end{figure}

We further quantified the response to imperfect prompts (\textcolor{IEEEBLUE}{Fig.~\ref{fig:fig_perturb}}), withholding prompts for regions that do contain lesions and issuing prompts for regions that do not. The two behave asymmetrically. Withholding degrades accuracy steadily, from 0.3602 to 0.3100 IoU at 40\% perturbation, and falls below the global-prompt baseline of 0.3407 at roughly 15\%. Over-prompting is far better tolerated, declining only to 0.3438 at the same level and remaining above that baseline throughout the range examined. The asymmetry itself is informative: that omission costs more than over-prompting means the model attends less to regions it has not been told about, which is the behaviour the mechanism is intended to produce. Where the shuffled-prompt condition shows that the model follows the region it is given, the omission curve shows that it looks less closely at regions it is not given. The two error types are not symmetric in the pipeline either: a spurious region enters the localization stage as a candidate and can still be rejected during refinement, whereas a region that never enters it cannot be recovered downstream. Two adversarial cases in \textcolor{IEEEBLUE}{Fig.~\ref{fig:fig6}} illustrate both sides: lesions are still localized where prompts are withheld, since visual evidence remains, and under misleading prompts candidates appear only where textual guidance and visual evidence coincide. In practical terms a prompt source must therefore achieve high recall over lesion-bearing regions, while spurious regions carry comparatively little cost.

\subsection{Failure Case Analysis}
\label{sec:failure}
\label{sec:failure_analysis}
\begin{figure}[t]
    \centering
    \setlength{\tabcolsep}{2pt}
    \renewcommand{\arraystretch}{1.2}
    \begin{tabular}{>{\centering\arraybackslash}m{0.22\columnwidth} >{\centering\arraybackslash}m{0.22\columnwidth} >{\centering\arraybackslash}m{0.22\columnwidth} >{\centering\arraybackslash}m{0.22\columnwidth}}
        \toprule
        \multicolumn{2}{c}{\footnotesize \textbf{(A)}} & \multicolumn{2}{c}{\footnotesize \textbf{(B)}} \\
        \cmidrule(lr){1-2} \cmidrule(lr){3-4}
        \footnotesize GT & \footnotesize Pred & \footnotesize GT & \footnotesize Pred \\ \midrule
        \includegraphics[width=\linewidth]{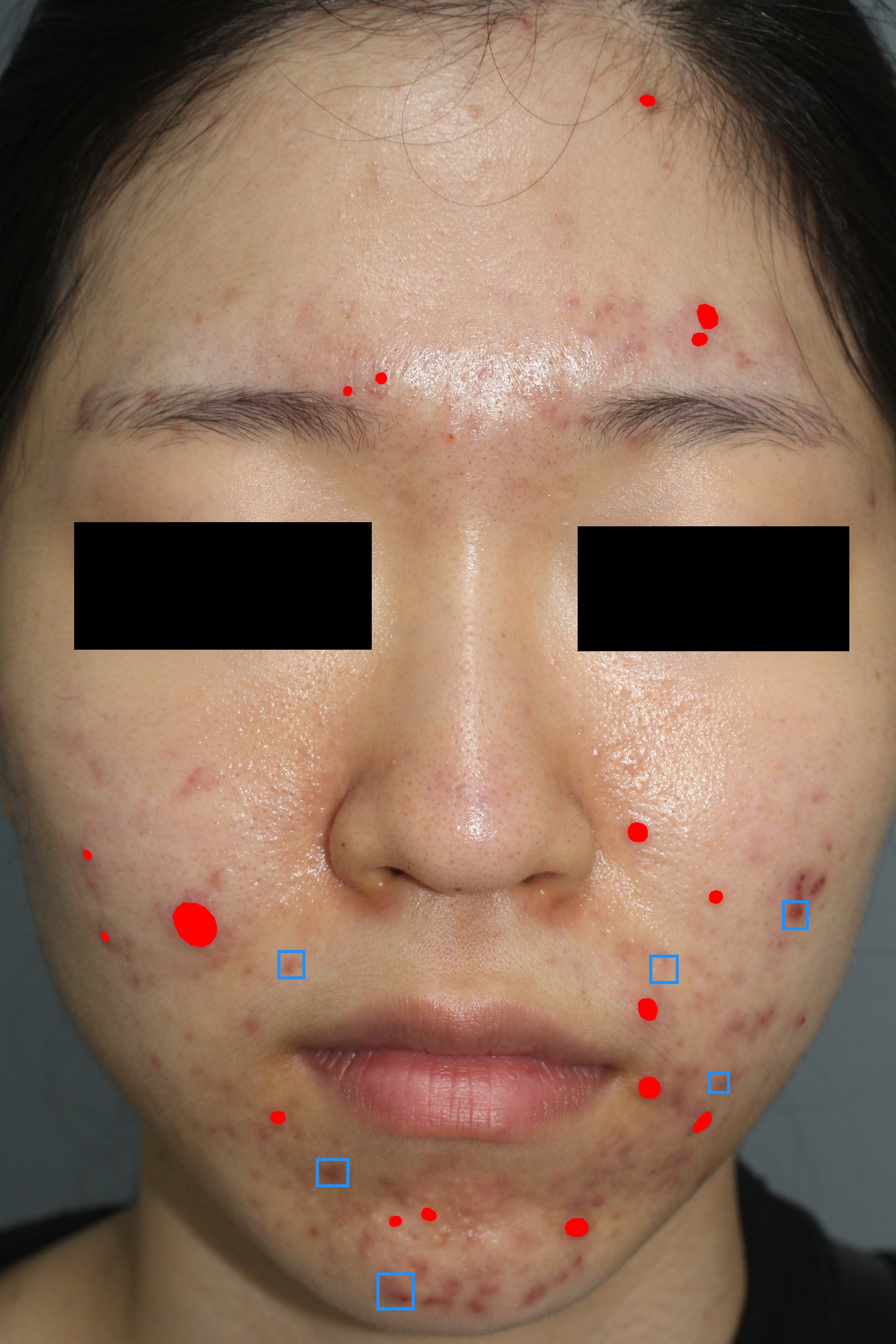} &
        \includegraphics[width=\linewidth]{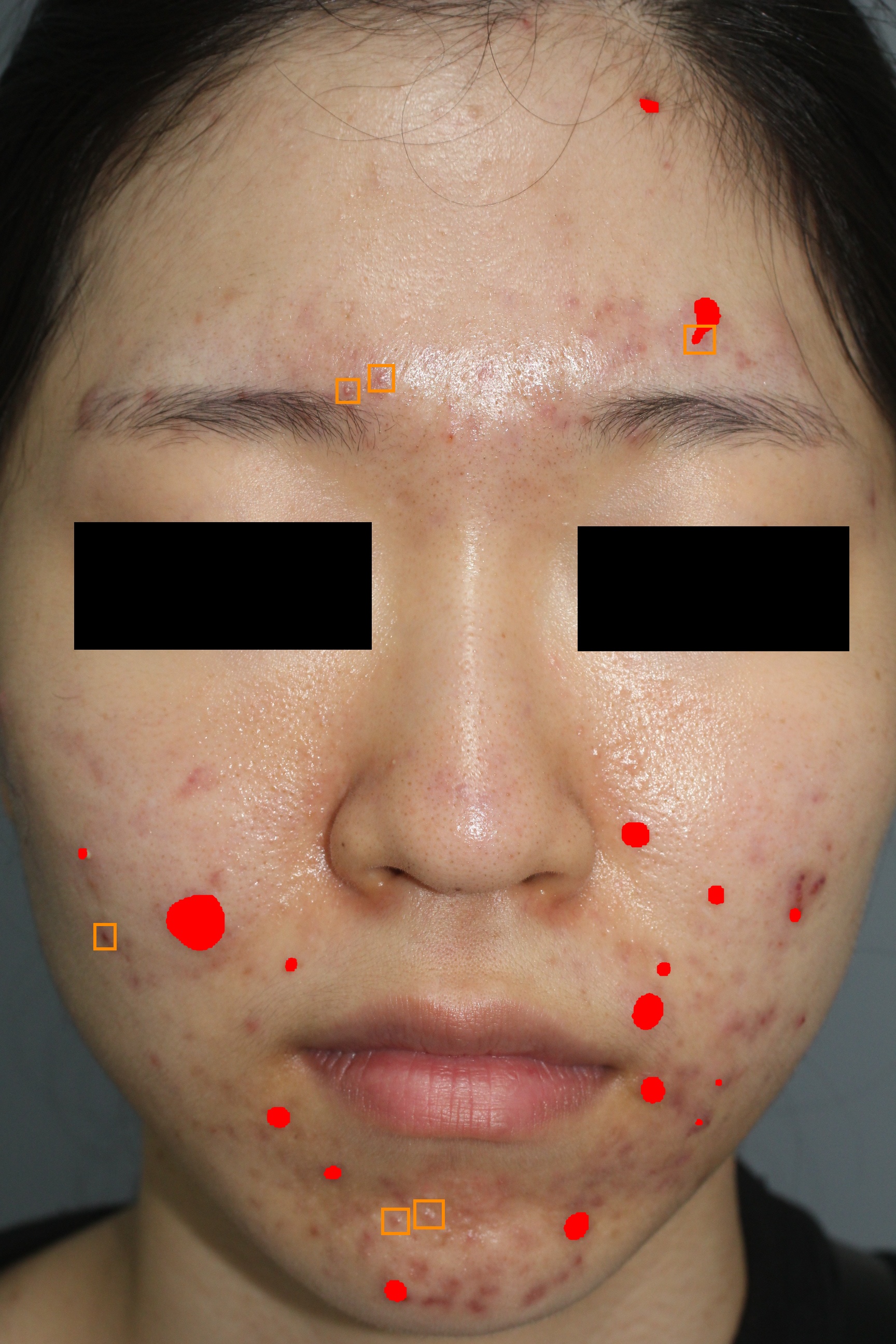} &
        \includegraphics[width=\linewidth]{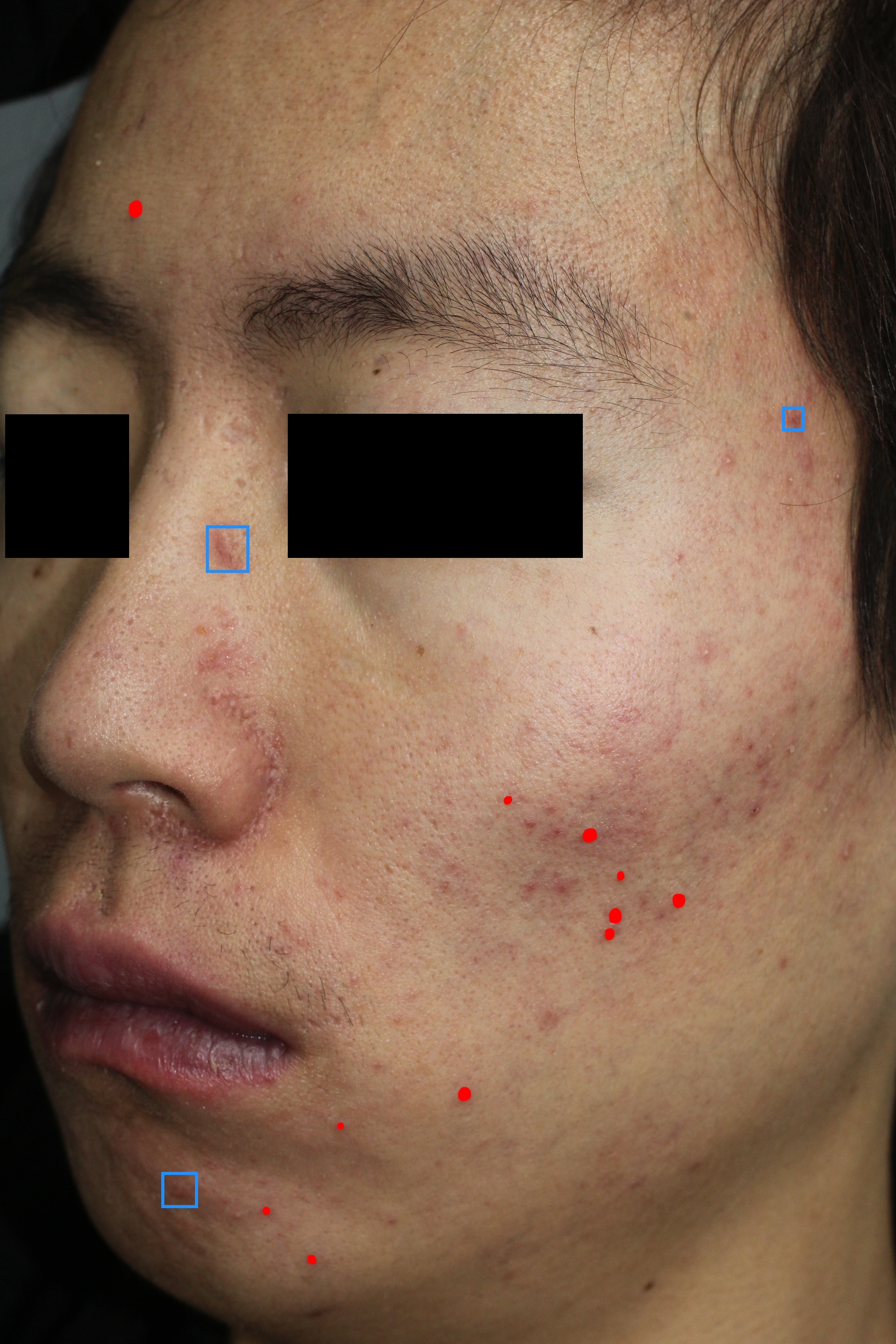} &
        \includegraphics[width=\linewidth]{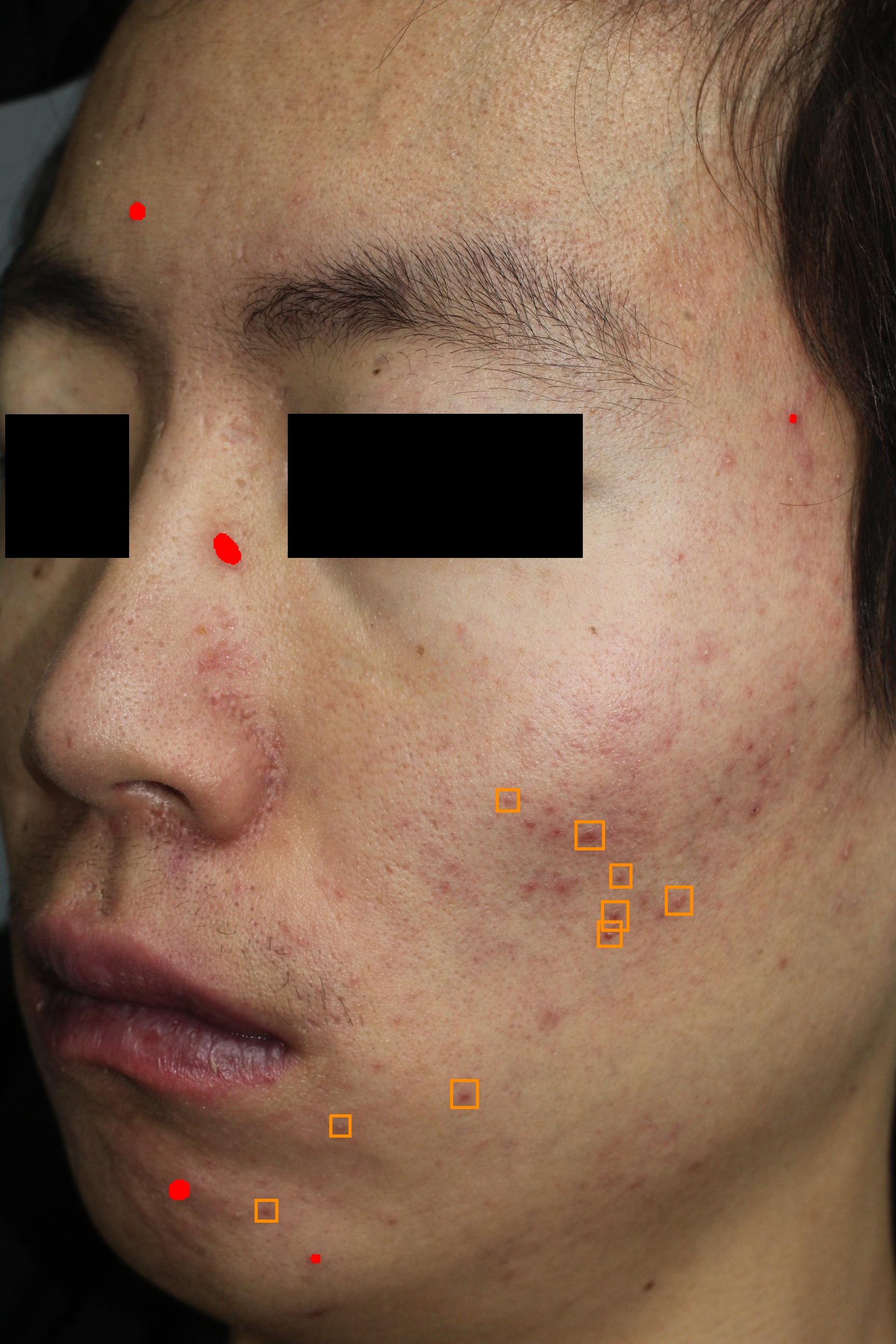} \\
        \fontsize{6pt}{7pt}\selectfont \textcolor{blue}{\textbf{Blue}}: PIH/scars/pores \textbf{(FP)} &
        \fontsize{6pt}{7pt}\selectfont \textcolor{orange}{\textbf{Orange}}: missed lesions \textbf{(FN)} &
        \fontsize{6pt}{7pt}\selectfont \textcolor{blue}{\textbf{Blue}}: PIH cluster \textbf{(FP)} &
        \fontsize{6pt}{7pt}\selectfont \textcolor{orange}{\textbf{Orange}}: missed papules \textbf{(FN)} \\ \bottomrule
    \end{tabular}
    \caption{\textbf{Failure case analysis of VL-AcneSeg.} Red regions denote acne masks. \textcolor{blue}{\textbf{Blue boxes}} mark false positives, predominantly PIH, scars and prominent pores; \textcolor{orange}{\textbf{orange boxes}} mark false negatives, typically small or low-contrast lesions.}
    \label{fig:fig8}
\end{figure}

The residual errors follow a consistent pattern (\textcolor{IEEEBLUE}{Fig.~\ref{fig:fig8}}). In (A) the cheek and chin carry active lesions, post-inflammatory hyperpigmentation and faded scars together, and the model both marks the residual features as lesions (blue boxes) and misses small papules embedded in densely pigmented skin (orange boxes). In (B) small inflammatory lesions are interspersed with widespread red marks, and the same two errors recur: subtle active lesions are overlooked and isolated non-acne marks elsewhere on the face are labelled as acne. Both modes arise from the same ambiguity, since active lesions and their residue share colour, shape and texture in clinical photographs. Confusions of this kind, rather than gross localization failures, account for most of the remaining error.

\subsection{Clinical Implementation}
\label{sec:iga}
\label{sec:clinical_implementation}
\begin{table}[t]
\centering
\caption{Correlation between the predicted lesion-to-face area ratio and IGA severity on the internal test set (22 visits from 16 patients). $p$ values compare each coefficient with that of the expert annotations using a test for dependent correlations sharing one variable; confidence intervals are obtained by patient-level cluster bootstrap.}
\label{tab:iga}
\footnotesize
\setlength{\tabcolsep}{3pt}
\resizebox{\columnwidth}{!}{
\begin{tabular}{lcccc}
\hline
\multirow{2}{*}{\textbf{Source of area ratio}} & \multicolumn{2}{c}{\textbf{Pearson}} & \multicolumn{2}{c}{\textbf{Spearman}} \\
\cline{2-5}
 & $r$ [95\% CI] & $p$ & $\rho$ [95\% CI] & $p$ \\ \hline
Expert annotation (GT) & 0.658 {\scriptsize [0.441, 0.807]} & --- & 0.710 {\scriptsize [0.541, 0.793]} & --- \\ \hline
\textbf{VL-AcneSeg (region)} & 0.719 {\scriptsize [0.477, 0.848]} & 0.430 & 0.684 {\scriptsize [0.485, 0.799]} & 0.719 \\
VL-AcneSeg (global) & 0.695 {\scriptsize [0.442, 0.833]} & 0.621 & 0.695 {\scriptsize [0.465, 0.843]} & 0.868 \\
EviVLM & 0.640 {\scriptsize [0.361, 0.796]} & 0.839 & 0.660 {\scriptsize [0.392, 0.809]} & 0.598 \\
CAT-Seg & 0.650 {\scriptsize [0.407, 0.797]} & 0.924 & 0.685 {\scriptsize [0.479, 0.812]} & 0.737 \\
SED & 0.630 {\scriptsize [0.279, 0.808]} & 0.770 & 0.645 {\scriptsize [0.372, 0.827]} & 0.526 \\
SAM & 0.650 {\scriptsize [0.320, 0.820]} & 0.932 & 0.572 {\scriptsize [0.299, 0.759]} & 0.187 \\
nnU-Net & 0.579 {\scriptsize [0.179, 0.793]} & 0.435 & 0.530 {\scriptsize [0.203, 0.776]} & 0.125 \\
Swin-UMamba$\dagger$ & 0.587 {\scriptsize [0.155, 0.801]} & 0.501 & 0.510 {\scriptsize [0.151, 0.766]} & 0.116 \\
PSPNet & 0.690 {\scriptsize [0.373, 0.863]} & 0.738 & 0.571 {\scriptsize [0.323, 0.755]} & 0.218 \\
U-Net & 0.630 {\scriptsize [0.336, 0.795]} & 0.722 & 0.628 {\scriptsize [0.376, 0.809]} & 0.372 \\
U-Net++ & 0.630 {\scriptsize [0.291, 0.844]} & 0.798 & 0.549 {\scriptsize [0.200, 0.790]} & 0.254 \\
TransUNet & 0.610 {\scriptsize [0.293, 0.782]} & 0.623 & 0.610 {\scriptsize [0.383, 0.771]} & 0.244 \\
DeepLabV3$+$ & 0.600 {\scriptsize [0.221, 0.808]} & 0.589 & 0.496 {\scriptsize [0.189, 0.728]} & 0.085 \\
CE-Net & 0.590 {\scriptsize [0.229, 0.779]} & 0.512 & 0.553 {\scriptsize [0.240, 0.778]} & 0.225 \\
SegFormer & 0.540 {\scriptsize [0.182, 0.753]} & 0.286 & 0.527 {\scriptsize [0.179, 0.780]} & 0.219 \\
Swin-UNet & 0.400 {\scriptsize [0.003, 0.617]} & 0.057 & 0.462 {\scriptsize [0.110, 0.722]} & 0.102 \\ \hline
\end{tabular}
}
\end{table}

To assess clinical relevance beyond pixel-level metrics, we correlated the lesion-to-face area ratio of each model with Investigator's Global Assessment (IGA) severity on the internal test set, pooling the views of a visit so that each of the 22 visits contributes one observation (\textcolor{IEEEBLUE}{Table~\ref{tab:iga}}). The area ratio derived from VL-AcneSeg tracks IGA at the level of the expert annotations themselves (Pearson 0.719 versus 0.658; Spearman 0.684 versus 0.710); the direction of this comparison reverses between the two coefficients and neither difference is significant (\textit{p} = 0.430 and 0.719), so the predicted and reference ratios cannot be distinguished in their agreement with IGA on this sample.

Comparing methods, no coefficient differs significantly from that of the reference annotations at this sample size. On Spearman, however, the six highest values belong to the reference annotations and to the five vision-language configurations, with the convolutional and Transformer architectures below them even where their pixel-level accuracy is comparable. Within this group the global prompt, our primary setting, correlates as well as the region-level one, as expected for a measure defined over the whole face; region-level prompts instead yield a separate estimate for each facial area. A high coefficient alone, however, does not imply an accurate area estimate: U-Net reaches a moderate value because its inflated areas still scale with lesion burden, whereas VL-AcneSeg reaches a comparable coefficient with precision and recall balanced, so its estimates track severity without a systematic offset. These results support the premise that lesion area measured from a segmentation can serve as a quantitative severity measure.

This is also where an area-based measure differs from counting. A count treats every lesion as one unit irrespective of extent, so it saturates precisely where severity is greatest: confluent papules that have merged into a plaque contribute the same value as a few discrete ones, and a lesion that shrinks under treatment without resolving contributes the same value throughout. An area ratio varies continuously with both, which is the property area-based assessment relies on.

A Dice score near 0.53 is modest by the standards of organ segmentation, but the quantity entering the severity measure is the aggregate lesion area rather than the boundary of each lesion, and a Dice score at this level is sufficient for that area to track IGA as closely as the reference annotations do. The difficulty of the task is also reflected in human performance: in a reader study on this cohort, eight dermatologists and residents reached a median lesion-detection F1 of 0.31 for inflammatory lesions from images alone \cite{kim2023automated}. Applications requiring precise per-lesion delineation would need higher accuracy than we report here.

\subsection{Deployment Considerations}
The framework runs in two settings. Under the global prompt no region information is required, which is why we report it as the primary setting. Under region-level prompting a clinician or the patient indicates which facial areas carry lesions---a coarse judgement, not the location of individual lesions---which raises accuracy and additionally gives a separate estimate for each area. The perturbation analysis indicates how accurate such prompts need to be. Withholding prompts for lesion-bearing regions falls below the global-prompt baseline at roughly 15\% omission, whereas issuing prompts for lesion-free regions stays above it throughout the range examined. In practice, therefore, over-inclusive prompting is safe, whereas leaving affected regions unmarked is what degrades the result.

\begin{table}[t]
\centering
\caption{Computational cost, measured on a single NVIDIA A6000 at the input resolution used throughout. For VL-AcneSeg, \textit{all regions} denotes prompts issued for all fifteen facial regions; the tiled CLIP encoding is shared across prompts.}
\label{tab:cost}
\footnotesize
\setlength{\tabcolsep}{4pt}
\begin{tabular}{lccc}
\hline
 & \textbf{Params (M)} & \textbf{FLOPs (G)} & \textbf{Time (ms)} \\ \hline
\textbf{VL-AcneSeg} (global prompt) & 432 & 1{,}417 & 858 \\
\textbf{VL-AcneSeg} (all regions) & 432 & 1{,}676 & 1{,}536 \\
SAM (ViT-L) & 312 & 1{,}312 & 594 \\
EviVLM & 130 & 567 & 176 \\
Swin-UMamba$\dagger$ & 27 & 25 & 64 \\
U-Net & 17 & 428 & 74 \\ \hline
\end{tabular}
\end{table}

VL-AcneSeg is slower than most baselines (\textcolor{IEEEBLUE}{Table~\ref{tab:cost}}), which follows from the tiled high-resolution encoding on which the accuracy also depends. In absolute terms, however, one image takes 858\,ms under the global prompt and at most 1.54\,s with all fifteen regions, so cost is not a constraint at either setting. The tiled encoding accounts for roughly 83\% of the computation but is shared across prompts, so each additional region prompt adds only 18.5\,GFLOPs, or about 48\,ms.

\subsection{Limitations and Future Directions}
Several limitations should be acknowledged. First, region-level prompting presupposes knowledge of which facial areas contain lesions; we therefore report the global prompt, which requires none, as our primary setting. With a closed set of fifteen regions the pretrained text prior is also not fully exploited: unseen synonyms are resolved (\textcolor{IEEEBLUE}{Table~\ref{tab:table5}}), but that capacity makes no difference to a deployment in which the vocabulary is fixed in advance.

Second, all internal and external data represent Korean populations, so the evidence does not extend to other ethnicities, skin tones or imaging devices, and claims of worldwide applicability would be unwarranted. This constraint is not particular to our work: the two public acne datasets are likewise drawn from a single ethnic group, and no published acne algorithm has yet undergone prospective clinical validation \cite{traini2025artificial}. Fitzpatrick phototype was not recorded in the clinical dataset and is not distributed with the external collection, so performance could not be stratified by skin type; nor could it be stratified by severity, since the 22 graded visits leave strata too small for stable estimates. The same sample size widens the intervals around the smaller between-method differences, so their magnitude is estimated less precisely than their direction. Constructing pixel-level acne datasets requires substantial dermatological effort, and evaluation on multi-ethnic, multi-site data remains the most important direction for establishing broader applicability.

Third, Dice and IoU remain moderate relative to other medical segmentation tasks, and the residual error is dominated by the ambiguity between active lesions and their own residue described in \textcolor{IEEEBLUE}{Section~\ref{sec:failure}}. Fourth, the correlation with IGA does not by itself show that the method is ready for clinical use; establishing that would require a prospective study against expert assessment.

Finally, inter-annotator agreement was not quantified. The annotation followed an adjudication design in which a single board-certified dermatologist fixed the final label set, so agreement between readers is not the quantity that defines the reference standard here; the underlying lesion markings were moreover produced for an earlier study \cite{kim2023automated}, and the drafts preceding adjudication were not retained as separate outputs. Quantifying observer variability on this task would help establish the practical ceiling for pixel-level acne segmentation and remains an objective for future work.

Several of these limitations suggest concrete next steps. The confusion is between an active lesion and its own residue, which are hard to separate in the spatial domain because they share colour, shape and texture; features that expose properties the RGB image does not make explicit, such as a frequency-domain representation, may provide evidence that spatial appearance alone does not. Our annotations distinguish five lesion types, and we plan to extend the model to a multi-class objective, which would also allow papules, pustules and nodules to be segmented jointly. Prompts that vary in what they describe---lesion type, extent or a clinician's own phrasing---would put the text encoder to work, and are a natural direction once multi-class supervision is in place. More broadly, the reformulation itself is not specific to acne: wherever lesions are small and numerous but distributed over an anatomy that is stable and nameable, text can supply the same kind of spatial constraint.

\section{Conclusion}
We introduced \textbf{VL-AcneSeg}, a multimodal framework for inflammatory acne lesion segmentation. Where open-vocabulary methods use text to specify which class to segment, we use it to specify where the target may appear, so that the text channel supplies a spatial rather than a categorical constraint---a formulation suited to a task with a single lesion class distributed over a nameable anatomy. Because region-level prompts presuppose knowledge of lesion locations, we report a single global prompt as the primary setting and treat region-level prompting as an oracle upper bound.

Across internal and external datasets VL-AcneSeg attains the highest Dice and IoU among the compared methods, including vision-language baselines that are themselves given region-level prompts, and retains this under domain shift where general-purpose architectures transfer less well. Controlled substitutions indicate that the prompt is read as a reference to a particular area rather than as an undifferentiated signal, although on a closed region vocabulary a learned region vector is mechanistically equivalent to a text embedding. The lesion areas the model produces track clinical IGA severity as closely as the expert annotations do, supporting area-based severity assessment as an alternative to lesion counting. We believe VL-AcneSeg provides a practical foundation for automated acne management.

\section*{Conflict of Interests}
The authors declare that they have no conflict of interest.

\bibliographystyle{IEEEtran}
\bibliography{references}

\clearpage
\setcounter{section}{0}
\setcounter{figure}{0}
\setcounter{table}{0}
\setcounter{equation}{0}
\renewcommand{\thesection}{S\arabic{section}}
\renewcommand{\thefigure}{S\arabic{figure}}
\renewcommand{\thetable}{S\arabic{table}}
\renewcommand{\theequation}{S\arabic{equation}}

\section*{Supplementary Material}

\section{Dataset Details}
\label{sec:supp_data}

\subsection{Internal cohort}
\noindent The internal cohort comprises 258 patients (38.4\% male; mean age 22.7 $\pm$ 5.97 years) who visited the Department of Dermatology, Seoul National University Hospital between February and July 2020. Photographs were acquired with Canon EOS 550D and Nikon D7100 cameras under standardized conditions. Fitzpatrick phototype was not routinely recorded in the clinical dataset; all participants were of Korean ethnicity, a population in which phototypes III and IV predominate. 

\subsection{IGA grading protocol}
\noindent Investigator's Global Assessment scores were assigned retrospectively at the visit level: three board-certified dermatologists independently reviewed all images acquired at a given visit and graded that visit on the standard five-point scale (0, clear; 1, almost clear; 2, mild; 3, moderate; 4, severe). The final grade was determined by majority voting; when the three graders assigned three different grades, the case was resolved through joint discussion until consensus was reached. Among the 22 visits in the internal test set, 3, 7, 8 and 4 visits were graded as IGA 1, 2, 3 and 4 respectively, and no visit was graded as IGA 0.

\subsection{External subset construction}
\noindent The AI Hub Korean Skin Condition Measurement Dataset includes 13{,}936 facial images, 84{,}688 skin condition measurement records and 125{,}424 labeled data entries. Because the collection contains many images without visible inflammatory lesions, we first applied an acne detection model trained on the public ACNE04 dataset as a screening step. A deliberately low confidence threshold (0.3) was used so that the screening favoured recall, retaining images with even weak detection responses. From this candidate pool we randomly sampled 100 images captured with digital cameras in a controlled environment and 100 images captured with smartphones under uncontrolled conditions. Each sampled image was then reviewed by the dermatologist who adjudicated the internal set, and those without confirmed inflammatory acne were excluded, yielding the final external test sets of 71 and 56 images. Consistent with the permissive screening threshold, only 71\% and 56\% of the sampled images were confirmed to contain inflammatory acne. A residual sampling bias nevertheless remains, restricted to images whose lesions the screening model failed to detect even at this low threshold, and this should be considered when interpreting the external validation results.

\subsection{External imaging conditions and metadata}
\noindent Because the AI Hub collection was compiled for general facial skin analysis rather than from an acne patient population, the two external subsets differ from the internal cohort in lesion burden. They also differ in the type of imaging shift they probe. The controlled subset was acquired with digital cameras under a standardized setup comparable to that of the internal cohort and therefore mainly reflects a shift in population and acquisition site rather than in imaging device. The real-world subset was acquired with smartphone cameras and includes images taken with both front- and rear-facing cameras, introducing additional variability in resolution, field of view and colour processing. The AI Hub distribution does not specify the individual camera models, and it provides neither participant demographics nor Fitzpatrick phototypes; these characteristics therefore cannot be reported for the external subsets.

\section{Text Prompt Details}
\label{sec:supp_prompt}

\begin{table}[t]
\centering
\caption{Anatomical synonyms used in the unseen-prompt experiment. Each training region name was replaced at inference by one of the two alternatives, chosen at random; none appears in training.}
\label{tab:supp_synonyms}
\footnotesize
\setlength{\tabcolsep}{3pt}
\begin{tabular}{lll}
\hline
\textbf{Seen (training)} & \textbf{Unseen 1} & \textbf{Unseen 2} \\ \hline
forehead & supraorbital area & frontal region \\
left cheek & left malar region & left buccal area \\
right cheek & right malar region & right buccal area \\
chin & mentum & mental region \\
nose & nasal region & nasal bridge \\
left temple & left pterion area & left temporal region \\
right temple & right pterion area & right temporal region \\
glabella & interciliary space & intercilium \\
left eyebrow & left supraorbital ridge & left superciliary arch \\
right eyebrow & right supraorbital ridge & right superciliary arch \\
left eye & left periorbital region & left orbital area \\
right eye & right periorbital region & right orbital area \\
lip & labial region & vermilion area \\
upper lip & upper labial region & infranasal area \\
neck & cervical region & anterior cervical area \\ \hline
\end{tabular}
\end{table}

\begin{table}[t!]
\centering
\caption{Ablation study on different text prompt strategies for VL-AcneSeg in internal data. Best values are in \textbf{bold} and second-best are \underline{underlined}.}
\label{tab:supp_promptstrategy}
\footnotesize
\begin{tabular}{lccc}
\hline
\textbf{Metrics} & \textbf{Region Prompt} & \textbf{Level Prompt} & \textbf{Num Prompt} \\ \hline
IoU          & \textbf{0.3602} & 0.3521 & 0.3594 \\
Dice         & \textbf{0.5296} & 0.5208 & 0.5288 \\
Precision    & \textbf{0.5202} & 0.4889 & 0.4940 \\
Recall       & 0.5393 & \underline{0.5572} & \textbf{0.5688} \\ \hline
\end{tabular}
\end{table}

\noindent Table~\ref{tab:supp_synonyms} lists the anatomical synonyms used in the unseen-prompt experiment reported in the main paper; none of the alternatives appears during training. Table~\ref{tab:supp_promptstrategy} compares three prompt contents on the internal dataset: \textbf{Region} supplies spatial guidance only, \textbf{Level} adds a qualitative severity descriptor, and \textbf{Num} adds an exact lesion count per region. Adding count descriptors raises recall by widening the search space but lowers precision, reflecting a trade-off between spatial guidance and quantitative prompt engineering.

\section{Localization under Different Patch Resolutions}
\label{sec:supp_patch}

\begin{figure}[t]
    \centering
    \setlength{\tabcolsep}{1pt} 
    
    \begin{tabular}{ccc}
        \includegraphics[width=0.31\columnwidth]{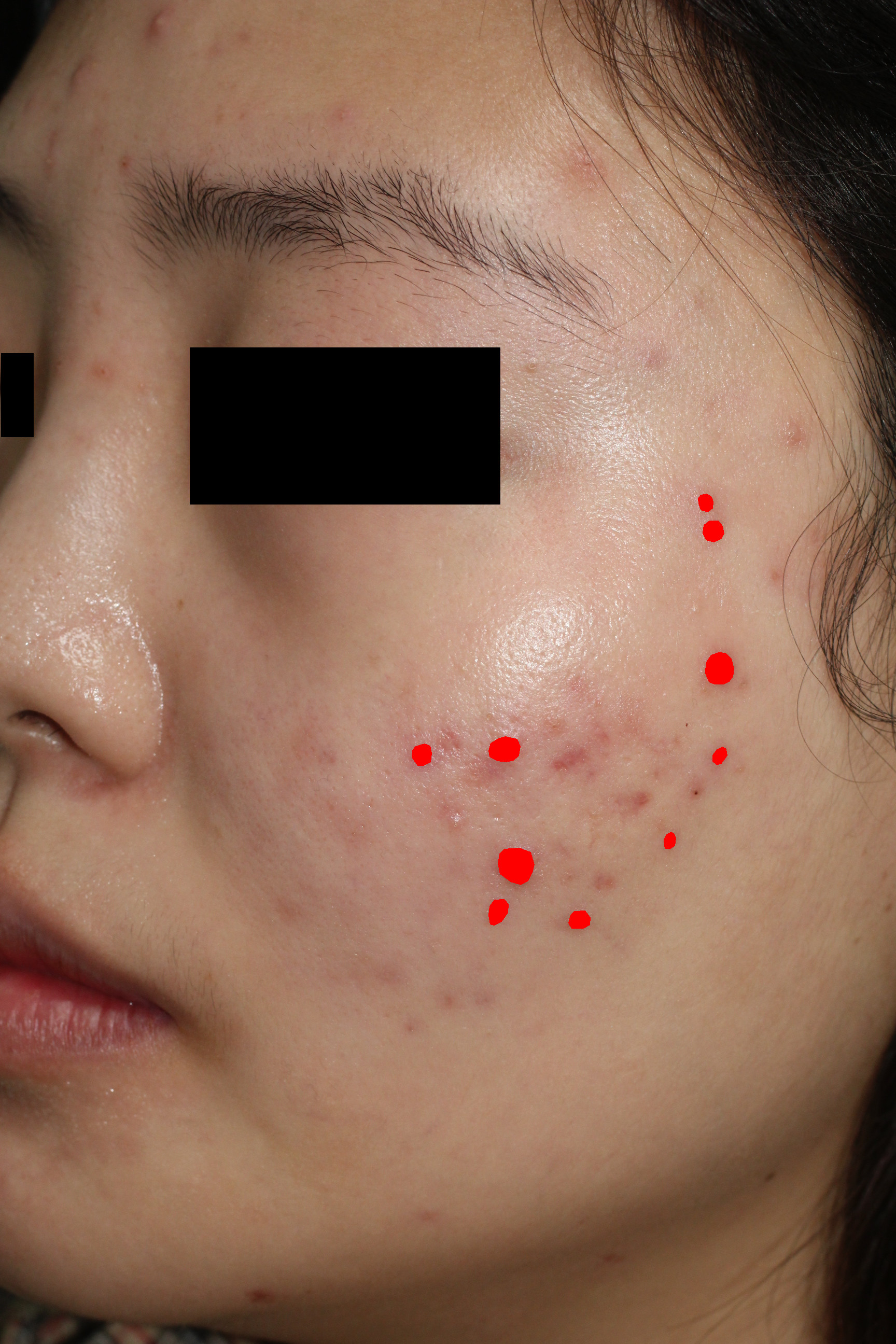} & 
        \includegraphics[width=0.31\columnwidth]{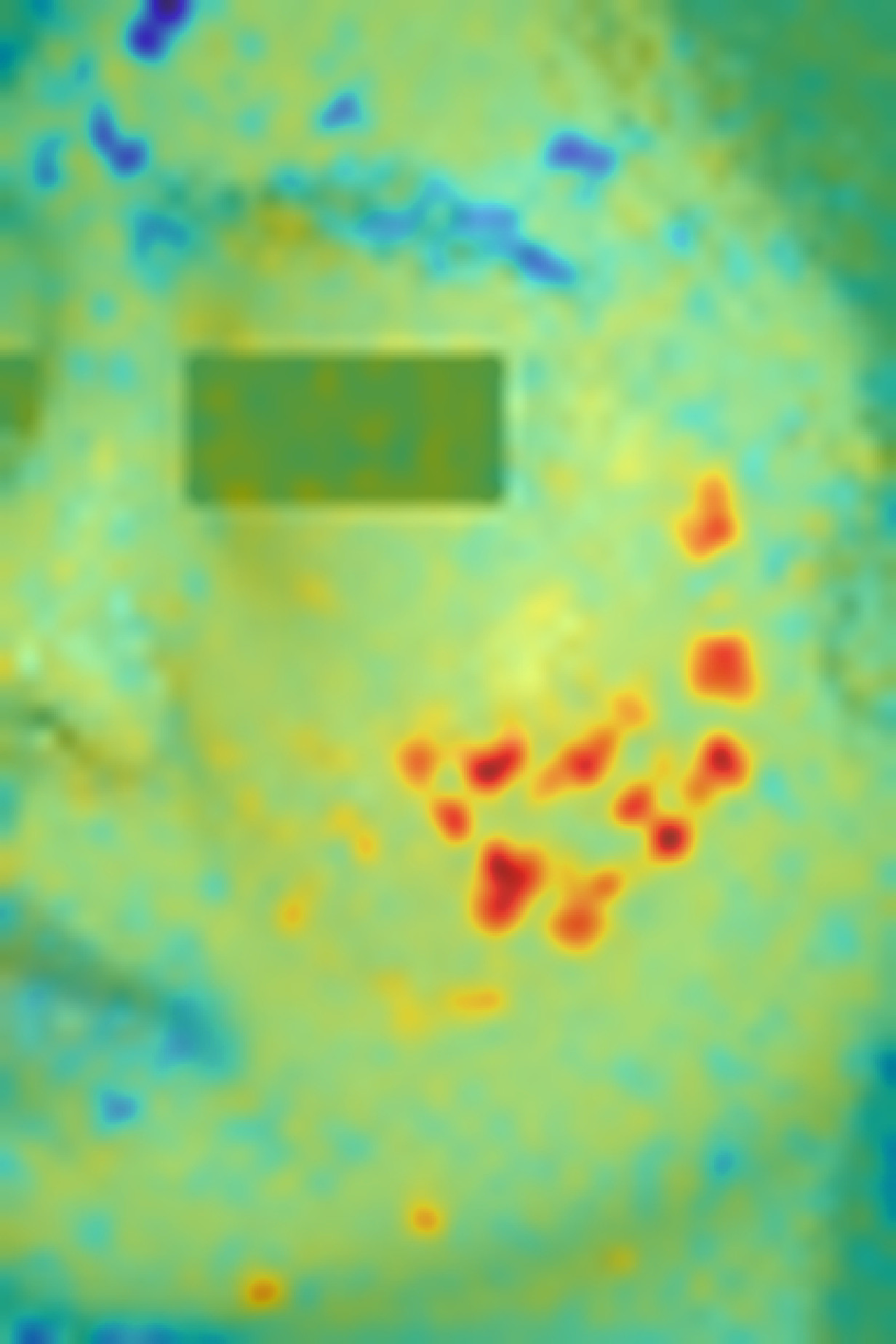} & 
        \includegraphics[width=0.31\columnwidth]{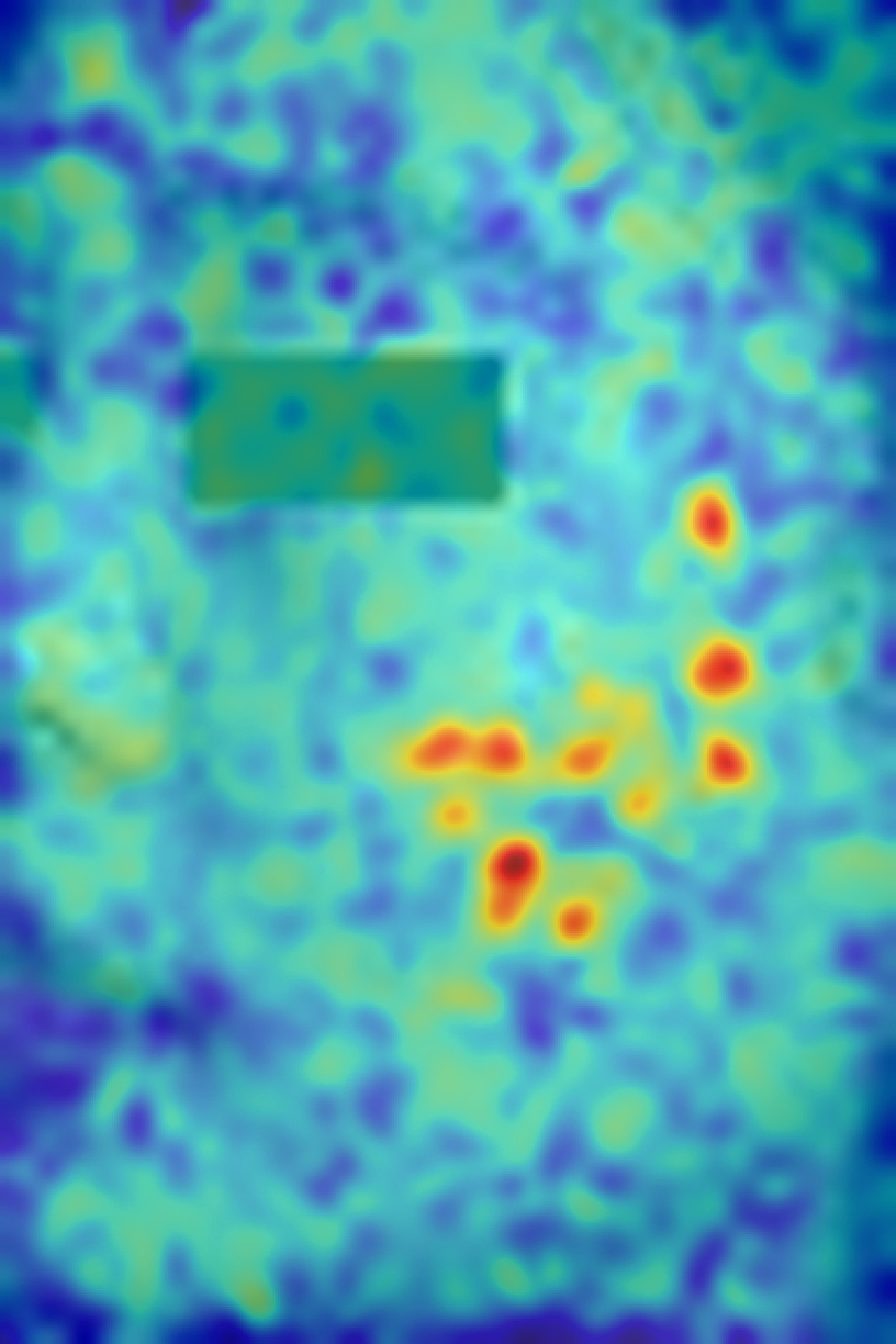} \\
        \footnotesize (a) GT & \footnotesize (b) $4 \times 3$ & \footnotesize (c) $3 \times 2$ \\ 
        \noalign{\vspace{1pt}} 
        
        \includegraphics[width=0.31\columnwidth]{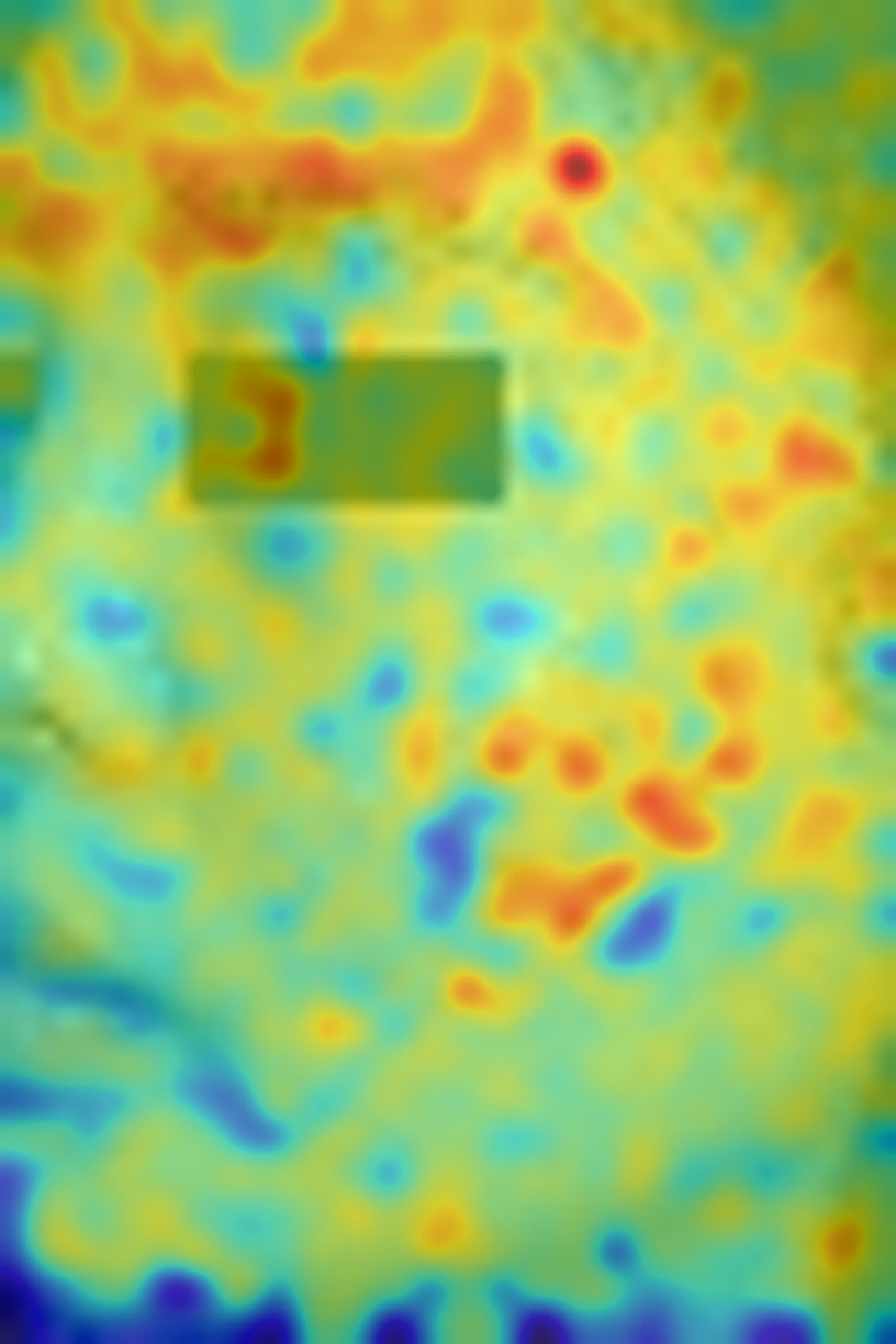} & 
        \includegraphics[width=0.31\columnwidth]{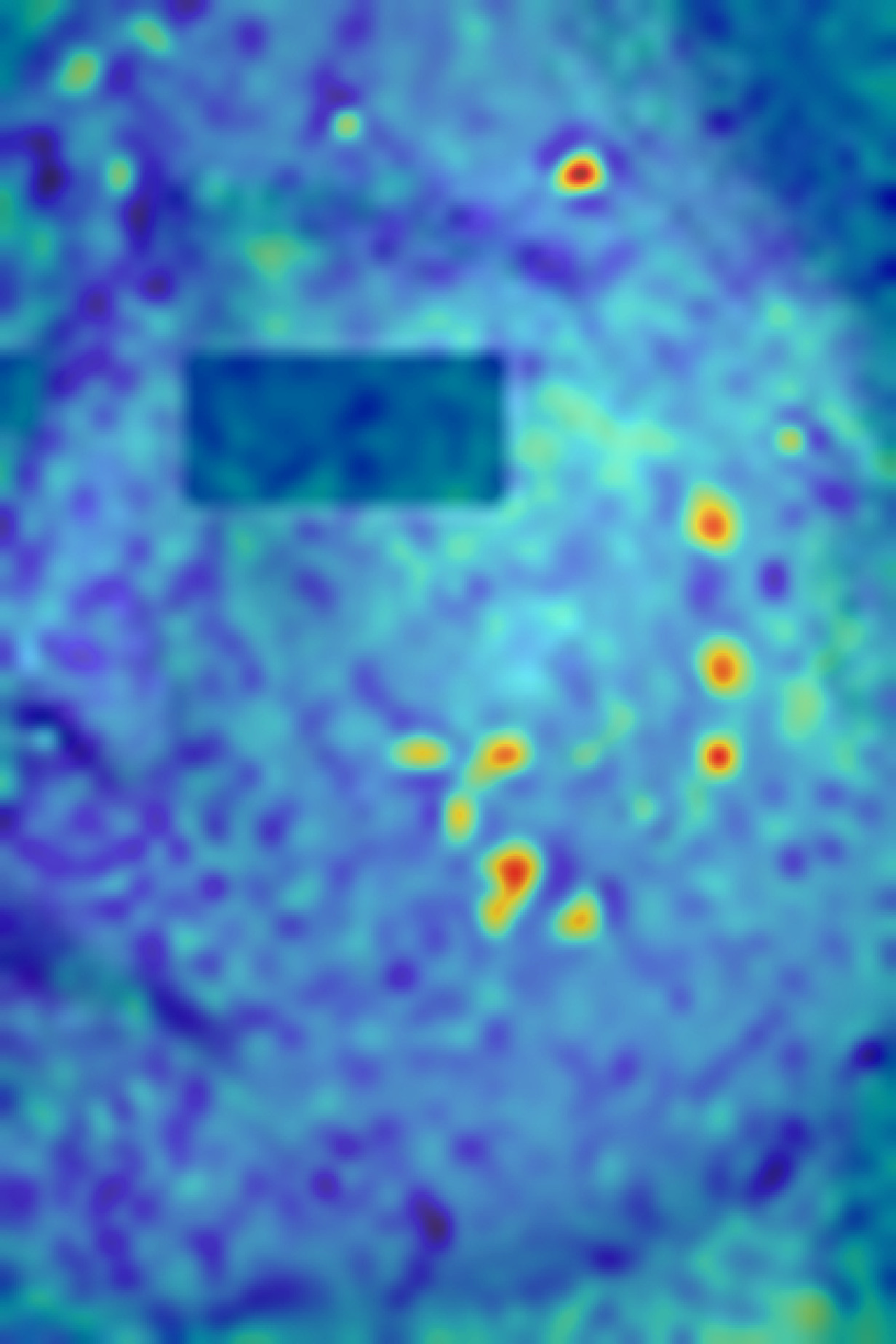} & \\
        \footnotesize (d) $2 \times 1$ & \footnotesize (e) Global Prompt & 
    \end{tabular}

    \caption{\textbf{Ablation study on patch granularity and regional textual cues.} Finer patch division improves localization precision. The model still localizes lesions under a global text prompt alone (e).}
    \label{fig:supp_patch}
\end{figure}

\noindent Localization maps for the three tiling configurations evaluated in Section~\ref{sec:ablation} of the main paper. Coarser tiling spreads activation over confounding skin features, whereas the finer division keeps the response on the lesions themselves.

\section{Qualitative Analysis of Localization in Smartphone Image}
\label{sec:appendix_localization}

\begin{figure}[ht]
    \centering
    \setlength{\tabcolsep}{2pt}
    \begin{tabular}{ccc}
        \includegraphics[width=0.31\linewidth]{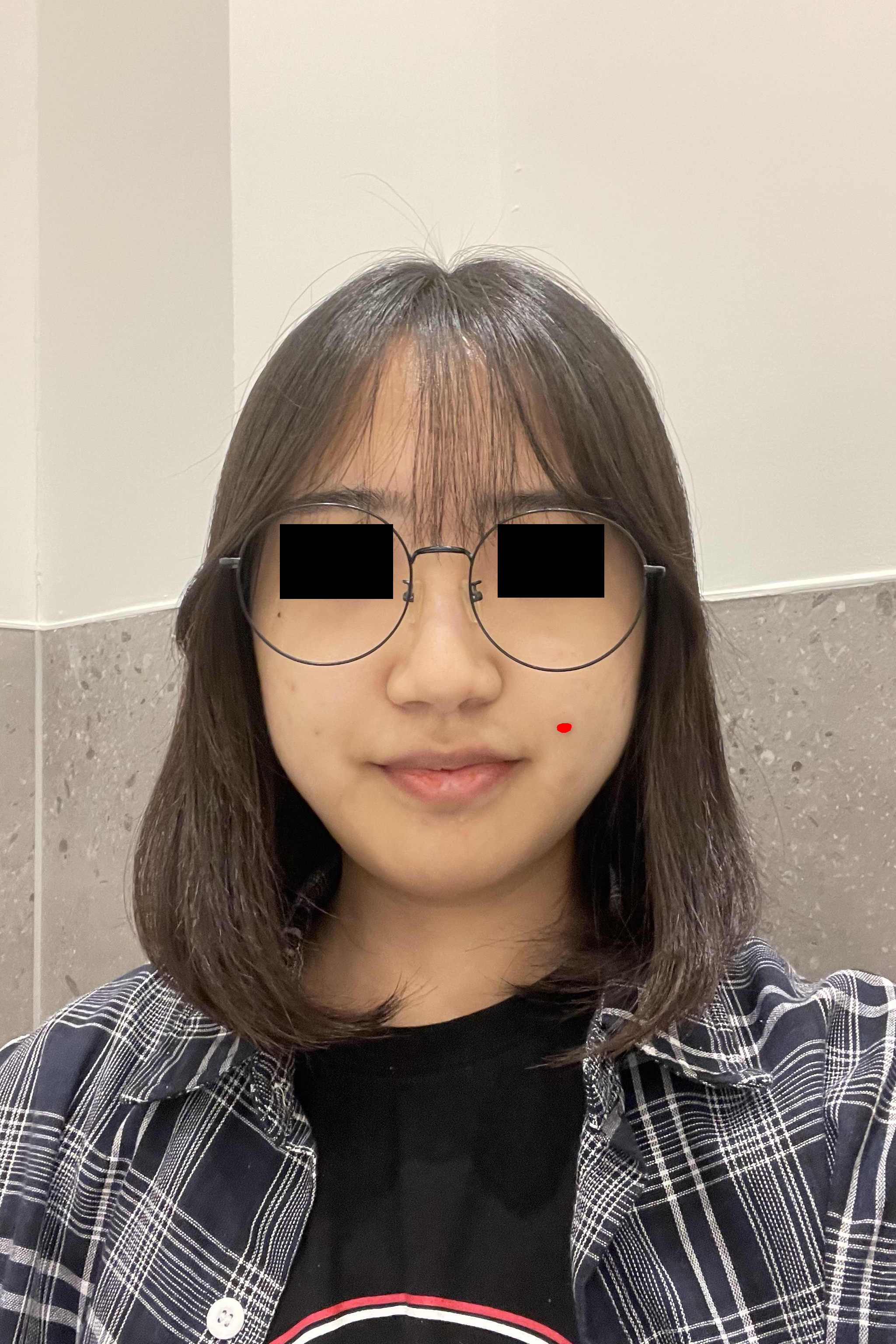} &
        \includegraphics[width=0.31\linewidth]{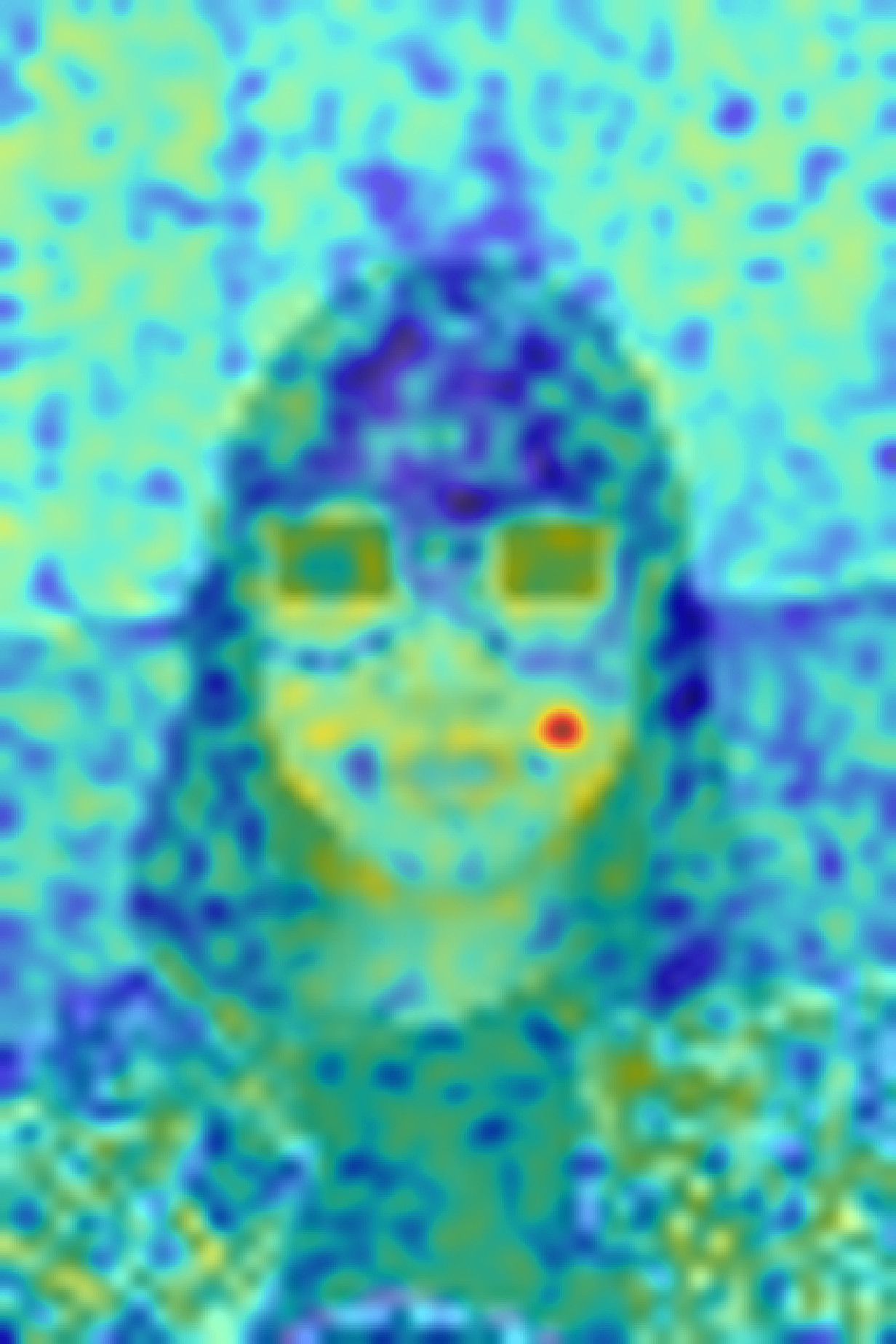} &
        \includegraphics[width=0.31\linewidth]{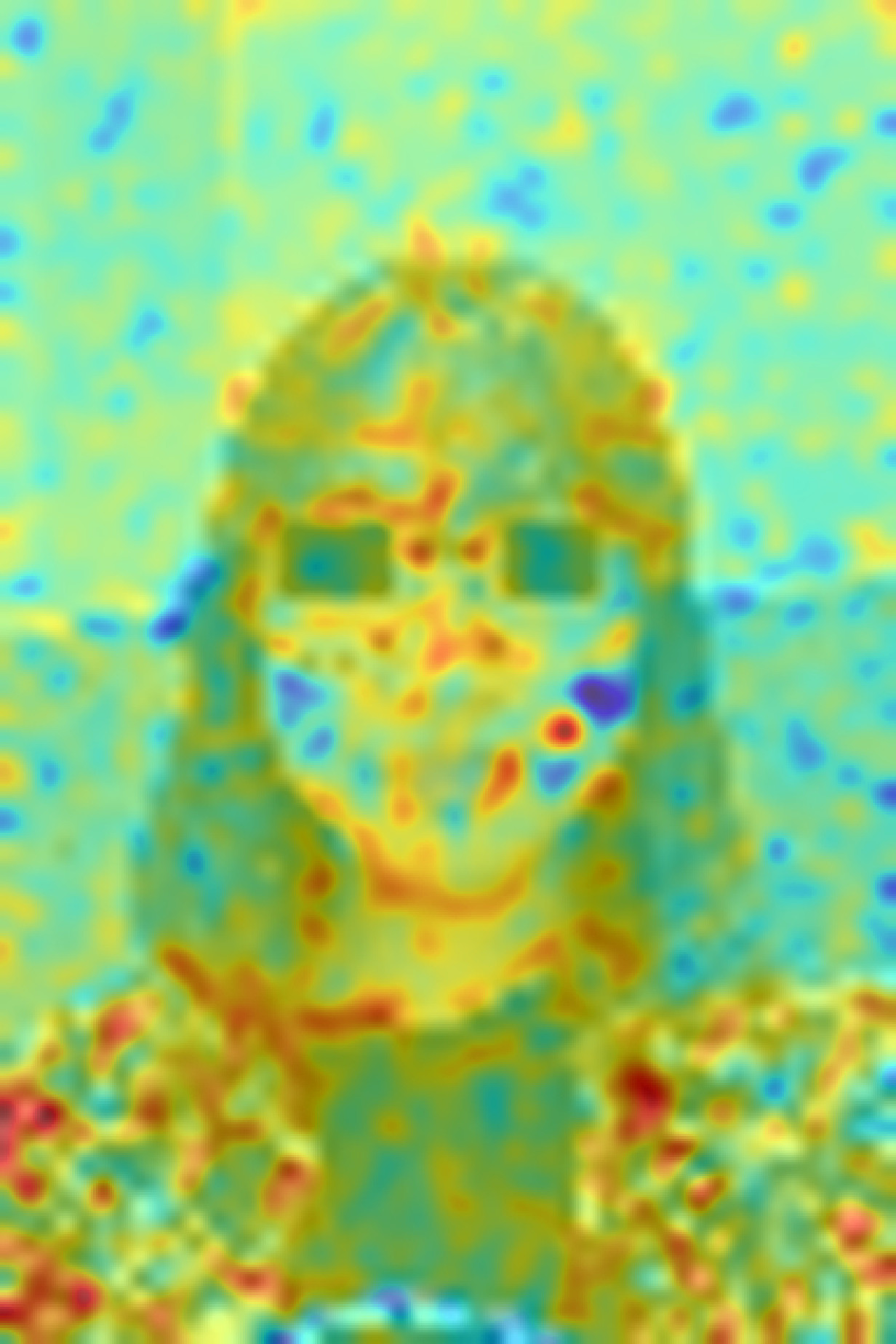} \\
        \footnotesize (a) Ground Truth & \footnotesize (b) Full Model & \footnotesize (c) Global Prompt
    \end{tabular}
    \caption{\textbf{Localization heatmaps in a smartphone environment.} (b) The full model identifies lesion-suspected areas with high precision and sharpness. (c) Although the global text prompt configuration exhibits broader localization extending beyond facial boundaries, it still captures the actual lesion sites.}
    \label{fig:smartphone_localization_ablation}
\end{figure}

\noindent Figure~\ref{fig:smartphone_localization_ablation} shows localization maps for a smartphone image under the two prompting conditions. With region-level prompts the response is concentrated on the lesion sites. With a global prompt the activation is broader and occasionally extends beyond the facial boundary, but the lesion sites are still recovered, so the model remains usable where region information is unavailable.

\section{Patch Resolution on External Datasets}
\label{sec:appendix_ablation}

\begin{table}[ht]
    \centering
    \caption{Effect of patch tiling resolution on the two external datasets. Component-wise ablations on the external datasets are reported in the main paper. Best values are in \textbf{bold} and second-best are \underline{underlined}.}
    \label{tab:appendix_ablation_final}
    \renewcommand{\arraystretch}{1.2}
    \resizebox{\linewidth}{!}{
        \begin{tabular}{l|cccccc|cccccc}
            \hline
            \multirow{3}{*}{\textbf{Ablation Model}} & \multicolumn{12}{c}{\textbf{External Datasets (Evaluation Only)}} \\
            \cline{2-13}
            & \multicolumn{6}{c|}{\textbf{Controlled Setting (1)}} & \multicolumn{6}{c}{\textbf{Real-world Setting (2)}} \\
            \cline{2-13}
            & IoU & Dice & Precision & Recall & TP/img & FP/img & IoU & Dice & Precision & Recall & TP/img & FP/img \\
            \hline
            \textbf{VL-AcneSeg (Full Model)} & \textbf{0.4232} & \textbf{0.5948} & \textbf{0.6110} & \textbf{0.5794} & 1.64 & 0.60 & \textbf{0.3044} & \textbf{0.4667} & \textbf{0.4882} & \underline{0.4471} & 1.50 & \textbf{0.82} \\
            $3 \times 2$ Patches & \underline{0.3812} & \underline{0.5520} & \underline{0.5664} & \underline{0.5383} & 1.83 & \textbf{0.79} & \underline{0.2443} & \underline{0.3927} & \underline{0.3787} & 0.4077 & 1.11 & \underline{0.66} \\
            $2 \times 1$ Patches & 0.3019 & 0.4638 & 0.4593 & 0.4683 & 1.58 & \underline{0.77} & 0.1213 & 0.2164 & 0.1354 & \textbf{0.5383} & 0.89 & 1.43 \\
            \hline
        \end{tabular}
    }
\end{table}

\noindent Table~\ref{tab:appendix_ablation_final} shows that the gain from finer tiling observed on the internal dataset also holds on both external datasets.
\section{Confidence Intervals for Between-Method Differences}
\label{sec:supp_diffci}

\begin{table}[t]
\centering
\caption{Differences in IoU and Dice between VL-AcneSeg and each baseline on the internal test set, with 95\% confidence intervals from paired patient-level cluster bootstrap over the 16 test patients. A positive value favours the proposed method. Vision-language baselines receive region-level prompts and are therefore compared against the region-level configuration; the remaining methods take no text input and are compared against the global prompt.}
\label{tab:supp_diffci}
\scriptsize
\setlength{\tabcolsep}{3pt}
\resizebox{\columnwidth}{!}{%
\begin{tabular}{lcccc}
\hline
\textbf{Baseline} & \textbf{$\Delta$IoU} \textbf{[95\% CI]} & \textbf{\textit{p}} & \textbf{$\Delta$Dice [95\% CI]} & \textbf{\textit{p}} \\ \hline
\multicolumn{5}{l}{\textit{Vision-language methods, compared against VL-AcneSeg (region)}} \\
EviVLM & $+$0.031 [0.001, 0.069] & 0.048 & $+$0.034 [0.001, 0.080] & 0.046 \\
CAT-Seg & $+$0.069 [0.012, 0.136] & 0.013 & $+$0.079 [0.011, 0.166] & 0.015 \\
SED & $+$0.042 [0.003, 0.097] & 0.034 & $+$0.047 [0.003, 0.112] & 0.038 \\
\hline
\multicolumn{5}{l}{\textit{Remaining methods, compared against VL-AcneSeg (global)}} \\
SAM & $+$0.019 [$-$0.037, 0.078] & 0.580 & $+$0.021 [$-$0.041, 0.091] & 0.592 \\
Swin-UMamba$\dagger$ & $+$0.053 [0.025, 0.076] & 0.005 & $+$0.061 [0.018, 0.085] & 0.005 \\
TransUNet & $+$0.078 [0.021, 0.144] & 0.005 & $+$0.092 [0.023, 0.182] & 0.005 \\
CE-Net & $+$0.119 [0.065, 0.170] & $<$0.001 & $+$0.146 [0.076, 0.211] & $<$0.001 \\
Swin-UNet & $+$0.132 [0.091, 0.170] & $<$0.001 & $+$0.163 [0.110, 0.217] & $<$0.001 \\
SegFormer & $+$0.136 [0.090, 0.177] & $<$0.001 & $+$0.168 [0.109, 0.228] & $<$0.001 \\
nnU-Net & $+$0.150 [0.110, 0.185] & $<$0.001 & $+$0.188 [0.136, 0.234] & $<$0.001 \\
PSPNet & $+$0.161 [0.132, 0.185] & $<$0.001 & $+$0.203 [0.172, 0.230] & $<$0.001 \\
DeepLabV3$+$ & $+$0.160 [0.131, 0.194] & $<$0.001 & $+$0.202 [0.164, 0.253] & $<$0.001 \\
U-Net++ & $+$0.183 [0.121, 0.236] & $<$0.001 & $+$0.236 [0.158, 0.311] & $<$0.001 \\
U-Net & $+$0.199 [0.165, 0.229] & $<$0.001 & $+$0.260 [0.225, 0.292] & $<$0.001 \\
\hline
\end{tabular}}
\end{table}

\noindent The main paper reports confidence intervals for each method separately, which do not indicate whether two methods differ, since the same patients contribute to both estimates. Table~\ref{tab:supp_diffci} therefore reports the difference itself, resampled in pairs so that the correlation between the two methods is preserved. Every difference is significant at the 0.05 level except that of SAM, whose interval is the only one to include zero; as reported in the main paper, that comparison was examined further by retraining both configurations with three seeds. The width of the intervals reflects the size of the test set: the differences against the conventional architectures are estimated to within a few hundredths of an IoU point, whereas those against the vision-language baselines, where the margin is smaller, are correspondingly less precise.

\section{Additional Evaluation Metrics}
\label{sec:supp_metrics}

\noindent This section reports two metrics that address aspects the main comparison leaves open. The first is small-lesion performance at the lesion level. Acne lesions vary widely in size, and the pixel-level metrics of the main paper are dominated by the larger ones simply because they contribute more pixels; a method could therefore score well while missing most of the small lesions. We isolate the smallest quartile and count lesions rather than pixels, so that each lesion carries equal weight. The second is the area under the precision--recall curve. All results in the main paper use a binarization threshold of 0.5, fixed for every method without per-model tuning, and AUPRC is computed from the probability maps before that threshold is applied, so it indicates whether the reported ordering depends on that choice.

\begin{table}[t]
\centering
\caption{Additional metrics on the internal test set. \textit{Small lesions} are those in the lowest quartile of the lesion-area distribution over the whole annotated set, corresponding to an area below 500\,px$^2$; recall, precision and F1 are computed over these lesions at the lesion level. AUPRC is the area under the precision--recall curve, computed from the predicted probability maps and therefore independent of the binarization threshold. Rows are ordered by small-lesion F1. Best values are in \textbf{bold} and second-best are \underline{underlined}.}
\label{tab:supp_metrics}
\footnotesize
\setlength{\tabcolsep}{4pt}
\begin{tabular}{lcccc}
\hline
\multirow{2}{*}{\textbf{Model}} & \multicolumn{3}{c}{\textbf{Small lesions} ($\leq 500$\,px$^2$)} & \multirow{2}{*}{\textbf{AUPRC}} \\
\cline{2-4}
 & Recall & Precision & F1 & \\ \hline
\textbf{VL-AcneSeg (region)} & 0.494 & \underline{0.451} & \textbf{0.472} & \underline{0.4138} \\
\textbf{VL-AcneSeg (global)} & 0.492 & 0.432 & \underline{0.460} & 0.4068 \\
SAM & 0.457 & 0.432 & 0.444 & 0.3400 \\
SED & 0.431 & 0.426 & 0.428 & \textbf{0.4167} \\
EviVLM & 0.453 & 0.405 & 0.428 & 0.4037 \\
CAT-Seg & 0.358 & \textbf{0.466} & 0.405 & 0.3922 \\
DeepLabV3$+$ & 0.381 & 0.392 & 0.386 & 0.3958 \\
CE-Net & 0.420 & 0.275 & 0.332 & 0.1655 \\
Swin-UMamba$\dagger$ & \underline{0.554} & 0.214 & 0.309 & 0.3963 \\
PSPNet & 0.503 & 0.208 & 0.294 & 0.3261 \\
U-Net++ & 0.531 & 0.194 & 0.284 & 0.1582 \\
Swin-UNet & 0.333 & 0.247 & 0.284 & 0.1882 \\
TransUNet & 0.506 & 0.182 & 0.268 & 0.1745 \\
nnU-Net & 0.420 & 0.169 & 0.241 & 0.3249 \\
U-Net & \textbf{0.840} & 0.060 & 0.112 & 0.3582 \\
SegFormer & 0.136 & 0.083 & 0.103 & 0.1840 \\ \hline
\end{tabular}
\end{table}

\noindent \textbf{Small lesions.} All quantities in this block are computed at the lesion level, over connected components rather than pixels. Recall alone is misleading here. U-Net recovers 84.0\% of the small lesions but only 6.0\% of what it marks is one, and the same pattern holds for Swin-UMamba$\dagger$, U-Net++, TransUNet and PSPNet, all of which reach recall comparable to or above the proposed method at less than half its precision. CAT-Seg sits at the other extreme, with the highest precision of any method and the lowest recall among the vision-language ones. Both configurations of VL-AcneSeg are in the upper range on both quantities, which is why they lead on F1, and the global-prompt configuration exceeds every vision-language baseline even though those baselines receive region-level prompts.

\noindent \textbf{Threshold independence.} Both configurations of the proposed method fall in the upper range, so the advantage reported in the main paper does not depend on the choice of threshold. The ordering below them differs from the pixel-level one, since AUPRC also reflects how well a model ranks pixels at thresholds it is never evaluated at. The top group is tightly packed, so AUPRC should be read as separating the stronger methods from the weaker ones rather than as distinguishing between them.

\section{Comprehensive Qualitative Comparison of Acne Segmentation}
\label{sec:appendix_qualitative}

\begin{figure*}[t!]
    \centering
    \includegraphics[width=\textwidth]{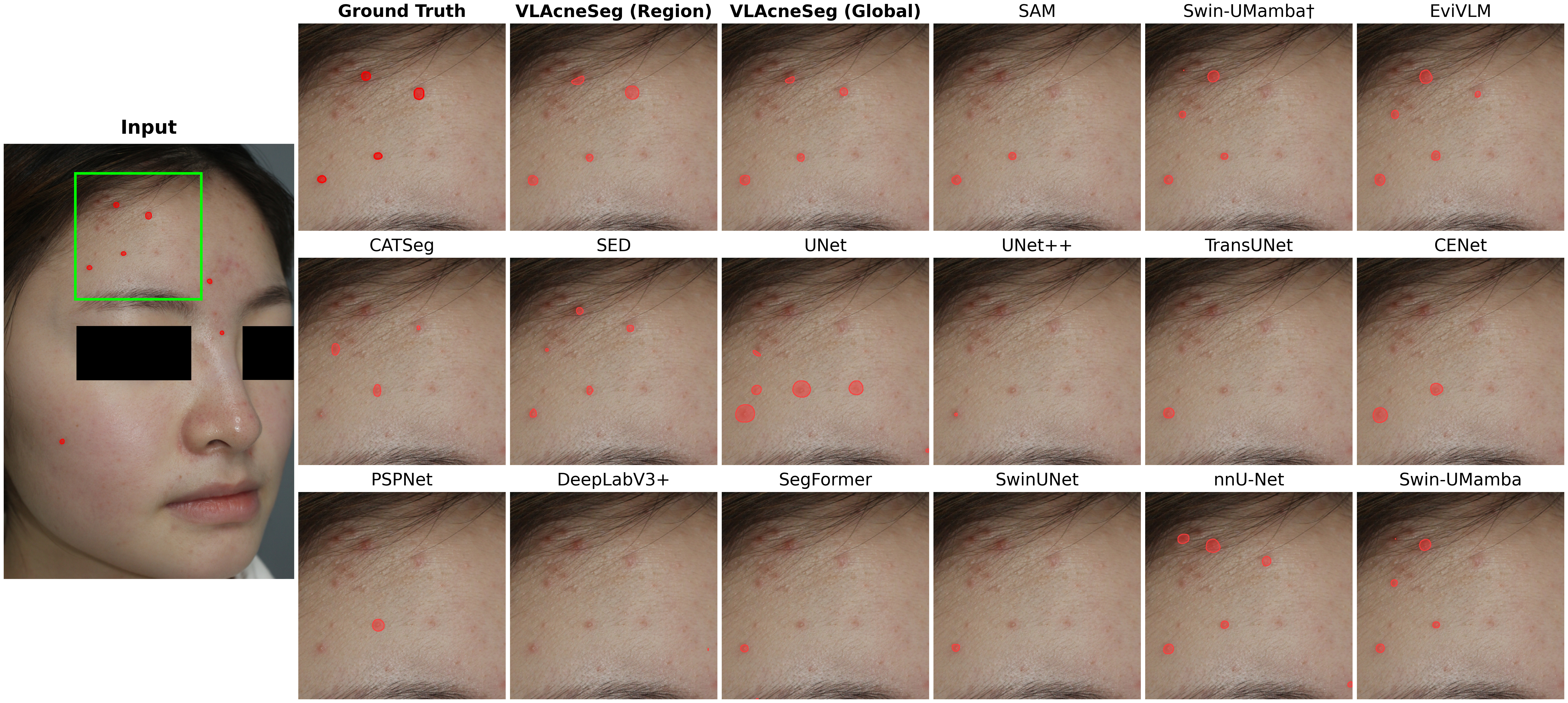}
    \caption{Comprehensive qualitative comparison on the internal dataset (Part I). The left panel shows the input with the reference annotation; the green box marks the region enlarged in the remaining panels. Predicted masks are overlaid in red.}
    \label{fig:comprehensive_a}
\end{figure*}

\begin{figure*}[t!]
    \centering
    \includegraphics[width=\textwidth]{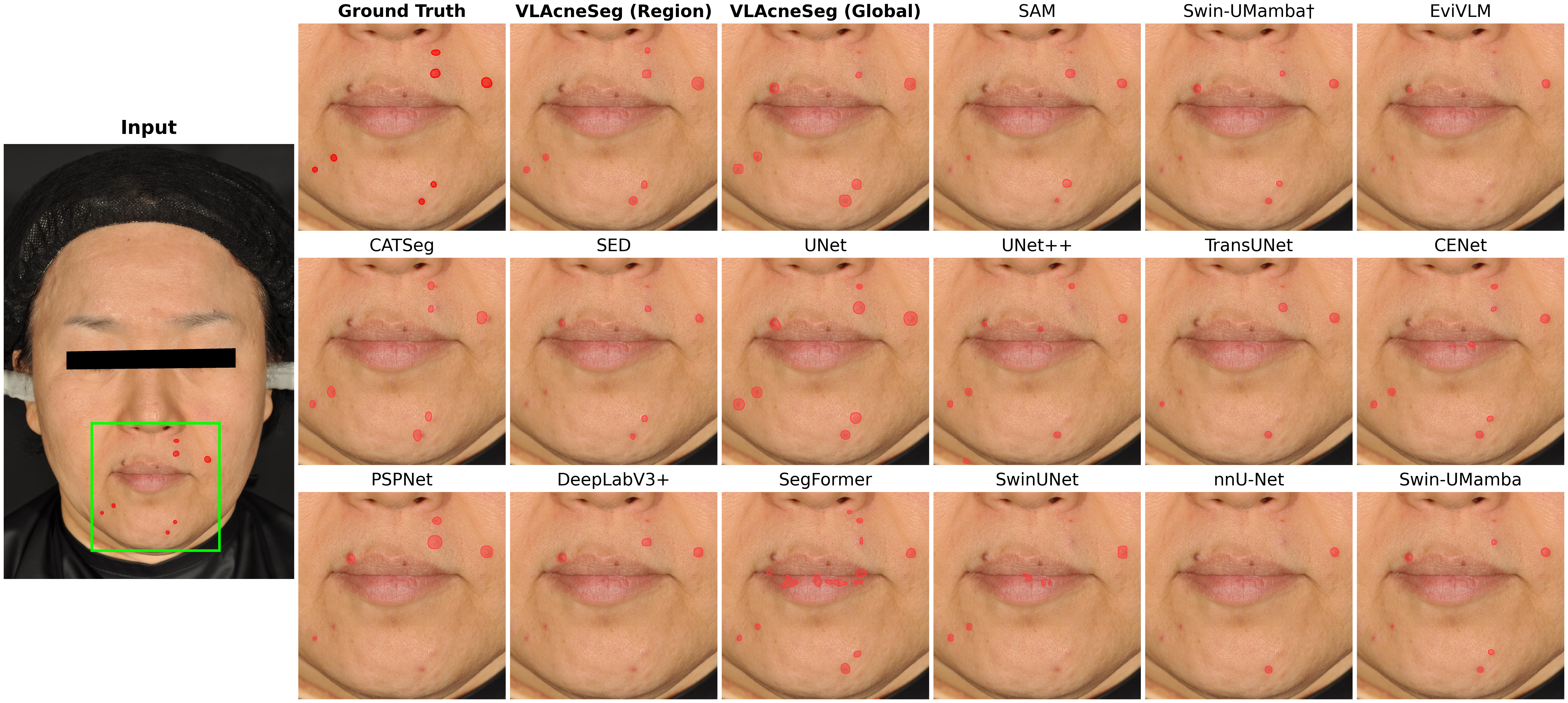}
    \caption{Comprehensive qualitative comparison on the external controlled dataset (Part II).}
    \label{fig:comprehensive_b}
\end{figure*}

\begin{figure*}[t!]
    \centering
    \includegraphics[width=\textwidth]{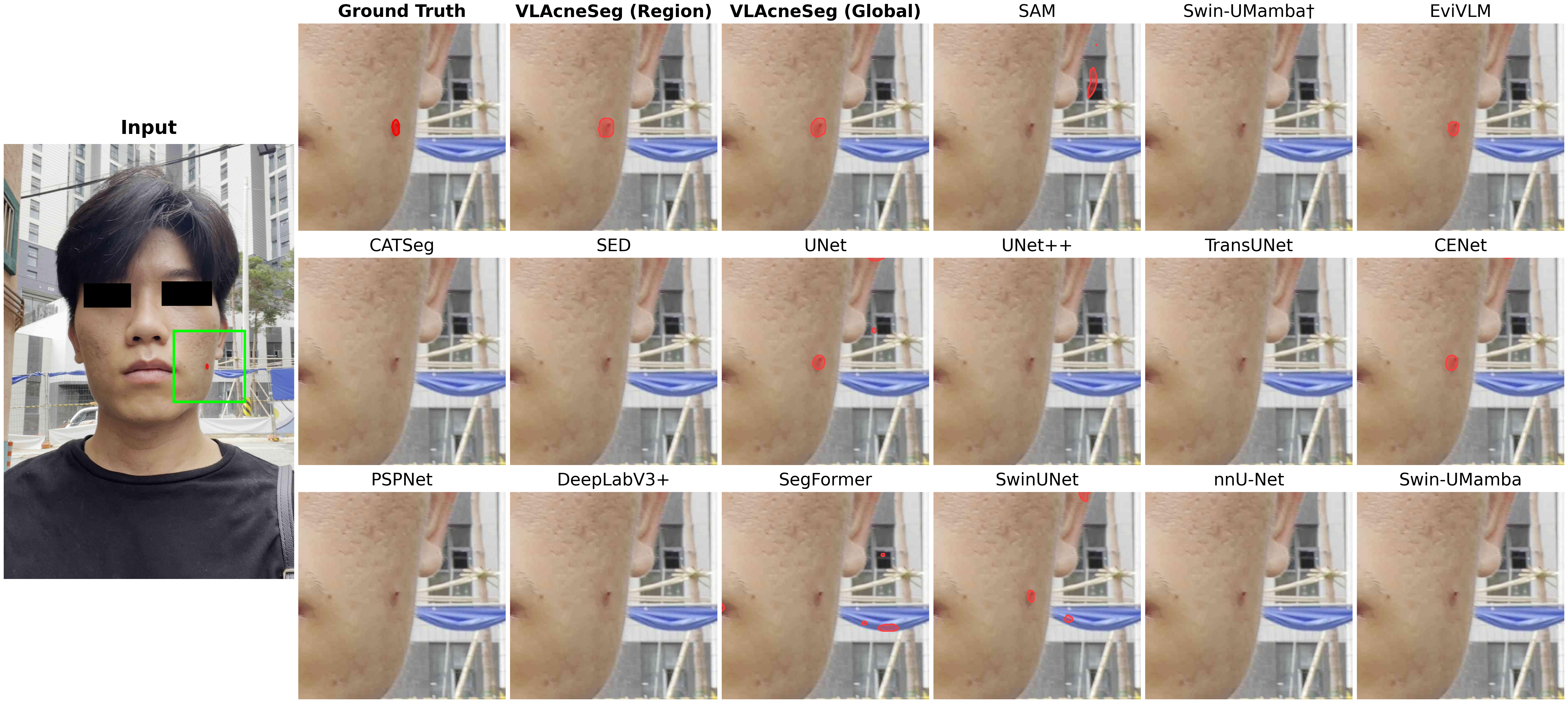}
    \caption{Comprehensive qualitative comparison on the external smartphone dataset (Part III).}
    \label{fig:comprehensive}
\end{figure*}

Figs.~\ref{fig:comprehensive_a}--\ref{fig:comprehensive} extend the qualitative comparison of the main paper to all baselines, on the internal, external controlled and external smartphone datasets. Each figure enlarges a single facial region so that lesions a few pixels across remain visible, and includes both prompting conditions of the proposed method.

\end{document}